\documentclass[lettersize,journal]{IEEEtran}
\usepackage{amsmath,amsfonts}
\usepackage{algorithmic}
\usepackage{algorithm}
\usepackage{array}

\ifCLASSOPTIONcompsoc
    \usepackage[caption=false, font=normalsize, labelfont=sf, textfont=sf]{subfig}
\else
    \usepackage[caption=false, font=footnotesize]{subfig}
\fi
\usepackage{float}

\usepackage{hyperref}
\usepackage[dvipsnames]{xcolor}

\usepackage{textcomp}
\usepackage{stfloats}
\usepackage{url}
\usepackage{verbatim}
\usepackage{graphicx}
\usepackage{cite}
\usepackage{amsfonts,amssymb,amsmath,mathtools,multicol}
\usepackage{physics}
\usepackage{booktabs}
\usepackage{multirow}
\usepackage{mathrsfs}
\usepackage{bm}

\newcommand{\V}[1]{{\boldsymbol{{#1}}}}

\newcommand{\Vdot}[1]{\dot{\V{#1}}}

\newcommand{\Def}{\coloneqq}

\newcommand{\bmat}[1]{\bmqty{#1}}
\newcommand{\MAT}[1]{\bmat{#1}} 

\begin{document}
\title{Adaptive Manipulation Potential and \\ Haptic Estimation for Tool-Mediated Interaction}
\author{Lin Yang$^{1}$, Anirvan Dutta$^{2}$, Yuan Ji$^{1}$, Yanxin Zhou$^{1}$, Shilin Shan$^{1}$, Lv Chen$^{1}$, \\ Etienne Burdet$^{2}$, Domenico Campolo$^{1*}$
\thanks{$^{1}$ authors are with the School of Mechanical and Aerospace Engineering, Nanyang Technological University (NTU), Singapore.}
\thanks{$^{2}$ authors are with the Imperial College of Science, Technology and Medicine, SW7 2AZ London, U.K}
\thanks{$^*$ Corresponding author: {\tt d.campolo@ntu.edu.sg}} 
\thanks{This research is supported by the National Research Foundation, Singapore, under the NRF Medium Sized Centre scheme (CARTIN).}}

\markboth{Journal of \LaTeX\ Class Files,~Vol.~14, No.~8, August~2021}%
{Shell \MakeLowercase{\textit{et al.}}: A Sample Article Using IEEEtran.cls for IEEE Journals}


\maketitle
\begin{abstract}

Achieving human-level dexterity in contact-rich, tool-mediated manipulation remains a significant challenge due to visual occlusion and under-determined nature of haptic sensing. This paper introduces a parameterized Equilibrium Manifold (EM) as a unified representation for tool-mediated interaction, and develops a closed-loop framework that integrates haptic estimation, online planning, and adaptive stiffness control. We establish a physical-geometric duality using an adaptive manipulation potential incorporating a differentiable contact model, which induces the manifold’s geometric structure and ensures that complex physical interactions are encapsulated as continuous operations on the EM. Within this framework, we reformulate haptic estimation as a manifold parameter estimation problem. Specifically, a hybrid inference strategy (haptic SLAM) is employed in which discrete object shapes are classified via particle filtering, while the continuous object pose is estimated using analytical gradients for efficient optimization. Crucially, by continuously updating the parameters of the manipulation potential, the framework dynamically reshapes the induced EM to guide online trajectory re-planning and implement uncertainty-aware impedance control, thereby closing the perception-action loop. The system is validated through simulation and over 260 real-world screw-loosening trials. Experimental results demonstrate robust identification and manipulation success in standard scenarios while maintaining accurate tracking. Furthermore, ablation studies confirm that haptic SLAM and uncertainty-aware stiffness modulation outperform fixed impedance baselines, effectively preventing jamming during tight tolerance interactions.
\end{abstract}

\section{Introduction}

\begin{figure}[!h]
\centering
\includegraphics[width=0.48\textwidth]{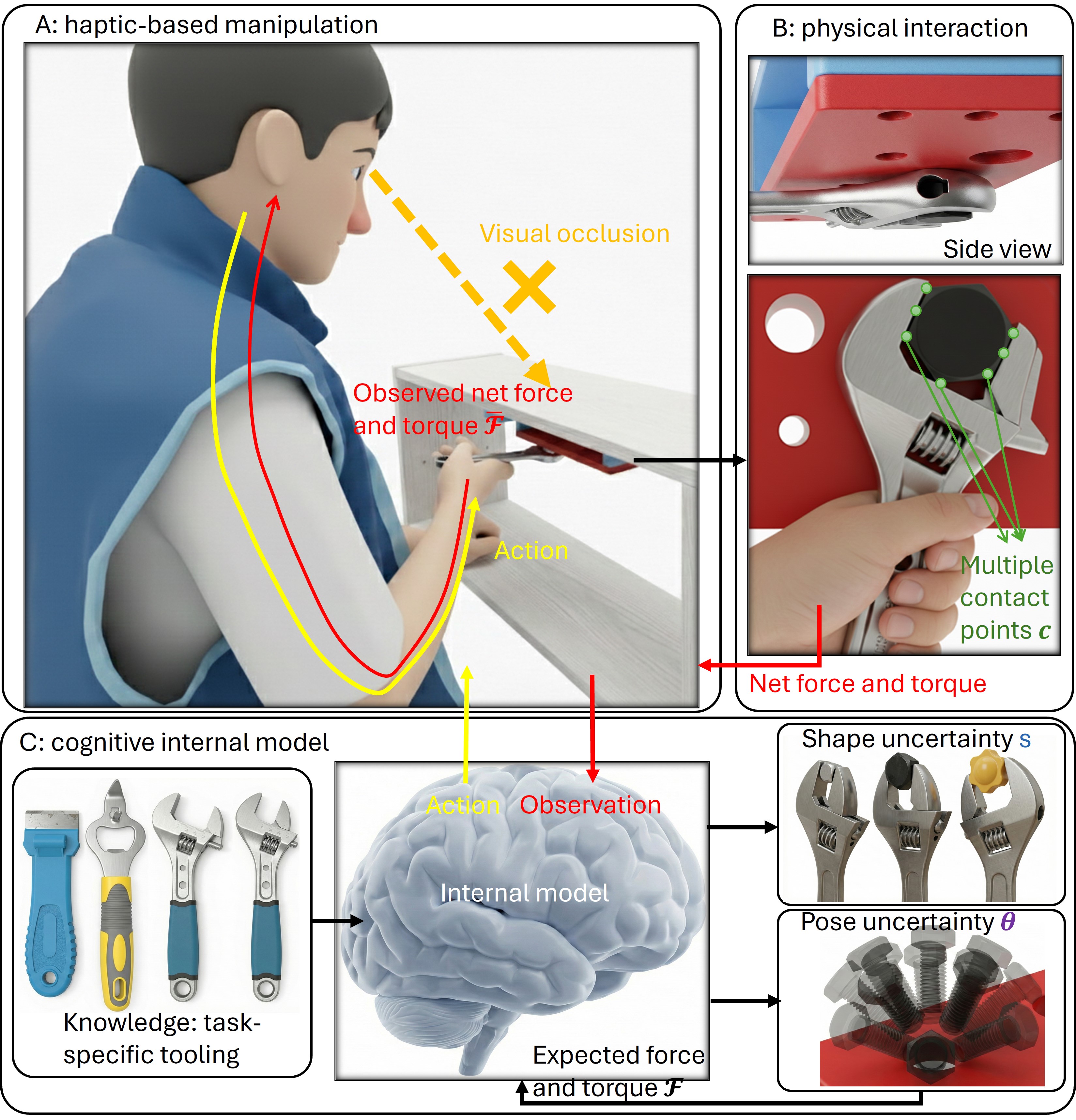}
\caption{Tool-mediated haptic manipulation under visual occlusion. (A) Humans can perform skillful haptic manipulation with tools despite unreliable visual information, although sensing is limited to net forces and torques at the human–tool interface. (B) Multiple contact points may exist at the tool tip while only a single net wrench is observed, leading to contact ambiguity. (C) Humans rely on cognitive reasoning and internal models of tool use to infer task-relevant object properties, such as pose and geometry, through haptic interaction. }
\label{fig:1}
\end{figure}

\IEEEPARstart{E}{quipping} robots with human-level dexterity in contact-rich environments has long been a fundamental pursuit in the field of robotics \cite{johannsmeier2025process}. In contrast to direct end-effector grasping, advanced tasks in industrial and domestic services require the use of diverse tools to accommodate varying task requirements \cite{qin2023robot}. While tool use can significantly expand a robot's capability, it fundamentally reshapes the perception problem: the task-relevant interaction occurs at the tool-environment interface, whereas sensing is typically limited to forces and torques (haptics) at the robot-tool interface. This indirect sensing is critical because equipping every individual tool with sensors is often infeasible due to cost and complexity. Furthermore, during such contact-rich interactions, visual information frequently becomes unreliable due to inevitable occlusion by the tool or the robot itself \cite{li2020review}, as illustrated in Fig. \ref{fig:1} (A). Consequently, in occluded, tool-mediated settings the robot must infer the tool–environment interaction from haptic signals measured at the robot–tool interface, since the contact itself cannot be sensed directly at the tool tip.

Beyond the indirect nature of sensing, tool-mediated perception is fundamentally ill-posed due to inherent measurement ambiguities \cite{Kim_SLAM}. Because sensing is restricted to forces and torques measured at the robot–tool interface, multiple distinct contact points at the tool tip may produce similar wrench observations, as illustrated in Fig.\ref{fig:1} (B). Existing literature demonstrates that if the object's geometry is known a priori, accurate localization can often be achieved by decoding contact cues \cite{miller2024extending}. However, in many real-world assembly tasks, the object's shape, category, and pose are jointly unknown, and the tool tip may engage in complex multi-point interactions. In such settings, these latent variables jointly influence the sparse wrench observations, making it difficult to distinguish between competing environment hypotheses. This challenge reflects a broader principle: haptic perception in tool use extends beyond local contact detection and requires inference of task-relevant states from sparse, partially observable and ambiguous measurements.

Within robotics, a wide range of approaches have investigated haptic perception and contact-rich manipulation. Many existing methods formulate haptic perception as an end goal, aiming to reconstruct complete object geometry or estimate the physical properties through exploration \cite{petrovskaya2011global, bonzini2025robotic, suresh2024neuralfeels}. While effective for comprehensive perception, such approaches often require global or exhaustive exploration and are therefore misaligned with task-driven tool use, where only local, task-relevant interaction is feasible. A complementary body of work focuses on planning and control for contact-rich manipulation, including planning with contact and online re-planning under force feedback \cite{Shirai10160480, zheng2024bayesian, le2024fast}. However, these methods typically assume a fixed and known contact or environment model, limiting their ability to cope with the ambiguity introduced by indirect, tool-mediated sensing \cite{pezzato2025sampling}. Crucially, in tool-mediated tasks, the state of the tool-environment interface is often instantaneously under-determined due to partial observability, necessitating complex temporal inference to resolve contact configurations over time \cite{ota2024tactile}. Existing adaptive control strategies modulate interaction behavior through predefined impedance schedules, demonstrations, or learning-based policies \cite{whitney1982quasi, michel2022safety, ge2024learning, kozlovsky2022reinforcement}, but they frequently treat perception and action as loosely coupled modules. This fragmentation makes it difficult to handle the stochastic and non-smooth dynamics inherent in tool use, such as frequent making and breaking of contacts or nonlinear stick-slip transitions, which remain computationally prohibitive for traditional complementarity-based or analytical models. Ultimately, these limitations highlight the need for a unified representation that integrates task state estimation, planning, and control within a single modeling framework, capable of resolving physical ambiguities through active, task-driven interaction.

This challenge reflects a broader principle widely studied in neuroscience and cognitive science. As pointed out by Helmholtz, perception is not a passive decoding of sensory signals, but an active process in which organisms rely on internal models to predict sensory outcomes and interpret observations through prediction error \cite{barlow1990conditions}. In tool use, humans do not directly perceive contact forces at the tool tip; instead, they leverage an internal representation of interaction mechanics that is continuously refined through action \cite{osiurak2020temporal, lederman1987hand}. In engineering terms, such an internal model can be viewed as a task-specific digital twin \cite{11059840} that generates expected sensory outcomes. As illustrated in Fig.~\ref{fig:1}(C), perception and action are thus coupled through a prediction error-driven loop: actions are selected to probe the interaction, sensory feedback is compared against model predictions, and discrepancies are used to refine the internal representation. However, formalizing this loop in conventional force space presents significant challenges due to the mathematical discontinuities of contact constraints. We argue that these difficulties can be overcome by re-framing the internal model as a geometric manifold. By shifting from fragmented force-based descriptions to a continuous geometric representation, non-smooth physical transitions can be treated as part of a unified, uncertainty-aware inference process.

Motivated by these observations, we develop a unified physics-based framework for tool-mediated, contact-rich manipulation that treats perception, planning, and control within a shared equilibrium manifold. The proposed framework is grounded in quasi-static mechanical formulations, a methodology that has demonstrated significant effectiveness in predicting contact forces during assembly \cite{tracy2025efficient} and in resolving complex jamming through a feedback controller \cite{halm2018quasi}. By neglecting inertial and Coriolis effects, this quasi-static approach focuses on interaction and enables the robot, object, and environment to be modeled as a unified manipulation potential subject to implicit equilibrium constraints \cite{campolo2023quasi, CAMPOLO2025116003}. This viewpoint provides a physically consistent and meaningful geometric representation of interaction that avoids the need for explicit contact mode scheduling \cite{tracy2025efficient} or complex complementarity solvers \cite{halm2018quasi}. Within this formulation, system behavior is characterized as motion along an equilibrium manifold. Latent environment variables, such as object pose and structural properties, parameterize this manifold, allowing the instantaneously under-determined task state to be inferred through operations on a shared equilibrium manifold. By grounding perception and action in this representation, our framework goes beyond existing decoupled methods, providing a principled foundation for resolving physical ambiguity through active interaction.

Our contributions are summarized as follows:
\begin{enumerate}
    \item \textbf{A unified physics-based framework:} We develop a real-time, physics-based framework built upon a parameterized Equilibrium Manifold that tightly integrates haptic state estimation, manipulation planning, and impedance control within a single formulation.

    \item \textbf{A tool-mediated manipulation perspective:} We formulate indirect, tool-mediated manipulation as a simultaneous perception–action problem, enabling effective manipulation through sparse, local, task-relevant exploration rather than full geometric reconstruction.

    \item \textbf{A differentiable contact model:}  We propose a smooth, differentiable contact model based on a signed-distance field and point clouds that embeds multi-point contact into a single manipulation potential. This formulation constructs a continuous equilibrium manifold endowed with an intrinsic metric.

    \item \textbf{Hybrid haptic SLAM:} We present a haptic SLAM framework that jointly estimates continuous object pose and discrete object types by combining analytical gradient-based optimization with particle-based inference driven by haptic mismatch.

    \item \textbf{Uncertainty-aware online planning and variance informed stiffness control:} We integrate haptic state estimation into online planning and impedance control, enabling adaptive regulation of interaction behavior and a principled transition from exploratory interaction to task execution.
\end{enumerate}

\section{Related Work}
This work lies at the intersection of haptic estimation, physical interaction modeling, and contact-rich manipulation. Accordingly, we organize the related literature into four interconnected themes: We first review (A) task formulation in haptic and tool-mediated interaction, distinguishing between perception-driven and task-driven objectives. We then discuss (B) physical modeling of contact and interaction, with an emphasis on force-based and equilibrium-based formulations. Next, we survey (C) haptic state estimation and inference methods, contrasting learning-based and physics-based approaches. Finally, we review (D) planning and control strategies for contact-rich manipulation, highlighting their assumptions about contact and perception–action coupling.
\vspace{-2mm}
\subsection{Task Formulation in Haptic and Tool-Mediated Interaction}\label{secA}

Haptic perception primarily relies on joint-level force torque sensors, which provide net interaction signals for perceiving the physical world when visual feedback is occluded or insufficient \cite{newbury2023deep}. A large body of prior work formulates haptic perception as an end goal, aiming to recover complete object properties such as shape, physical parameters, or pose through active exploration. Within this perception-driven paradigm, exploration strategies are designed to maximize information about the object itself: shape estimation often requires traversing the full object contour or probing from multiple directions \cite{petrovskaya2011global, zheng2024bayesian, suresh2024neuralfeels}; physical property estimation using interactive prehensile or non-prehensile actions to excite dynamic modes \cite{dutta2025predictive}; and pose estimation is frequently preceded by an explicit reconstruction phase to build a geometric or physical model \cite{rostel2022learning}. While effective for comprehensive perception, this perception-first and full model acquisition strategy is unnecessary for many manipulation tasks whose objective is task completion rather than object reconstruction. Moreover, the very conditions that motivate haptic sensing, such as visual occlusion or confined workspaces, often make global exploration physically infeasible, necessitating local, task-relevant interaction instead \cite{pan2023hand}.

Beyond direct contact scenarios, increasing attention has been given to tool-mediated haptic interaction, where the robot infers properties of the environment through an intermediate tool rather than via direct end-effector contact. This setting is commonly referred to as extrinsic contact sensing \cite{ma2021extrinsic} or indirect perception \cite{aoyama2025few}. In such problems, sensing is limited to forces and torques measured at the robot–tool interface, while the contact of interest occurs at the tool–environment interface, fundamentally altering the perceptual problem. Multiple extrinsic contact configurations may generate similar intrinsic wrenches at the handle, rendering inference inherently under-determined \cite{Kim_SLAM}. Prior studies have shown that resolving this ambiguity often requires exploiting additional structure, such as task constraints, motion history, or high-dimensional tactile feedback \cite{sipos2022simultaneous, ota2024tactile}. Thus, many existing approaches treat extrinsic contact inference as an isolated sensing problem. In contrast, practical tool-use tasks require a task-driven formulation, where contact is integrated into manipulation and used to infer task-relevant states. This motivates a task-conditioned and local haptic perception framework tightly coupled with the manipulation process.
\vspace{-2mm}
\subsection{Physical Modeling of Contact and Interaction}
Contact modeling in robotics is commonly categorized into two paradigms: force based constraint formulations and energy based potential formulations. As analyzed in the comparative study of contact models in robotics \cite{le2024contact}, most modern physics engines (e.g., MuJoCo \cite{todorov2012mujoco}, RaiSim \cite{hwangbo2018per}, and Drake \cite{drake}) adopt force-based approaches, typically formulated as Nonlinear or Conic Complementarity Problems (NCPs and CCPs). These methods enforce non-penetration and friction constraints through contact forces or impulses, often relying on physical relaxations such as artificial compliance or dissipation to ensure numerical tractability \cite{le2024contact}. While effective for dynamic simulation, such relaxations may introduce internal forces or inconsistent energy behavior, which can be problematic in haptic settings where force–torque measurements are central.

An alternative perspective, explored primarily in quasi-static manipulation \cite{whitney1982quasi} and differentiable simulation, models interaction through energy-based or equilibrium-consistent formulations, where contact forces arise from potential gradients \cite{CAMPOLO2025116003}. Such formulations yield smoother and differentiable interaction mappings, which are particularly important for gradient-based optimization and manipulation planning \cite{toussaint2018differentiable}. Compared to complementarity-based solvers \cite{halm2018quasi}, these approaches offer smoother interaction mappings but have been less studied in complex, tool-mediated manipulation scenarios. These differences become particularly consequential when contact models are used not only for simulation, but as internal models for state estimation and planning under uncertainty.
\vspace{-1mm}
\subsection{Haptic State Estimation and Inference}
To address the estimation challenges outlined in Section \ref{secA}, existing approaches to haptic state estimation can be broadly divided into learning-based and physics-based methodologies. Learning-based approaches have become increasingly popular due to their ability to handle high-dimensional sensor data and unmodeled dynamics. Representative works employ graph neural networks to learn interaction dynamics \cite{dutta2025predictive}, neural fields for in-hand haptic SLAM \cite{suresh2024neuralfeels}, imitation learning for few-shot tool-use perception \cite{aoyama2025few}, convolutional networks to interpret tactile observations \cite{Kim_SLAM}, or linear model learning (LML) to learn mapping matrices \cite{tracy2025efficient}. These methods have demonstrated strong empirical performance in challenging sensing scenarios, but typically operate as black-box models and do not explicitly expose the underlying physical structure or analytical gradients of the interaction model.

An alternative line of work adopts physics-based estimation, emphasizing interpretability and data efficiency through explicit modeling of contact and motion. Early approaches formulated Bayesian MCMC methods for 6-DOF object localization using haptic feedback \cite{petrovskaya2011global}, followed by particle filter based estimators \cite{zheng2024bayesian}, kinematic optimization techniques \cite{ma2021extrinsic}, and factor graph based tactile SLAM formulations \cite{Kim_SLAM}. While effective in their respective settings, these methods often simplify aspects of physical interaction, such as frictional effects, and commonly rely on sampling or generic numerical solvers. Moreover, existing frameworks typically focus on either continuous state estimation or discrete hypothesis selection, rather than jointly addressing both within a unified inference process. Together, these characteristics motivate approaches that combine structured physical modeling with analytical differentiation and hybrid inference capabilities for contact-rich manipulation.
\vspace{-1mm}
\subsection{Planning and Control for Contact-Rich Manipulation}
Traditional robot motion planning is primarily designed to avoid contact \cite{lavalle1998rapidly}, whereas many manipulation tasks require planning with contact to deliberately exploit physical interaction \cite{zheng2024bayesian}. Contact-Implicit Trajectory Optimization (CITO) provides a principled formulation for non-smooth contact but is typically limited to offline use due to its computational cost \cite{kurtz2022contact}. Recent equilibrium manifold formulations reason about interaction through physical consistency \cite{yang2025planning_book}, yet similarly rely on offline optimization and offer limited adaptability under uncertainty. MPC and MPPI-based methods enable online re-planning for contact-rich manipulation \cite{le2024fast}, but fundamentally rely on accurate forward models and do not explicitly resolve contact ambiguity under partial observability \cite{pezzato2025sampling}. In tool-mediated settings, haptic feedback has been incorporated into MPC to estimate object pose for re-planning \cite{Shirai10160480}, though the environment and contact model are still typically assumed fixed and known. When uncertainty is addressed in planning, information gain is commonly used as an objective, but existing formulations largely assume direct, single point contact \cite{dutta2025predictive} (Sec.~\ref{secA}), leaving tool-mediated interaction with complex contact mechanisms insufficiently addressed.
 
Beyond trajectory generation, successful execution of contact-rich tasks critically depends on compliance. Fixed impedance control is prone to jamming and excessive internal forces under major misalignment \cite{whitney1982quasi}. Variable Impedance Control (VIC) and adaptive strategies have therefore been proposed to modulate stiffness during interaction. Approaches such as virtual energy tanks ensure passivity and stability \cite{michel2022safety}, but rely on predefined attractors obtained from demonstration, resulting in time based rather than perception driven stiffness modulation that cannot accommodate unknown target poses. Other methods learn stiffness adaptation from human strategies \cite{ge2024learning} or via reinforcement learning under small misalignment \cite{kozlovsky2022reinforcement}. As a result, perception, planning, and control are often loosely coupled. This motivates frameworks that tightly integrate haptic state estimation with planning and control, allowing uncertainty arising from contact-rich interaction to directly inform both decision-making and interaction regulation.
\begin{figure*}[!t]
\centering
\includegraphics[width=0.8\textwidth]{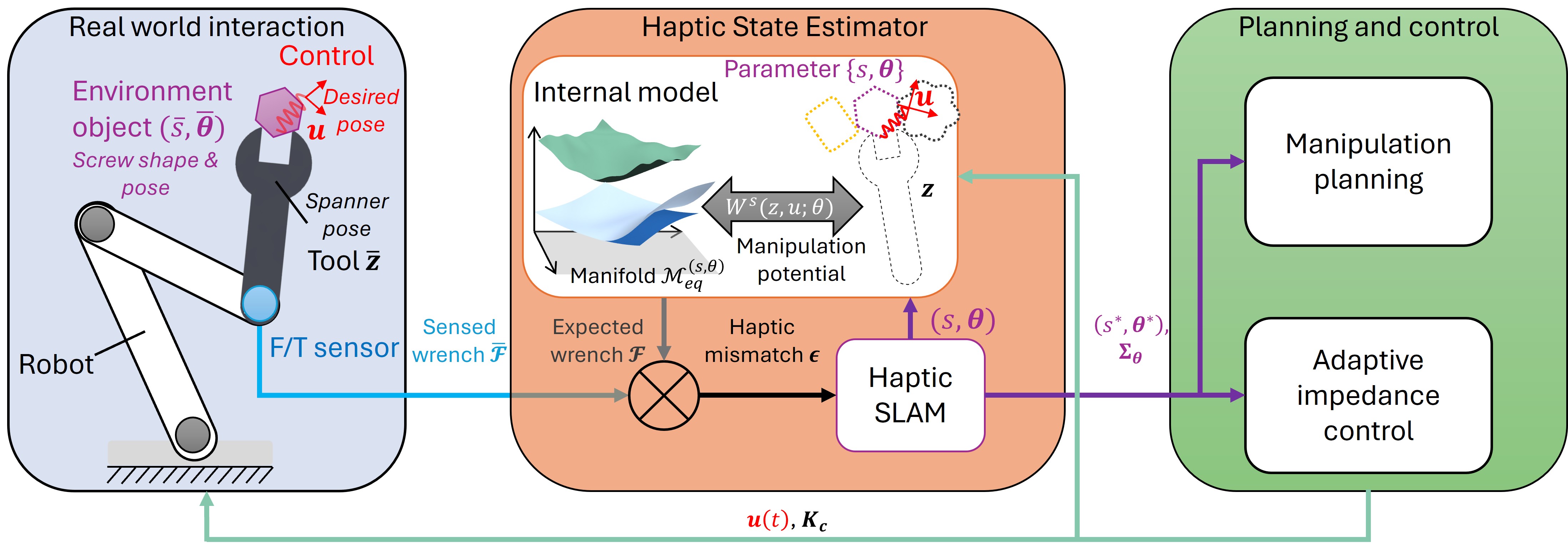}
\caption{Overall workflow of tool-mediated manipulation framework. Left: the robot uses a tool to interact with the object and senses the resulting net wrench $\overline{\bm{\mathcal{F}}}$. Middle: $\overline{\bm{\mathcal{F}}}$ is fed into a haptic state estimator that incorporates a manifold-based internal physical model to predict the expected wrench $\bm{\mathcal{F}}$. The discrepancy between the observed and predicted wrenches—termed \textit{haptic mismatch}—is used to update the estimated object pose $\V \theta$ and type $s$.
Right: Based on the optimal estimates $(s^*, \V \theta^*)$ and associated uncertainty $\V \Sigma_{\V \theta}$, a planner and controller generate the control $\V u(t)$ and stiffness $\V K_c$ for robot.
}
\label{fig:workflow}
\end{figure*}

\begin{table}[!t]
\centering
\caption{Nomenclature and Key Symbols}
\label{tab:nomenclature}
\renewcommand{\arraystretch}{1.2} 
\begin{tabular}{l p{0.75\linewidth}} 
\toprule
\textbf{Symbol} & \textbf{Description} \\
\midrule
\multicolumn{2}{l}{\textit{A. Internal Model \& Equilibrium Manifold (Sec. \ref{sec: equilibrium mfd})}} \\
$\V z \in \mathcal{Z}$ & Internal system state (e.g., spanner pose), evolving on the manifold \\
$\V u \in \mathcal{U}$ & Robot control input (e.g., impedance equilibrium pose) \\
$\V \theta \in \Theta$ & Continuous environment parameter (e.g., screw pose) to be estimated \\
$s \in \mathcal{S}$ & Discrete environment parameter (e.g., spanner and screw shape/type) \\
$W^{(s)}(\V z, \V u; \V \theta)$ & Quasi-static manipulation potential function for shape hypothesis $s$ \\
$\mathcal{M}_{eq}^{(s, \V \theta)}$ & Equilibrium manifold defined by the constraint $\partial_{\V z} W = 0$ \\
$\V c_i(\V z)$ & Position of the $i$-th point cloud node in the world frame \\
$F_{SQ}(\cdot)$ & Superquadric implicit function defining object geometry \\
\midrule
\multicolumn{2}{l}{\textit{B. Haptic SLAM \& Hybrid Inference (Sec.\ref{sec:pose} to \ref{sec:shape})}} \\
$\overline{\bm{\bm{\mathcal{F}}}}$ & \textbf{Observed} wrench (net force and torque) from the F/T sensor \\
$\bm{\bm{\mathcal{F}}}$ & \textbf{Predicted} wrench derived from the internal physical model \\
$\V \epsilon$ & Haptic mismatch (residual) vector, $\V \epsilon = \overline{\bm{\mathcal{F}}} - \bm{\mathcal{F}}$ \\
$\V \Sigma$ & Covariance matrix of the measurement noise \\
$J(\V u, \V \theta)$ & Explicit analytical Jacobian $\partial \V \epsilon / \partial \theta$ \\
$\mathcal{B}_b$ & Index set of time steps belonging to the $b$-th batch \\
$\theta_b^*[s^{(i)}]$ & Representative pose estimate for the $i$-th shape particle in batch $b$ \\
$\{s^{(i)}, w^{(i)}\}$ & Set of discrete shape particles and their associated probability weights \\
\midrule
\multicolumn{2}{l}{\textit{C. Planning \& Control (Sec. \ref{sec:planning_control})}} \\
$\V \beta$ & Parameters of the Dynamic Motion Primitives (DMP) \\
$\psi(t)$ & Auxiliary state representing accumulated haptic cost along a trajectory \\
$\V \Sigma_{\V \theta}$ & Covariance matrix of the estimated pose distribution \\
$\V K_c$ & Impedance stiffness matrix \\
$\kappa$ & Stiffness anisotropy ratio ($k_n / k_{\phi}$) modulated by uncertainty \\
\bottomrule
\end{tabular}
\end{table}
\section{Problem Definition}
In this work, we study tool-mediated manipulation, using a screw-loosening task as a representative example. A robot equipped with a tool (e.g., a spanner) $\mathbf{z}$ interacts with an object $(s, \mathbf{\theta})$ (e.g., a screw). The robot maintains an internal model that predicts sensory outcomes and guides action generation. Given candidate hypotheses of object shape $s$ and pose $\V \theta$, the model computes the expected interaction wrench $\bm{\mathcal{F}}$ for the tool pose $\V z$. Crucially, both the physical system and the internal model share the same control input $\V u$, representing the robot’s end-effector pose. For clarity, internal variables follow standard notation, while real-world observations are indicated with an overline $\overline{[\cdot]}$. For example, $\overline{s}$ denotes the screw shape, $\overline{\V \theta}$ its measured pose, and $\overline{\bm{\mathcal{F}}}$ the observed interaction wrench (net force and torque) measured at the tool handle.

The expected interaction wrench $\bm{\mathcal{F}}$ is derived from an adaptive manipulation potential $W^{(s)}(\V z, \V u; \V \theta)$ and its induced equilibrium manifold $\mathcal{M}_{eq}^{(s, \V \theta)}$ (Sec. \ref{sec: equilibrium mfd}). The discrepancy between the observed and predicted wrenches, defined as the \textit{haptic mismatch} $\V \epsilon = \overline{\bm{\mathcal{F}}} - \bm{\mathcal{F}}$, is used to recursively refine the estimates $(s, \V \theta)$ (Sec.\ref{sec:pose} to \ref{sec:shape}). Finally, the optimal estimates $(s^*, \V \theta^*)$ and uncertainty $\V \Sigma_\theta$ inform an online re-planner to generate the next trajectory $\V u(t)$ and impedance stiffness $\V K_c$ for closed-loop (Sec.\ref{sec:planning_control}). Fig.~\ref{fig:workflow} illustrates the overall workflow of the proposed framework. Essential notations are listed in Table \ref{tab:nomenclature} and discussed in detail in subsequent sections.

\section{Parameterized Equilibrium Manifolds and Physical Modeling}\label{sec: equilibrium mfd}
This section establishes the physics-based framework for tool-mediated, contact-rich manipulation. We present a unified formulation that explicitly bridges physical interaction modeling and geometric manifold constraints through a shared mathematical structure. Building upon the background of quasi-static equilibrium manifolds \cite{campolo2023quasi, CAMPOLO2025116003} and navigation strategies while avoiding haptic singularities \cite{yang2025planning_book}, we introduce here two key advancements over existing formulations. First, we propose a novel parameterization of the mechanical system where the environment is characterized by unknown discrete and continuous variables, rather than assuming prior knowledge of these variables. Second, unlike previous methodologies limited to simpler geometries, we propose a differentiable, manifold-inducing contact model. By integrating \textit{signed-distance fields} (SDF) and \textit{point clouds} (PC) into the manipulation potential, the framework can accommodate complex, multi-point contact geometries while preserving analytical tractability for gradient-based optimization. This correspondence ensures that the system's equilibrium behavior is constrained by its underlying physics. By grounding our modeling in this physical-geometric duality, we provide a principled foundation for the haptic SLAM, planning, and adaptive control developed in Sec. \ref{sec:haptic_SLAM}.

\subsection{Overview: Physical and Geometric Dual Perspective}
\label{sec:discussion_physics_geometry}
The formulation presented in this section admits two complementary interpretations, establishing a physical-geometric duality as illustrated in the internal model of Fig.\ref{fig:workflow}. These two views are linked through the manipulation potential $W^{(s)}(\V z, \V u; \V \theta)$, which maps physical interaction model to equilibrium constraints and induces the corresponding manifold geometry. 
From a geometric perspective, the same system can be viewed as a family of equilibrium manifolds $\mathcal{M}_{eq}^{(s, \V \theta)} \subset \mathcal{Z} \times \mathcal{U}$. In this representation, the environment variables $(s, \V \theta)$ [e.g., discrete shape hypotheses $s^{(i)}$ and their corresponding continuous poses $\V \theta^{(j)}(s^{(i)})$] parameterize the manifold and determine its geometry. Consequently, the system's motion $\V z$ then emerges in response to control inputs $\V u$ acting upon this geometric structure. For details, please refer to Sec.\ref{sec:qs_system}.
From a physical perspective, the system is described by control $\V u$, tool $\V z$ and object in the environment $(s, \V \theta)$. Interactions are modeled explicitly through contact geometry and impedance control. This perspective gives rise to a manipulation potential $W^{(s)}(\V z, \V u; \V \theta)$, which encapsulates the physical constraints of the task and interactions within the system. Detailed instantiations are provided in Sec.\ref{sec:contact_model}.

\subsection{Parameterized Quasi-Static Mechanical System} \label{sec:qs_system}
Under the quasi-static assumption, we model the robot control, tool, and environment as a single interconnected mechanical system. Let 
\begin{itemize} 
\item $\V z \in \mathcal{Z} \subset \mathbb{R}^N$ denote the \emph{internal state} of the system, which can be interpreted as the tool that evolves implicitly through physical interaction; 
\item $\V u \in \mathcal{U} \subset \mathbb{R}^K$ be the \emph{control input} of the robot, interpreted as the desired end-effector pose under impedance control. 
\end{itemize}
The environment is characterized by two types of parameters: 
\begin{itemize} 
\item a \emph{discrete structural parameter} $s \in \mathcal{S}$, representing categorical properties of the environment (e.g., spanner or screw type), which do not admit differentiation and instead index distinct mechanical models.
\item a \emph{continuous parameter} $\V \theta \in \V {\Theta}$, representing quantities such as environment object pose, which admit a differentiable structure.
\end{itemize}
For each physical model $s$, the quasi-static interaction between the robot, tool, and environment is governed by a \emph{manipulation potential}
\begin{equation}
W^{(s)} : \mathcal{Z} \times \mathcal{U} \times \V \Theta \rightarrow \mathbb{R}, \; (\V z,\V u;\V \theta) \mapsto W^{(s)}(\V z,\V u;\V \theta), 
\end{equation} 
that encodes all necessary potential to describe the mechanical system (e.g., contact potentials and impedance control). Throughout this section, all derivatives are taken with respect to $(\V z,\V u)$, conditioning on environment parameters $(s,\V \theta)$. To describe the interaction between the robot and the objects, we define the control forces as $\bm{\mathcal{F}}_\text{ctrl} =  -\partial_\V{u}W$ and the net generalized internal forces as $\bm{\mathcal{F}}_\text{int} =  -\partial_\V{z}W$ \cite{CAMPOLO2025116003}, where $\partial_{\V \alpha} W$ is defined component-wise: for $\partial_{\V \alpha} = [\partial_{\alpha_1}, \ldots, \partial_{\alpha_a}]^T$, $\partial_{\V \alpha} W \equiv [\partial_{\alpha_1} W, \ldots, \partial_{\alpha_a} W]^T$. Furthermore, we define $\partial^2_{\V \alpha \V \beta} \equiv \partial^{}_{\V \alpha} \partial_{\V \beta}^T$ for second  order derivatives.

For an environment hypothesis $(s,\V \theta)$, we consider a manipulation potential $W^{(s)}(\V z, \V u;\V \theta)$. An \textit{equilibrium configuration} $\V z^*(\V u; \V \theta,s)$ is then defined implicitly by the force balance condition
\begin{equation} 
\partial_{\V z} W^{(s)}(\V z^*,\V u;\V \theta) = 0\,, 
\label{eq:fz0}
\end{equation} 
which corresponds to the state where $\bm{\mathcal{F}}_\text{int} = \V 0$. Assuming that the Hessian $\partial^2_{\V z \V z} W^{(s)}$ is full rank at equilibrium, the implicit function theorem guarantees the existence of a smooth local mapping $\V u \mapsto \V z^*(\V u;\V \theta,s)$ \cite{spivak2018calculus}. The corresponding \emph{equilibrium manifold} is defined as \begin{equation} 
\mathcal{M}_{eq}^{(s,\V \theta)} \Def \{(\V z,\V u)\in\mathcal{Z}\times\mathcal{U} | \partial_{\V z} W^{(s)}(\V z,\V u;\V \theta)=0\}
\end{equation}
For each parameter pair $(s,\V \theta)$, $\mathcal{M}_{eq}^{(s,\V \theta)}$ is a smooth embedded submanifold of $\mathcal{Z} \times \mathcal{U}$. 

\begin{figure}[H]
\centering
\includegraphics[width=\columnwidth]{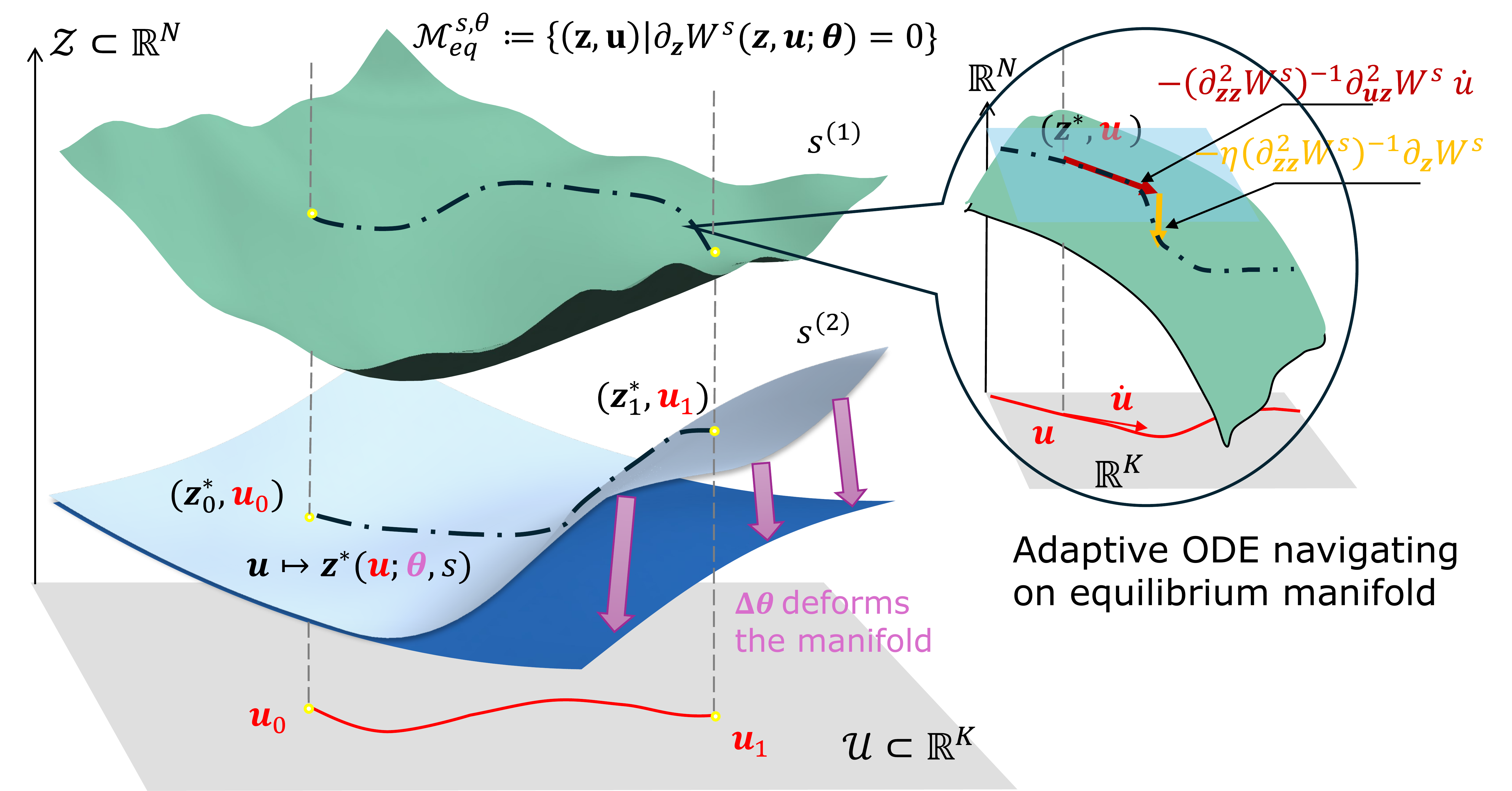}
\caption{%
Parameterized equilibrium manifolds and induced motion.
For fixed environment parameters $(s,\V \theta)$, the quasi-static interaction between the robot and the environment defines an
equilibrium manifold $\mathcal{M}_{eq}^{(s,\V \theta)}$
embedded in the state-control space.
Different values of the discrete parameter $s$ select different
manifolds, while the continuous parameter $\V \theta$ continuously
deforms the manifold geometry within a given structural model.
For fixed $(s,\V \theta)$, control inputs $\V u$ induce equilibrium configurations $\V z^*(\V u;\V \theta,s)$ along the manifold. The resulting motion on $\mathcal{M}_{eq}^{(s,\V \theta)}$ is governed by an adaptive ODE.}\label{fig: adaptive ODE}
\end{figure}

As illustrated in Fig.\ref{fig: adaptive ODE}, the discrete parameter $s$ indexes a set of candidate mechanical models, where the number of distinct manifolds depends on the number of possible shape hypotheses. Varying $s$ selects a different mechanical model (e.g., the jump from green manifold $s^{(1)}$ to blue manifold $s^{(2)}$), while varying $\V \theta$ deforms the manifold to minimize the gap between the internal expectation (light blue) and the physical reality (dark blue). Importantly, neither $s$ nor $\V \theta$ are treated as state variables on the manifold itself; instead, they parameterize a family of equilibrium manifolds. 

Following \cite{campolo2023quasi, CAMPOLO2025116003}, we define a Riemannian metric on the control space $\mathcal{U}$ induced by the quasi-static interaction. Conditioning on $(s,\V \theta)$, \emph{haptic metric} is given by 
\begin{equation}
\resizebox{0.9125\linewidth}{!}{$
\V G^{(s)}(\V u;\V \theta) := \partial^2_{\V u \V u} W^{(s)}-\partial^2_{\V u \V z} W^{(s)} \left[\partial^2_{\V z \V z} W^{(s)} \right]^{-1} \partial^2_{\V z \V u} W^{(s)}
$}
\end{equation}
Here, we focus on a local analysis around a single, specific equilibrium solution: all derivatives are evaluated at the equilibrium $\V z^*(\V u;\V \theta,s)$, which is computed as the Schur complement of the Hessian of the potential function. 
For a control trajectory $\V u(\gamma):[0,1] \to \mathcal{U}$, the associated \emph{haptic distance} is defined as 
\begin{equation}\label{eq:haptic_dis}
\Gamma[\V u] = \int_0^1 \!\!\! \sqrt{ \Vdot u^\top \V G^{(s)}(\V u;\V \theta) \, \Vdot u } \,\, d\gamma \,. 
\end{equation}

\subsection{Motion Along the Equilibrium Manifold} \label{sec:em_motion}
\subsubsection{Navigation on the Equilibrium Manifold}
As the equilibrium manifold is defined implicitly, object motion must be computed jointly with the evolution of the control input. Conditioning on $(s,\V \theta)$, the evolution of the internal state is governed by the adaptive ordinary differential equation when the system is frictionless \cite{yang2025planning_book} 
\begin{equation} 
\resizebox{0.9125\linewidth}{!}{$
\Vdot z_{\mathrm{free}} = - \left[ \partial^2_{\V z \V z} W^{(s)}\right]^{-1} \partial^2_{\V u \V z} W^{(s)} \Vdot u - \eta \left[\partial^2_{\V z \V z} W^{(s)}\right]^{-1} \partial_z W^{(s)}
$}
\label{eq:ode0}
\end{equation} 
where the first term captures the infinitesimal coupling between control variation and object motion, and the second term enforces numerical stability by correcting deviations from equilibrium. The scalar $\eta > 0$ controls the convergence rate toward the manifold \cite{schneebeli2011newton}. This dynamics ensures that trajectories remain on $\mathcal{M}_{eq}^{(s,\V \theta)}$ while the robot executes control commands.

\subsubsection{Haptic Obstacle}
To ensure robot stability and avoid singularities on the equilibrium manifold haptic obstacles are defined as \cite{yang2025planning_book}:
\begin{equation}\label{eq: Haptic Obstacle}
\text{det} \! \left[ \partial^2_{\V z \V z} W^{(s)} \right] \geq \lambda > 0
\end{equation}
This constraint ensures a non-critical region where the Hessian is strictly positive definite, satisfying the requirements for the implicit function theorem and guaranteeing a well-defined inverse for the system dynamics in Eq. \ref{eq:ode0}.
\vspace{-1mm}
\subsection{Instantiation of the Manipulation Potential}
\label{sec:contact_model}
To move from the general theoretical framework to a physical model, we must define a concrete analytical form for the manipulation potential $W^{(s)}(\V z,\V u; \V \theta)$ introduced in Sec. \ref{sec:qs_system}. This instantiation is essential to bridge the gap between geometric formulation and manifold representation, allowing the complex mechanics of contact-rich tool use to be characterized as a structured manipulation potential. While the proposed formulation is applicable to both planar and spatial manipulation, we present a two-dimensional (2D) example. The extension to 3D will follow directly from the same principles.

\subsubsection{Parameterization and Notation} In 2D, the manipulated tool is parameterized by a continuous pose $\V z = [x, y, \phi]^\top$, thus the transform matrix $\V T(\V z)\in SE(2)$ is defined as
\begin{equation}  
\V T(\V z) := \MAT{\V R(\phi)  & \begin{array}{c} x \\ y \end{array} \\ \V 0^\top & 1}.
\label{eq:T}
\end{equation}
For the environment, the transformation $\V T(\V \theta)\in SE(2)$ describes the object pose in world frame. In addition, a discrete shape parameter $s \in \mathcal{S}$ indexes different geometric hypotheses (e.g., screw types).  The pair $(s,\V \theta)$ fully determines the contact geometry and, consequently, the form of the manipulation potential.

Assuming a rigid tool, the point cloud is represented by a finite set of points $\{\V c'_i\}$ whose positions are invariant in the tool-fixed frame $\V z$, determined by its physical geometry. Rigid transformation is used to compute point cloud in world frame
\begin{align} 
\V c_i(\V z) & = \V T (\V z) \circ \V c'_i \nonumber \\
& = \MAT{1 & 0 & 0\\0&1&0}\V T (\V z) \MAT{\V c'_i\\ 1} .
\end{align}
To represent the shape of the object, we utilize superquadric SQ (in 2D called super-ellipses) defined implicitly by the inside-outside function $F_{SQ}(x,y)$ \cite{jaklic2000segmentation}:
\begin{align}
    F_{SQ}(x,y) &= \left(\frac{x}{a_1}\right)^{\frac{2}{\upsilon}} + \left(\frac{y}{a_2}\right)^{\frac{2}{\upsilon}} - 1
    \label{eq:io}
\end{align}
where $a_1$ and $a_2$ determine the scale of the shape and $\upsilon$ controls its curvature. The function $F(x,y)$ is positive outside the surface, zero on the boundary, and negative inside the object. $F_{SQ}$ acts as a differentiable inside-outside function that induces an SDF-like representation for contact modeling.

\begin{figure}[!h]
    \centering
    \includegraphics[width=0.98\columnwidth]{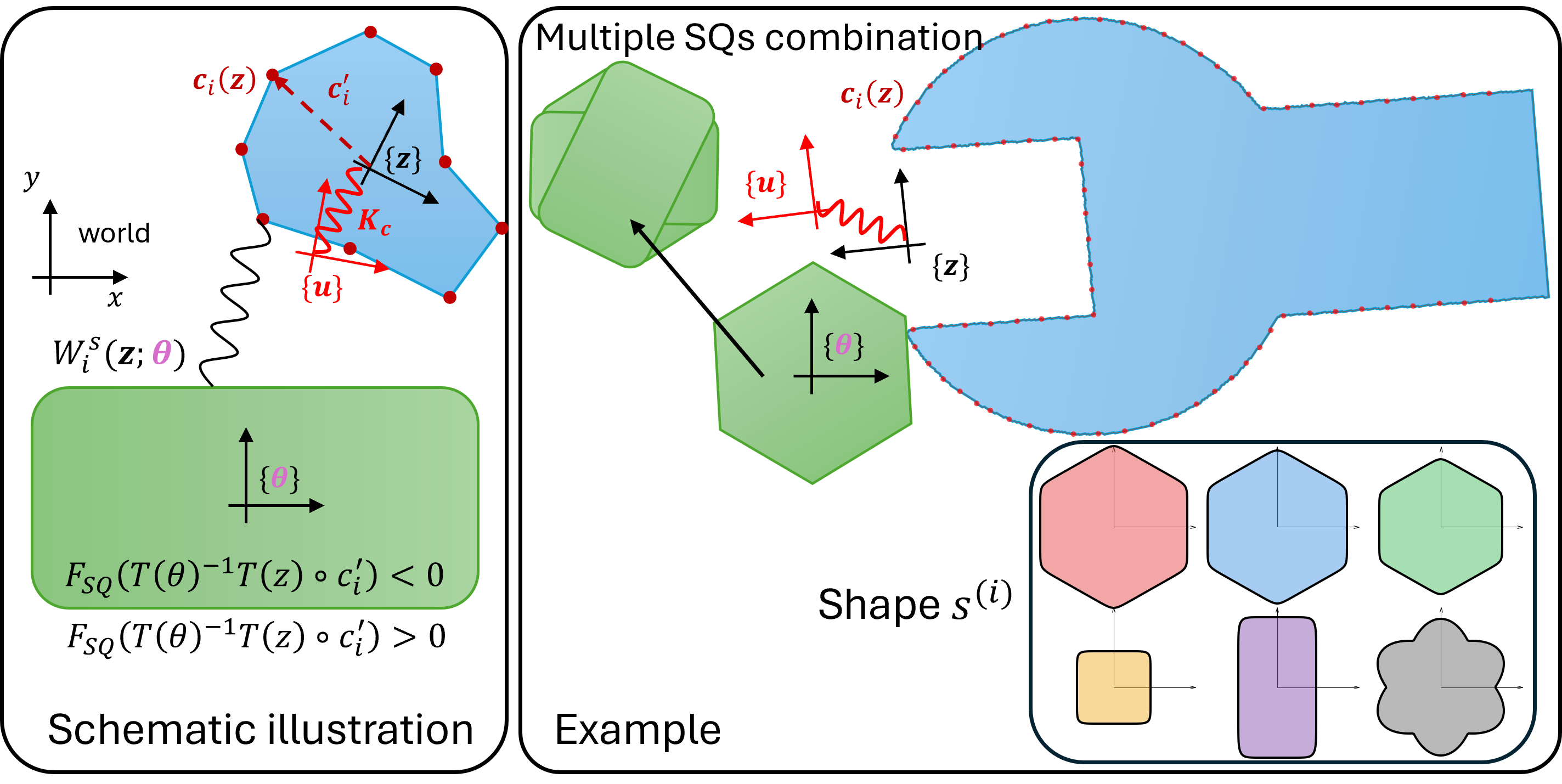}
    \caption{Instantiation of the manipulation potential  $W^{(s)}(\V z,\V u;\V \theta)$ through contact geometry and impedance control. Left: Schematic representation of the general case. Right: Illustrative example of a spanner–screw operation.}
    \label{fig: point_clou}
\end{figure}

\subsubsection{SQ-PC Contact Potential}
Contact between the tool and the object is modeled by evaluating the distance of each tool point $\V c_i(\V z)$ to the object surface. For a given shape hypothesis $s$, the contact potential associated with point $i$ is defined as 
\begin{align}
    W^{(s)}_i(\V z; \V \theta) &= \left[ 1+\exp(-F_i/\zeta_1)\right]^{\zeta_1\zeta_2}
        \label{eq:contact_W}
\end{align}
where $F_i=F_{SQ}[\V T(\V \theta)^{-1} \V T(\V z) \circ \V c_i']$. Eq. \ref{eq:contact_W} yields a $C^\infty$ one-sided approximation of a contact barrier, ensuring continuity of both gradients and Hessians. The parameter $\zeta_1$ controls the smoothing bandwidth of the transition region around the contact boundary ($F_{SQ}=0$), while $\zeta_2$ regulates the effective stiffness of the contact interaction. Compared to a naive exponential potential $W=\exp(-\zeta F)$, the proposed design suppresses spurious long-range gradients.

To safely regulate physical interaction in uncertain environments, we employ impedance control, which introduces compliance and prevents excessive contact forces during exploration. The controller acts as a virtual spring with stiffness $\V K_c$, coupling the actual tool pose $\V z$ to the desired command $\V u$. This representation is illustrated in Fig.~\ref{fig: point_clou}. By incorporating this elastic energy (impedance potential) with the environmental contact potential, the total system manipulation potential becomes:
\begin{equation}
\resizebox{0.8875\linewidth}{!}{$
W^{(s)}(\V z,\V u; \V \theta) = \underbrace{\frac{1}{2}(\V u - \V z)^\top \V K_c (\V u - \V z)}_{\text{Impedance potential}} + \underbrace{\sum_i W^{(s)}_i(\V z; \V \theta)}_{\text{Contact potential}}
$}
\end{equation}

\subsubsection{Regularized Friction as a Forward Correction Term}
\label{subsec:fri}
To capture realistic contact behaviors (e.g., stick-slip transitions) while preserving differentiability and tractable gradients, we incorporate friction as a velocity-regularized, non-conservative correction to the quasi-static update, rather than as part of the equilibrium constraint.

Following the kinematic and force projections detailed in Appendix~\ref{app:friction_derivation}, we derive a generalized friction wrench $\bm{\mathcal{F}}^{fri}_i$ for each point cloud node $\V c_i$ from its predicted tangential slip velocity and normal contact force. Importantly, friction is computed from the frictionless quasi-static update (Eq.~\eqref{eq:ode0}) and used only to modify the forward integration path. The resulting adaptive update of $\mathbf{z}$ is then given by:
\begin{equation}
\resizebox{0.8925\linewidth}{!}{$
    \dot{\mathbf{z}} =
    -\left(\partial^2_{\mathbf{zz}} W\right)^{-1} \! \partial^2_{\mathbf{uz}} W \, \dot{\mathbf{u}}
    - \eta \left(\partial^2_{\mathbf{zz}} W \right)^{-1}
    \! \big( \partial_{\mathbf{z}} W - \sum_i \bm{\mathcal{F}}^{fri}_i \big)
    \label{eq:ode_fri}
$}
\end{equation}
where $\bm{\mathcal{F}}^{fri}_i$ acts as a forward-only dissipative correction. As friction is modeled as a velocity-regularized dissipative correction rather than an equilibrium constraint, it does not contribute to the equilibrium gradient and thus leaves the quasi-static condition $\partial_{\V z} W = 0$ unchanged.

\section{Hybrid Haptic SLAM and planning framework}\label{sec:haptic_SLAM}
This section translates the previously established physical and geometric dual perspective into an operational robotic perception and control architecture. Haptic SLAM is formulated as a manifold parameter estimation problem, where the objective is to jointly infer discrete environmental shape parameters $s$ and continuous state variables (poses) $\V \theta$ by decoding sparse, tool-mediated haptic signals (wrench). By leveraging the parameterized equilibrium manifold $\mathcal{M}_{eq}^{(s, \theta)}$, the framework resolves the under-determined state through sequential haptic observations while concurrently facilitating online planning and adaptive control on the same manifold structure. Finally, the integrated framework is evaluated through numerical validation to verify its effectiveness and convergence properties before experimental deployment.

\subsection{Haptic SLAM Formulation}\label{sec:hybrid}
We formulate the haptic perception problem by bridging the gap between continuous physical interaction and discrete state estimation.

\subsubsection{Continuous Interaction and Discrete Sampling}
Let the interaction evolve over a continuous time horizon $t \in [0, T]$. At any continuous time $t$, the control input $\V u(t)$ induces an equilibrium configuration of robot-tool-environment system with manipulation potential $W^{(s)}$ described in Sec. \ref{sec: equilibrium mfd}. Theoretically, for a given environment hypothesis $(s, \V \theta)$, the system generates an instantaneous haptic mismatch vector $\V \epsilon(t) = \overline{\bm{\mathcal{F}}}(t) - \bm{\mathcal{F}}[\V z^*(t), \V u(t); s, \V \theta]$.

Although physical interactions evolve continuously, perception and inference operate on discrete measurements. We sample the continuous trajectory at global sample index as $m$, yielding a sequence of action-observation pairs (AOP) $\{(\V u_m, \overline{\bm{\mathcal{F}}}_m)\}_{m=1}^{M}$. Under this setting, the objective is to jointly infer the discrete environment shape parameter $s$ and environment poses $\V \theta_{1:M}$, conditioned on the history of controls and measurements. This hybrid formulation falls under the class of SLAM (Simultaneous Localization and Mapping) techniques \cite{thrun2002probabilistic}. From a probabilistic perspective, the full joint posterior distribution follows the standard recursive SLAM form:
\begin{align}\label{eq:joint1}
p(s, \V \theta_{1:M} \mid \overline{\bm{\mathcal{F}}}_{1:M}, \V u_{1:M}&) 
\propto P(s) \, p(\V \theta_1) \nonumber \\
\prod_{m=2}^{M} \! p(\V \theta_m \mid \V \theta_{m-1}, \V u_m) 
\prod_{m=1}^{M} & p(\overline{\bm{\mathcal{F}}}_m \mid \V \theta_m, s, \V u_m).
\end{align}

\subsubsection{Windowed Batch Estimation}
In contrast to classical SLAM formulations, we do not employ an explicit stochastic motion model $p(\V \theta_m \mid \V \theta_{m-1}, \V u_m)$ to propagate the pose over time.
Grounded in the quasi-static assumption, we treat the time-varying pose as a piecewise constant state. By further assuming that the haptic observations within each temporal window are independent and identically distributed (i.i.d.) conditioned on the local pose, we aggregate information to compensate for the sparsity of informative haptic signals, where a single AOP is insufficient to fully resolve the object's pose. Consequently, we partition the interaction sequence into local segments, referred to as batches of $N$ samples:
\begin{equation}
    \mathcal{B}_b = \{m \mid (b-1)N < m \leq bN\}, \qquad b=1, ..., B
\end{equation}
where $B=M/N$ is the number of batches. Within each batch $\mathcal{B}_b$, the measurements are assumed to be generated from a common local pose parameter i.e.
\begin{subequations}
\begin{align}
 p(\V \theta_m \mid \V \theta_{m-1}, \V u_m) &= \delta(\V \theta_m - \V \theta_{m-1}) \\
\V \theta_m &=\V \theta_b \quad \forall m \in \mathcal{B}_b 
\end{align}
\end{subequations}
By marginalizing the intra-batch transitions, Eq. \eqref{eq:joint1} simplifies to a batch-based formulation:
\begin{align}\label{eq:joint_batch}
p(s, &\V \theta_{1:B} \mid \overline{\bm{\mathcal{F}}}_{1:M}, \V u_{1:M}) 
\propto  \nonumber \\
P(s) \, p(\V \theta_1) 
\prod_{b=1}^{B} &\prod_{m \in \mathcal{B}_b}  \!p(\overline{\bm{\mathcal{F}}}_m \mid \V \theta_b, s, \V u_m).
\end{align}
Crucially, the term $ p(\overline{\bm{\mathcal{F}}}_m \mid \V \theta_b, s, \V u_m)$ defines the observation likelihood of the measured wrench $\overline{\bm{\mathcal{F}}}_m$ relative to the control wrench predicted by the manipulation potential gradient, $\bm{\mathcal{F}}_\text{ctrl}  = -\partial_{\V u} W^{(s)}(\V z^*, \V u_m; \V \theta_b)$, thereby encapsulating the haptic mismatch between prediction and observation.
\begin{figure}[h]
\centering
\includegraphics[width=0.5\textwidth]{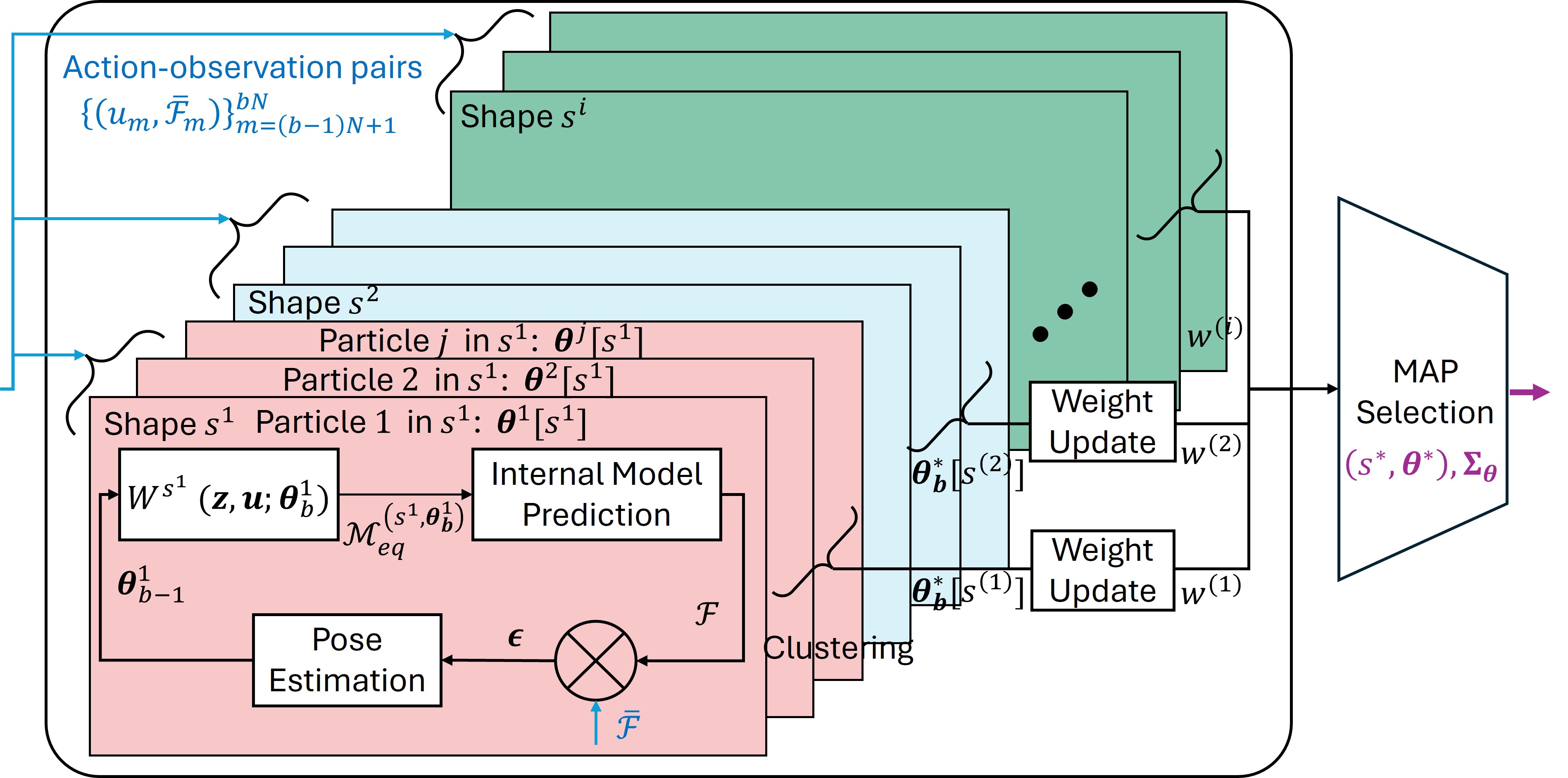}
\caption{Workflow of the hybrid inference in Haptic SLAM. For each discrete shape hypothesis $s^{(i)}$, multiple pose particles $\V \theta^{(j)}[s^{(i)}]$ are initialized to represent candidate configurations. These particles process sensed haptic observations $\overline{\mathcal F}$ in parallel, independently updating their pose parameters $\V \theta^{(j)}[s^{(i)}]$ via a local optimizer driven by the internal physical model. A representative pose $\V{\theta}^*[s^{(i)}]$ is then identified for each shape, followed by a MAP selection and clustering process to determine the estimated object shape $s^*$ and pose $\V{\theta}^*$ for the current batch. This inference cycle is re-executed in each subsequent batch as new haptic observations become available.}
\label{fig:haptic_SLAM_logic}
\end{figure}

\subsubsection{Hybrid Inference Strategy}
Directly evaluating the joint posterior in Eq.~\eqref{eq:joint_batch} is computationally intractable as it couples a discrete global variable $s$ indexing alternative physical hypotheses with continuous batch-wise poses $\{\V \theta_b\}_{b=1}^B$. 
We adopt a Rao-Blackwellized (RB) \cite{thrun2002probabilistic} strategy to factorize the posterior into a discrete marginal distribution over shapes and a continuous conditional density over poses:
\begin{align}\label{eq:RB_chain_correct}
&p(s,\boldsymbol{\theta}_{1:B}\mid \overline{\bm{\mathcal{F}}}_{1:M}, \V u_{1:M})
= \nonumber \\
p(\boldsymbol{\theta}_{1:B}\mid s, & \ \overline{\bm{\mathcal{F}}}_{1:M}, \V u_{1:M})\;
p(s\mid \overline{\bm{\mathcal{F}}}_{1:M}, \V u_{1:M}).
\end{align}
Such factorization enables a dual-layer inference scheme in which discrete shape hypotheses are maintained through particle weights, while the continuous pose component is estimated via deterministic optimization. Compared to purely sampling-based approaches, it significantly reduces sampling variance and improves computational efficiency. By treating the discrete shape as a latent state and the continuous pose as a conditional parameter, we frame the perception problem within a Generalized Expectation-Maximization (GEM) logic \cite{dempster1977maximum}, enabling iterative refinement of the belief over shapes and pose estimates. As illustrated in Fig. \ref{fig:haptic_SLAM_logic}, unlike standard Expectation-Maximization that seeks a single global parameter, our framework operates as a bank of parallel estimators conditioned on each shape hypothesis. The workflow proceeds recursively for each data batch $\mathcal{B}_b$ through the following steps:

\begin{itemize}
\item \textit{Initialization}: At the onset of interaction ($b=1$), we initialize a set of discrete shape particles $\{s^{(i)}\}$, where $i$ indexes the shape particles. For each shape hypothesis, a set of pose particles $\{\V \theta^{(j)}[s^{(i)}]\}$ is initialized, where $j$ indexes the pose particles, based on the wrench-axis analysis from the first contact (detailed in Sec. \ref{sec:wrench_axis}).

\item \textit{M-step (Conditional Pose Maximization)}: For the current batch $\mathcal{B}_b$, the haptic SLAM framework refines the pose parameters $\V \theta_{b}$ conditioned on each discrete shape hypothesis $s^{(i)}$. Since the manipulation potential $W^{(s)}$ is shape-dependent, this step is parallelized across all shape hypotheses. For each shape $s^{(i)}$, the associated pose particles $\{\V \theta^{(j)}[s^{(i)}]\}$ process the sensed haptic observations $\overline{\bm{\mathcal{F}}}_m$ in parallel. They are independently updated using gradient-based optimization driven by the haptic mismatch (Sec.~\ref{sec:pose}). Subsequently, a representative pose estimate $\V \theta_{b}^{*}[s^{(i)}]$ is identified for each shape by clustering the converged particles to capture the dominant mode:
\begin{equation}\label{eq:MAP}
\V \theta_b^*[s^{(i)}] = \arg\max_{\V \theta_b} 
\prod_{m \in \mathcal B_b} \!\!
p(\overline{\mathcal F}_m \mid \V \theta_{b}, s^{(i)}, \V u_m)
\end{equation} 
Instead of performing full RB marginalization over the pose $\V \theta$, the clustering-based MAP estimate resolves symmetry-induced multi-modality and ensures physically consistent solutions.

\item \textit{E-step (Shape Posterior Weighting)}: 
Once the optimal poses $\V \theta_{b}^{*}[s^{(i)}]$ are determined, we update the posterior belief over the discrete shapes. The weight $w^{(i)}$ for each shape particle is updated recursively based on how well its corresponding internal model explains the observed haptic data in the current batch (Sec.~\ref{sec:shape}). 
\begin{equation}\label{eq:weight_update}
w^{(i)} \propto P[s^{(i)}] 
\prod_{b=1}^B \prod_{m \in \mathcal B_b} 
\!\!  p(\overline{\mathcal F}_m \mid \V{\theta}^*_b[s^{(i)}], s^{(i)}, \V u_m)
\end{equation}
The final output is the Maximum A Posteriori (MAP) shape estimate $s^* = \arg\max_{s^{(i)}} w^{(i)}$, which facilitates the closure of the perception-action loop with the most physically consistent environment model.
\end{itemize}

\subsection{Pose Particles Initialization}\label{sec:wrench_axis}
To initialize the recursive haptic SLAM framework, the prior distribution over the initial pose $p(\V{\theta}_1)$ for each shape hypothesis is derived from the first haptic observation, obtained at the onset of interaction.

\subsubsection{Single contact point estimation}
Haptic SLAM is initialized at the instant of first contact, which typically corresponds to a single contact point between the tool and the object. Since it is rare for multiple contact points to occur simultaneously at the very first moment of contact, this single-point assumption is reasonable. Under this condition, the contact location can be inferred from the wrench-axis~\cite{turlapati2022towards}, noting that this formulation is only valid in the single-point contact regime. In our framework, sensed (control) wrench equals
$$\bm{\mathcal{F}}_\text{ctrl} := \begin{bmatrix}\V f  \\ \V \tau \end{bmatrix} =  -\partial_\V{u}W^{(s)}(\V z^*,\V u; \V \theta)$$
where $\overline{\V f}$ and $\overline{\V \tau}$ denote the force and torque measured at the tool handle.
For a single contact point, the wrench-axis admits a reference point at the onset of contact
\begin{equation}\label{eq:wx}
    \mathbf{r}_0 = \frac{\overline{\V{f}} \times \boldsymbol{\overline{ \V \tau}}}{\|\overline{\V f}\|^2+\varepsilon},
\end{equation}
where $\varepsilon$ is a very small positive constant for numerical stability, and Eq.\ref{eq:wx} is parameterized as
\begin{equation}\label{eq:wrench_axis}
 \mathbf{r}(\lambda) = \mathbf{r}_0 + \lambda \overline{\V{f}}, \quad \lambda \in \mathbb{R}
\end{equation}
Intersecting this line with the tool boundary yields a set of candidate contact locations.

\subsubsection{Uniform sampling and validation}

Given the feasible contact locations inferred from the wrench-axis, we generate initial pose hypotheses by uniformly sampling the object boundary.
For each sampled boundary point with position $\V p_c$ and outward normal $\V n_c$, we enforce the geometric constraints
\begin{align}\label{eq:find_normal}
    \V T(\V \phi_i) \circ \V p_c &= \mathbf{r}(\lambda) \nonumber \\
    \V R(\V \phi_i) \V n_c &= \xi \overline{\V{f}}, \quad \xi < 0
\end{align}
which require the sampled boundary point to coincide with a point on the wrench-axis and its surface normal to be anti-aligned with the measured force direction. Together, these constraints define a set of candidate object poses consistent with a single-point contact assumption.

\begin{figure}[!ht]
\centering
\includegraphics[width=0.5\textwidth]{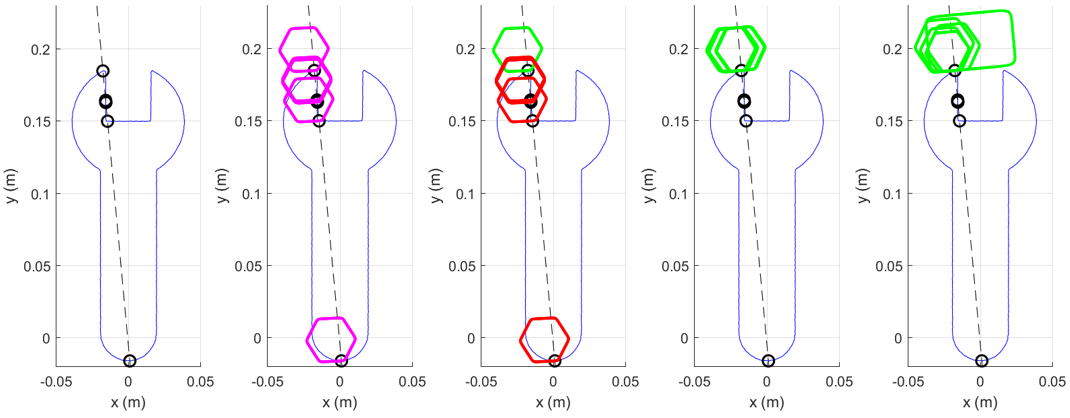}
\caption{Pipeline for pose and particle initialization from single-point contact.}
\label{fig:all priors}
\end{figure}

Figure~\ref{fig:all priors} illustrates the complete initialization pipeline, proceeding from left to right. Starting from the first contact, the measured force and torque define a wrench-axis (Eq.~\ref{eq:wrench_axis}), whose intersections with the tool boundary yield a set of candidate contact locations.
Given each candidate contact, uniform sampling is performed over the object boundary, generating pose hypotheses (magenta) that satisfy the geometry as Eq.~\ref{eq:find_normal}. These hypotheses are then validated by rejecting configurations that result in interpenetration between the tool and the object (red). The remaining candidates (green) correspond to multiple feasible poses for each object shape, denoted by $\V \theta^{(j)}[s^{(i)}]$, and collectively form the initial particle set over both pose and shape hypotheses.

\subsection{Pose Estimation via Least-Squares}\label{sec:pose}
Assuming Gaussian noise on the measured wrench $\overline{\mathcal F}$, the likelihood in Eq.~\eqref{eq:MAP} is equivalent to the least-squares minimization, which serves as the $Q$-function in GEM framework.
\begin{equation}
\V{\theta}_b(s)
= \arg\min_{\V \theta_b}
\sum_{m \in \mathcal B_b}
\|\overline{\mathcal F}_m - {\mathcal F}_m(\V \theta_{b}, s, \V u_m)\|^2_{\Sigma^{-1}},
\label{eq:theta_ls}
\end{equation}
where the initial estimate $\V \theta_1$ is obtained from the wrench-axis initialization (Eq.\ref{eq:find_normal}). $\Sigma$ is defined as measurement noise covariance matrix determined by tool's geometry to balance force and torque components. In practice, we solve this problem incrementally using a Levenberg-Marquardt (LM) scheme.

As discussed in M-step in Sec.~\ref{sec:hybrid}, pose variables are optimized conditionally for each discrete shape hypothesis. In practice, the following pose estimation procedure is applied independently to each pose particle associated with a given shape hypothesis. For clarity, we omit the particle index and present the derivation for a generic pose variable.

Let the reduced manipulation potential be defined as $W^*(\V u,\V \theta) := W[\V z^*(\V u, \V \theta),\V u, \V \theta]$, where $\V z^*(\V u,\V \theta)$ denotes the equilibrium configuration on the corresponding equilibrium manifold. Here, $\V \theta$ is treated as a parameter of the internal model rather than a state variable. The haptic residual at equilibrium is defined as
\begin{align}
\V \epsilon(\V u, \overline{\V \theta},\V \theta)  & =  \overline{\bm{\mathcal{F}}}- \bm{\mathcal{F}}  \\
& = -\partial_{\V{u}}W^*(\V u, \overline{\V \theta}) + \partial_{\V{u}}W^*(\V u,\V \theta) \nonumber \\
& = - \partial_{\V{u}}W[\V z^*(\V u, \overline{\V \theta}),\V u, \overline{\V \theta}] + \partial_{\V{u}}W[\V z^*(\V u, \V \theta),\V u, \V \theta]  \nonumber
\end{align}
where $\overline{\bm{\mathcal F}}$ is the observed wrench and $\bm{\mathcal F} = -\partial_{\V u}W^*(\V u,\V \theta)$ is the predicted wrench generated by the internal model. Defining the per-sample loss as
\begin{equation}
\mathcal{L}(\V u, \overline{\V \theta},\V \theta) := \; \V \epsilon ^T \Sigma^{-1} \V \epsilon
\end{equation}
its gradient with respect to $\V \theta$ is given by
\begin{equation}
    \partial_{\V \theta} \mathcal{L}  = 2 \V J^T \Sigma^{-1} \V \epsilon
\end{equation}
where the Jacobian
\begin{align}\label{eq:gradient}
 \V J( \V u, \V \theta) := \frac{\partial \V \epsilon}{\partial\V \theta}  & =  \frac{\partial}{\partial\V \theta}  \partial_{\V{u}}W[\V z^*(\V u, \V \theta),\V u, \V \theta]
\end{align}
is obtained via implicit differentiation of the equilibrium constraint.
Taking the total differential of the residual yields
\begin{align}\label{eq: total differential2}
\delta \V \epsilon & = \partial^2_{\V{uu}}W \; \delta\V u +  \partial^2_{\V{uz}}W \; \delta \V z +  \partial^2_{\V{u\theta}}W \; \delta\V \theta
\end{align}

Applying the implicit function theorem to the equilibrium condition
$\partial_{\V{z}} W[\V z^*(\V u, \V \theta), \V u, \V \theta] = \V 0$, we differentiate only through the conservative potential. Since non-conservative corrections (e.g., friction) do not define a state-based equilibrium constraint, they are excluded from the analytic Jacobian.
\begin{align}\label{eq:IFT2}
 \V 0 & = \partial^2_{\V{zu}}W \; \delta\V u +  \partial^2_{\V{zz}}W \; \delta\V z +  \partial^2_{\V{z\theta}}W \; \delta\V \theta  \nonumber \\
\delta\V z & = -(\partial^2_{\V{zz}}W)^{-1} (\partial^2_{\V{zu}}W \; \delta\V u+\partial^2_{\V{z\theta}}W \; \delta\V \theta)
\end{align}
Substituting into Eq.~\eqref{eq: total differential2} yields
\begin{align}\label{eq:subs}
\delta \epsilon = &  \,\, [\partial^2_{\V{uu}} W - \partial^2_{\V{uz}} W(\partial^2_{\V{zz}}W)^{-1}\partial^2_{\V{zu}}W]\; \delta\V u  \nonumber \\
& +
[\partial^2_{\V{u\theta}} W - \partial^2_{\V{uz}} W(\partial^2_{\V{zz}}W)^{-1}\partial^2_{\V{z\theta}}W ]\; \delta\V \theta \nonumber \\
\delta \epsilon = & \,\,\V G( \V u, \V \theta) \delta \V u + \V J( \V u, \V \theta) \delta\V \theta
\end{align}
where, at equilibrium, $\delta\V u=0$ and
\begin{align}
 \V J( \V u, \V \theta) & = \partial^2_{\V{u\theta}} W - \partial^2_{\V{uz}} W(\partial^2_{\V{zz}}W)^{-1}\partial^2_{\V{z\theta}}W 
\end{align}
Finally, the pose update is computed using an LM step:
\begin{align}
    \Delta \V \theta = (\V J_b^T  \V \Sigma^{-1} \V J_b+\lambda \V I)^{-1} \V J_b^T \Sigma^{-1} \V \epsilon_b
    \label{eq:GN}
\end{align}
where $\V J_b$ and $\V \epsilon_b$ stack all observations within batch $b$.
\begin{figure}[!ht]
\centering
\includegraphics[width=0.5\textwidth]{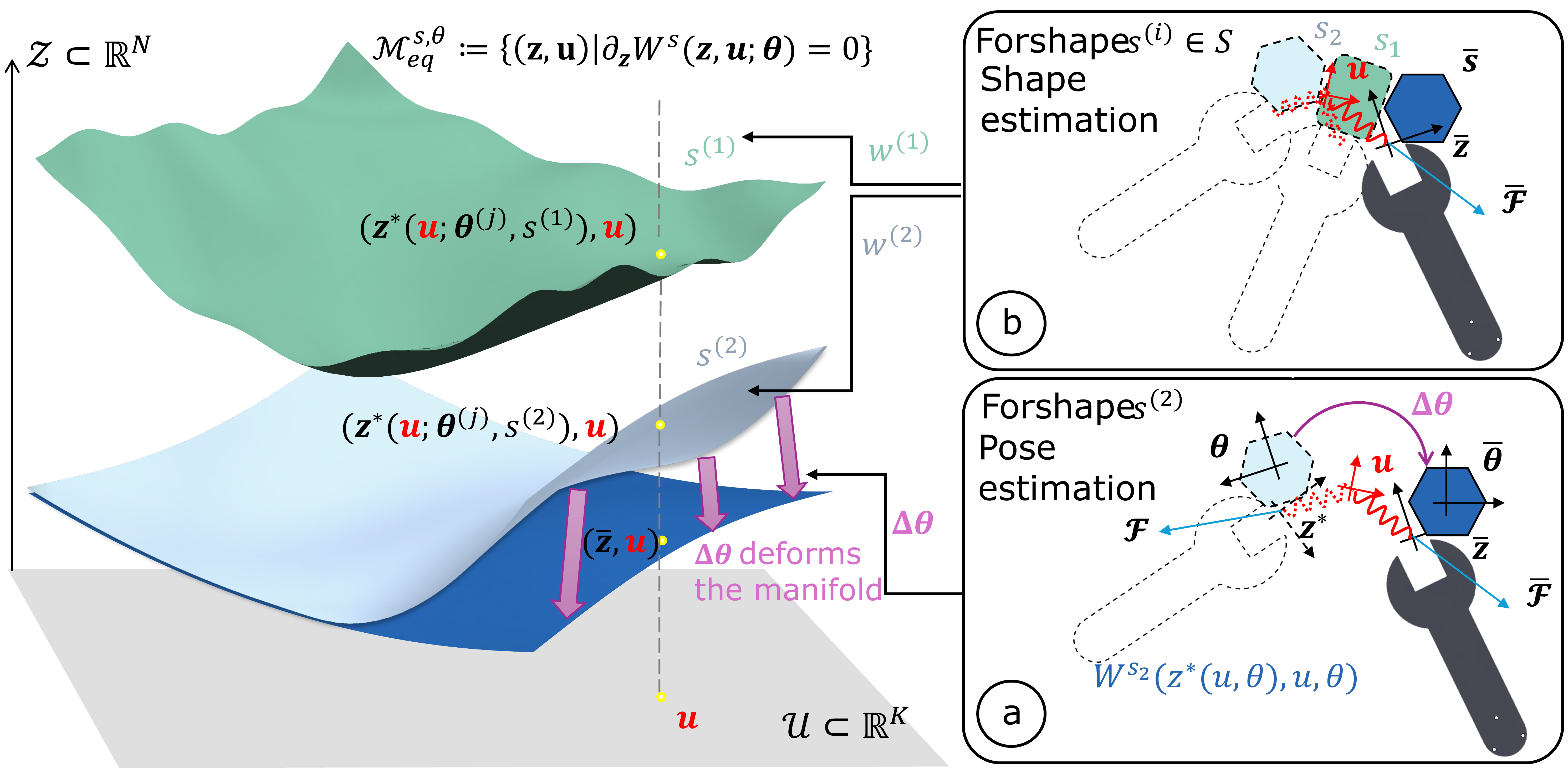}
\caption{%
Physical and geometric interpretation of haptic SLAM: (a) Pose estimation (continuous update): For a fixed shape $s^{(2)}$, haptic mismatch $\V \epsilon =  \overline{\bm{\mathcal{F}}}- \bm{\mathcal{F}}$ drives the deformation $\Delta \V \theta$ to align the expected manifold (light blue) with reality (dark blue). (b) Shape estimation (discrete selection): Parallel manifolds for distinct shapes, such as a green rectangle ($s^{(1)}$) and a blue hexagon ($s^{(2)}$), are compared to select the candidate based on  $\V \epsilon$.
}
\label{fig:fig_SLAM_v2}
\end{figure}

Figure~\ref{fig:fig_SLAM_v2} (a) provides an intuitive interpretation of the pose update in haptic SLAM. For a given control input $\V u$, the real system and the internal model generally settle at different equilibrium configurations, denoted by $\overline{\V z}$ and $\V z$, producing the measured and predicted wrenches $\overline{\mathcal F}$ and $\mathcal F$, respectively. The resulting haptic mismatch drives the parameter update $\Delta\V \theta$ through Eq.~\eqref{eq:GN} as indicated by the pink arrow, deforming the equilibrium manifold such that the predicted equilibrium configuration (light blue) moves closer to the true one (dark blue).

\subsection{Shape Estimation via Particle Weights}\label{sec:shape}
The posterior over the discrete shape parameter is represented using a finite set of particles $\{s^{(i)}, w^{(i)}\}$, where each particle corresponds to a candidate shape hypothesis. Conditioned on a given $s^{(i)}$, multiple pose particles are refined in parallel as described in Sec.~\ref{sec:pose}. These refined pose particles are subsequently summarized by a representative pose estimate $\V\theta_b^*[s^{(i)}]$ through greedy distance-based clustering \cite{thrun2002probabilistic}, which groups geometrically equivalent configurations.

The particle weights approximate the posterior distribution over shape hypotheses. They are initialized according to the prior
\begin{equation}
w^{(i)}_0 = P(s^{(i)}).
\end{equation}
and updated by evaluating the observation likelihood at the optimized pose estimates:
\begin{equation}
w^{(i)} \propto w^{(i)}_0
\prod_{b=1}^B \prod_{m \in \mathcal B_b}
p\!\left(\overline{\mathcal F}_m \,\middle|\,
\V\theta_b^*[s^{(i)}],\, s^{(i)},\, \V u_m\right).
\end{equation}
The weights are then normalized as
\begin{equation}
\tilde{w}^{(i)} = \frac{w^{(i)}}{\sum_j w^{(j)}},
\label{eq:weight}
\end{equation}
yielding a discrete posterior distribution over shape hypotheses. The maximum a posteriori (MAP) particle identifies the most probable shape hypothesis $s^*$ at the current batch, which is subsequently used for downstream planning. Fig. \ref{fig:fig_SLAM_v2}(b) provides a geometric interpretation of this update. Each shape hypothesis $s^{(i)}$ induces its own equilibrium manifold and a corresponding posterior weight $w^{(i)}$. In essence, this process identifies the discrete geometry whose induced manifold generates the expected wrench that best aligns with the observations.

\subsection{Haptic planning and control}\label{sec:planning_control}
\subsubsection{Manipulation planning}
Dynamic Motion Primitives (DMPs) provide a compact parameterization of smooth control trajectories via a finite set of weights \cite{ijspeert2013dynamical}. Previous work has shown that DMPs can be combined with black-box optimization (BBO) to search for an optimal trajectory on the equilibrium manifold for contact-rich tasks \cite{yang2025planning_book}. However, such optimization is typically performed offline and is tied to a fixed target condition. Here we instead adopt an online receding-horizon formulation inspired by Model Predictive Path Integral (MPPI) control~\cite{williams2017model}. Following this line of work, we formulate haptic planning as a receding-horizon planning problem, where a short-horizon control trajectory is repeatedly re-planned and executed over short segments.

In our formulation, the control trajectory $\V u(t)\in\mathbb{R}^K$ is parameterized by a DMP with parameters $\V\beta$, mapping a finite-dimensional parameter $\V\beta$ to a smooth time-varying control trajectory $\V u(t)$ subject to boundary conditions $\V u(0)=\V u_0$ and $\V u(T)=\V u_1$. The initial control $\V u_0$ is obtained from the robot state, while the target $\V u_1(s^*,\V\theta^*)$ is determined from the current haptic SLAM estimation.

Coupling the DMP formulation with the adaptive ODE (Eq.\ref{eq:ode_fri}) yields the following haptic DMP dynamics:
\begin{align}
\tau\dot{\V u} = & \ \V v, \nonumber \\
\tau\dot{\V v} = & \ \alpha_1 [\alpha_2(\V u_T - \V u) - \V v] + \V f(\V \beta), \nonumber  \\
\Vdot z = -(\partial^2_\V{zz}W)^{-1}\partial^2_\V{uz}W \Vdot u &- \eta (\partial^2_\V{zz}W)^{-1} \big(\partial_\V z W  - \sum_i \bm{\mathcal{F}}^{fri}_i \big), \nonumber \\
\dot{\psi} = & \ \sqrt{\V v^T \V G^2(\V u) \,\V v}\,. \label{eq:allODE}
\end{align}

While the haptic distance $\Gamma[\V u]$ is defined as a path functional as Eq.\ref{eq:haptic_dis}, the auxiliary state $\psi(t)$ accumulates the same quantity along a time-parameterized realization of the path. Eq. \ref{eq:allODE} defines the forward rollout dynamics used in MPPI. Multiple rollouts are generated by sampling perturbed DMP parameters $\V\beta^r\sim\mathcal N(\V\beta,\V \Lambda)$, where $\V \Lambda$ is the covariance matrix of the exploration noise. For each rollout, the cost function is designed as the accumulated haptic cost $\psi$, together with a terminal penalty:
\begin{equation}
C_r = \psi(\V \beta^r) + \|\V z_T(\V \beta^r)-\V z_{\mathrm{tgt}}\|_{\Sigma^{-1}}^2.
\label{eq:cost}
\end{equation}
where both $\psi(\V \beta^r)$ and the terminal tool state $\V z_T(\V \beta^r)$ are obtained by integrating Eq. \ref{eq:allODE}. 
The term $\V z_{\mathrm{tgt}}$ is set to control target $\V u_1$.
After evaluating all rollouts, they are ranked by cost and a fixed percentage of the lowest-cost rollouts is selected as an elite set. The DMP parameters are updated by averaging over this elite set, and the resulting control policy is executed over one receding horizon before re-planning.

\begin{figure}[!h]
\centering
\includegraphics[width=0.5\textwidth]{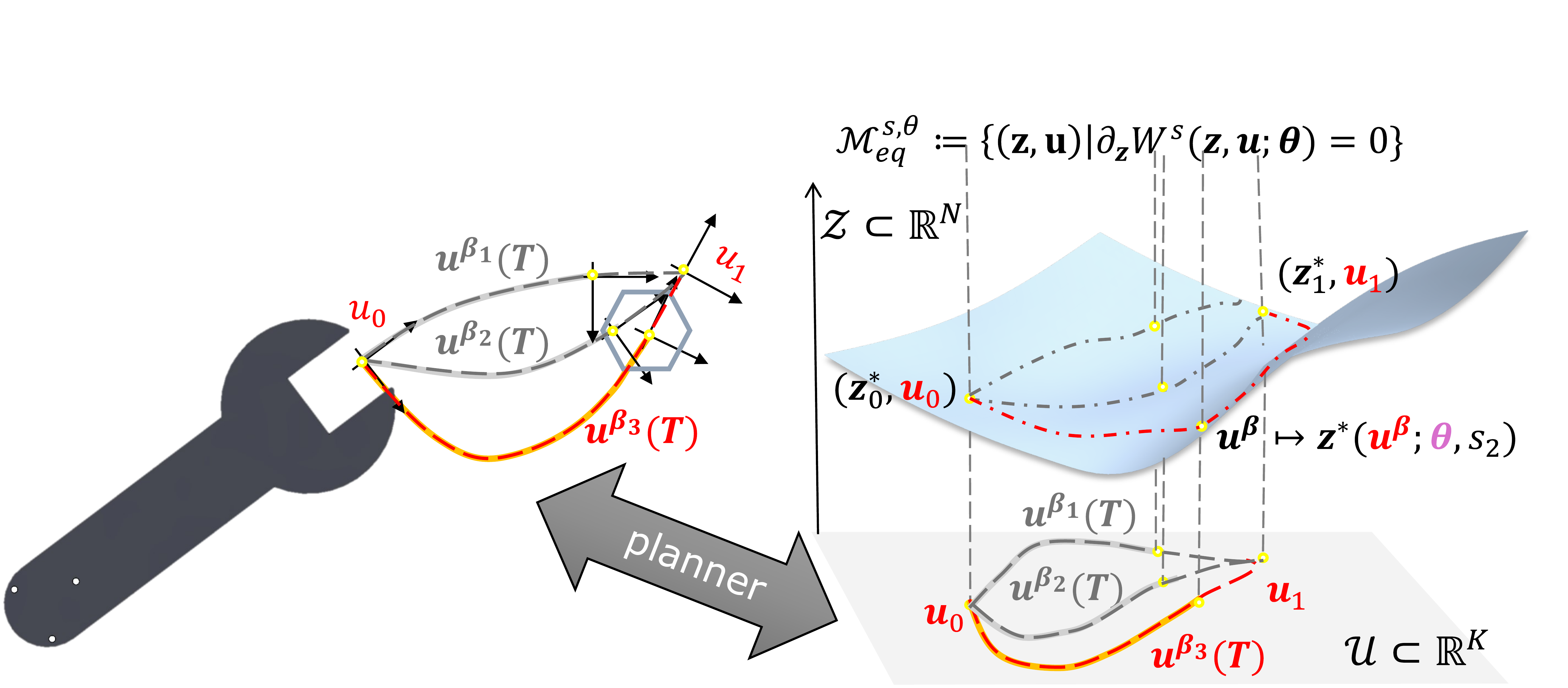}
\caption{%
Schematic illustration of MPPI-based planning. Rollouts are evaluated using a haptic cost and an elite set is used to update the policy in a receding-horizon manner. An experimental replotted execution of this process is shown in Fig.~\ref{fig:MPPI_exp}.}
\label{fig:plan}
\end{figure}

Fig. \ref{fig:plan} illustrates the conceptual workflow of MPPI-based haptic planning. At each iteration, multiple control trajectories are sampled and forward simulated along the estimated equilibrium manifold. Their accumulated haptic costs are evaluated, and a fixed percentage of the lowest-cost rollouts is selected as an elite set to update the control policy. Only a short segment of the updated policy is executed before state estimation is refreshed and replanning is performed.

The target configuration and terminal state are task-dependent and may admit multiple valid choices due to object symmetry. A replotted execution of this planning process using real experimental data is provided later in Fig.~\ref{fig:MPPI_exp}.

\subsubsection{Variance informed impedance control}
Haptic SLAM provides pose estimates for each particle together with an associated uncertainty, which can be summarized by a pose covariance matrix $\V \Sigma_{\V \theta}$. In impedance control, stiffness can be modulated inversely with the variance (e.g., $\V K_c \propto \V \Sigma_{\V \theta}^{-1}$)  to allow sufficient compliance in the presence of inaccuracy \cite{calinon2010learning}. However, in insertion tasks such as spanner-screw assembly, it is additionally necessary to ensure adequate rigidity along the insertion direction, thus isotropic stiffness \cite{Takagi2020} . Therefore, we define a stiffness anisotropy ratio $\kappa := k_n / k_\phi$, which controls the relative stiffness between translation along the insertion axis and rotation about it. This ratio is adapted based on pose uncertainty as
\begin{equation}\label{eq: stiffness}
\kappa = \kappa_{min} + \frac{1-\tanh [\text{Tr}(\V \Sigma_{\V \theta})/\sigma_0]}{2} (\kappa_{max}-\kappa_{min}).
\end{equation}
where $\kappa_{min}$ and $\kappa_{max}$ bound the admissible range.

Here, $k_n$ and $k_\phi$ represent the translational stiffness along the insertion axis (normal) and the rotational stiffness, respectively. This formulation is grounded in the mechanics of the Remote Center of Compliance (RCC) \cite{whitney1982quasi}. The resulting task-space stiffness matrix is defined as $\V K_{RCC} := \mathrm{diag}(k_n, k_t, k_\phi)$ and is mapped to the robot control frame via
\begin{equation}
\V K_{C} = [\mathrm{Ad}_{\V T(\V z)}]^{-T} \V K_{RCC}  \, [\mathrm{Ad}_{\V T(\V z)}]^{-1}
\label{eq:RCC}
\end{equation}
where $[\mathrm{Ad}_{\V T(\V z)}]$ denotes the adjoint of the transformation $\V T(\V z)$.

The adaptive modulation of $\kappa$ serves a dual purpose. When pose uncertainty is high, $\kappa$ is reduced, resulting in lower normal stiffness relative to rotational stiffness. This encourages compliant exploration, reduces contact forces, and mitigates jamming. As the pose estimate converges and uncertainty decreases, $\kappa$ increases, stiffening the insertion axis while preserving rotational compliance, thereby providing sufficient force for mating while maintaining alignment robustness.

\subsection{Numerical Validation}\label{sec:simu_analysis}
We evaluate the proposed method in simulation on the tool insertion stage of a screw-loosening task. This study aims to validate the convergence behavior of the hybrid inference framework and the online planner before real-world experiments.

\subsubsection{Simulation Setup} The simulation environment is constructed using the geometric models presented in Sec.~\ref{sec:contact_model}, following the \textit{Shape-Size} naming convention (where the suffix indicates the characteristic size in millimeters). To simulate the complete closed-loop workflow (Fig.~\ref{fig:workflow}), the SQ-PC contact model with the manipulation potential serves as the interaction simulator. For each trial, a ground-truth screw shape and a randomly sampled pose define the environment configuration used to generate the \textit{simulated} contact wrench observations. These contact wrenches are treated as the measurements $\overline{\bm{\mathcal{F}}}$ for the haptic SLAM approach. We instantiate a spanner-30 tool interacting with three representative screw geometries: {Rec-30} (matched), {Hex-30} (matched), and {Hex-36} (oversized). These scenarios, corresponding to the discrete hypotheses ${s^{(1)}, s^{(2)}, s^{(3)}}$, are designed to evaluate the system's ability to handle both correct size matching and geometric incompatibility.

\subsubsection{Shape Classification and Particle Evolution} 
\begin{figure}[!htb]
    \centering
    \includegraphics[width=0.98\columnwidth]{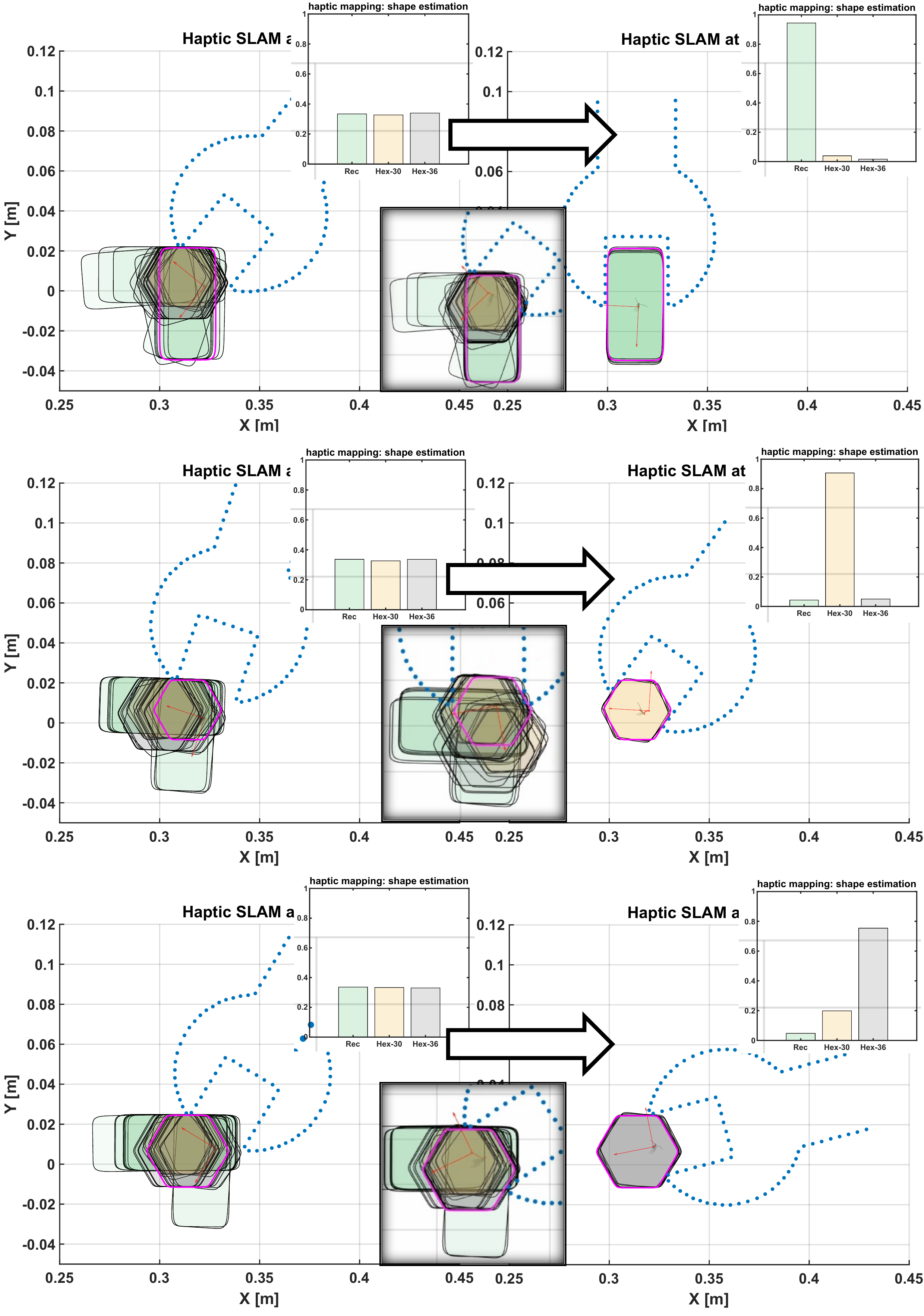}
    \caption{Particle evolution in simulation during the tool insertion stage for three screw configurations.
    From left to right: particle initialization, intermediate evolution under tool-screw interaction, and final converged belief.
    The ground-truth screw geometry is shown in magenta.}
    \label{fig:resu_simu}
\end{figure}

Fig.~\ref{fig:resu_simu} illustrates the evolution of particles during the interaction. The magenta contours denote the ground-truth screw shape $\overline{s}$ and pose $\overline{\mathbf{\theta}}$. To distinguish between hypotheses, the three candidate screw shapes are represented by distinct colors in both the spatial particle distribution and the probability weight plots. Additionally, the robot control input $\mathbf{u}$ is visualized as a red coordinate frame, while the tool geometry is depicted as a blue point cloud.
\begin{itemize}
\item Initialization (Left Column): Particles are initialized according to Eq.~\ref{eq:find_normal}. At this stage, the system exhibits large uncertainty regarding both the discrete screw type and the relative pose, covering multiple hypotheses.
\item Evolution (Middle Column): As the tool interacts with the screw head under commands from the planner, the accumulation of haptic observations progressively updates the particle poses and weights.
\item Convergence (Right Column): The final distribution shows that the estimator successfully converges to the correct shape and pose. In the matched cases (Rec-30, Hex-30), the planner successfully inserts the tool into the screw while the particle weight associated with the correct screw type approaches one. In the oversized case (Hex-36), the planner stops in front of the screw and correctly identifies the geometry.
\end{itemize}

\subsubsection{Pose Convergence Analysis}
\begin{figure}[!h]
    \centering
    \includegraphics[width=0.8\columnwidth]{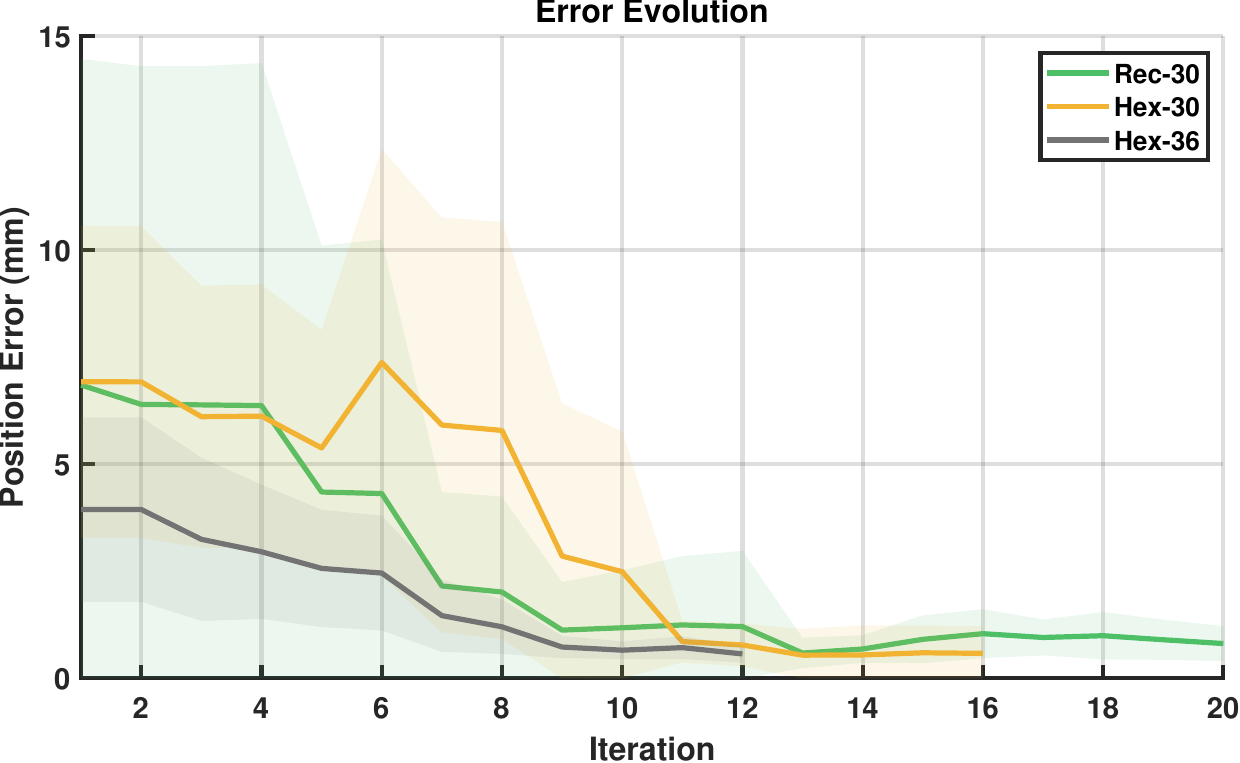}
    \caption{Position error evolution of the estimated screw pose during the insertion stage in simulation.}
    \label{fig:simu_error}
\end{figure}

We further quantify the estimation accuracy by analyzing the convergence of the screw pose over time. Fig.~\ref{fig:simu_error} reports the mean and variance of the position error, computed using particles whose weights exceed a predefined confidence threshold. As shown, both the mean error and variance decrease rapidly as observations accumulate. The final position error converges to below $1,\mathrm{mm}$, where the residual error primarily arises from numerical effects in the contact model.

\section{Experimental Evaluation}\label{sec:experiments}
We conduct a series of experiments to evaluate the proposed closed-loop framework on a screw-loosening task, encompassing both insertion and rotational loosening phases. In total, we conduct 260 real-world trials to comprehensively validate the framework. Specifically, 120 trials across four three-hypothesis study scenarios are designed to demonstrate robustness under typical conditions (Sec. \ref{sec:3class}) , while 60 trials in a challenging six-hypothesis stress test are implemented to stress-test the framework and probe its performance boundaries under structural ambiguity (Sec. \ref{sec:6class}). Finally, 80 trials are dedicated to ablation studies (Sec. \ref{sec:abl}) to analyze the individual contributions of key system components, specifically the adaptive stiffness and perception modules.

\begin{figure}[H]
    \centering
    \includegraphics[width=0.49\textwidth]{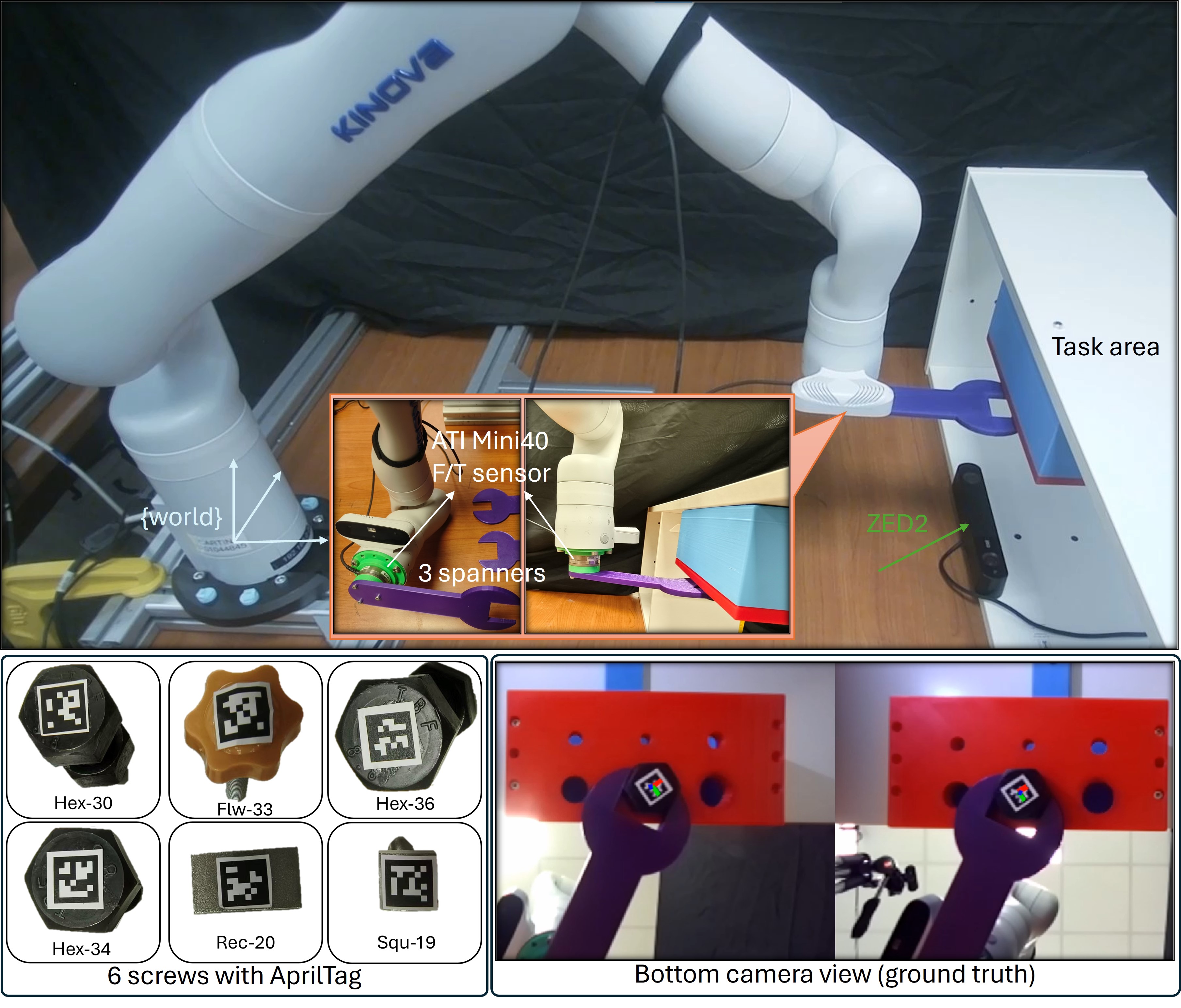}
    \caption{Experimental setup for the screw-loosening task.}
    \label{fig:setup}
\end{figure}
\vspace{-5mm}
\subsection{Experimental Setup}
\label{sec:exp_setup}
All experiments are conducted in task space using a Kinova Gen3 robotic manipulator equipped with an ATI Mini40 force-torque sensor (FT sensor) as shown in Fig.\,\ref{fig:setup}. The system is controlled from a workstation with an Intel i9-14900K CPU and 64 GB RAM. The system is implemented in a dual-layer structure: the high-level inference and planning algorithms utilize the MATLAB Parallel Computing Toolbox for parallelized particle updates, while the low-level impedance controller runs via a C++ interface at approximately $1\,\mathrm{kHz}$ to ensure stable interaction. In our implementation, each discrete shape hypothesis $s^{(i)}$ is sampled by 10 pose particles. The batch length $N$ is set to 20 for the tool insertion stage and reduced to 5 for the rotational loosening phase to accommodate the differing interaction dynamics. For the online planning, 20 perturbed rollouts are generated per iteration.

To enable systematic evaluation and ensure ease of attachment with the robot, we 3D printed three custom open-ended spanners: spanner-36, spanner-34, and spanner-21, where the suffix indicates the nominal aperture size in millimeters. These physical tools correspond to the point-cloud models defined in Sec. \ref{sec:contact_model}. The interaction targets include six distinct screw geometries: Hex-36, Hex-34, Hex-30, Squ-19, Rec-20, and Flw-33. To introduce positional variability, a custom experimental fixture containing six mounting sockets is positioned within the robot's workspace. For each trial, screws are randomly assigned to these sockets.

To obtain ground truth pose measurements for quantitative evaluation, an AprilTag is rigidly attached to the center of each screw. An external ZED2 camera, with extrinsic parameters calibrated to the robot base frame, tracks these tags. Crucially, this vision-based pose data is used exclusively for error analysis and metric computation; it is not provided to the haptic estimation system, ensuring the experiments strictly validate the robot's ability to operate under visual occlusion. Unless otherwise specified, all poses are expressed in robot base frame.

\begin{figure}[!h]
    \centering
    \subfloat[Representative samples of $s^{(1)}$: Hex-36 (Oversized)]{
        \includegraphics[width=0.9\columnwidth]{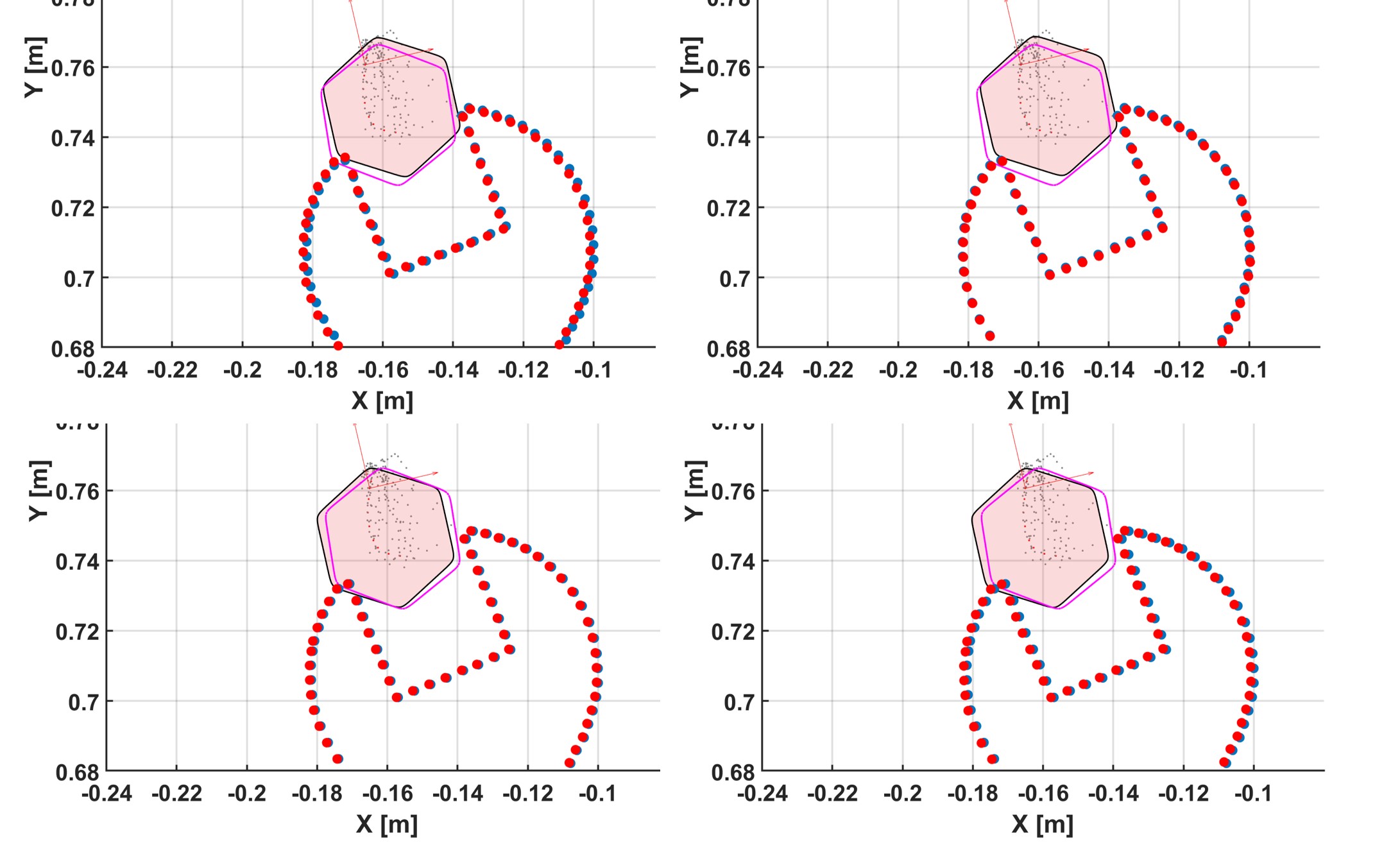}%
        \label{fig:mismatch_sub_a} 
    } \\ 
    \vspace{1pt} 
    \subfloat[Representative samples of $s^{(2)}$: Hex-34 (Matched)]{
        \includegraphics[width=0.9\columnwidth]{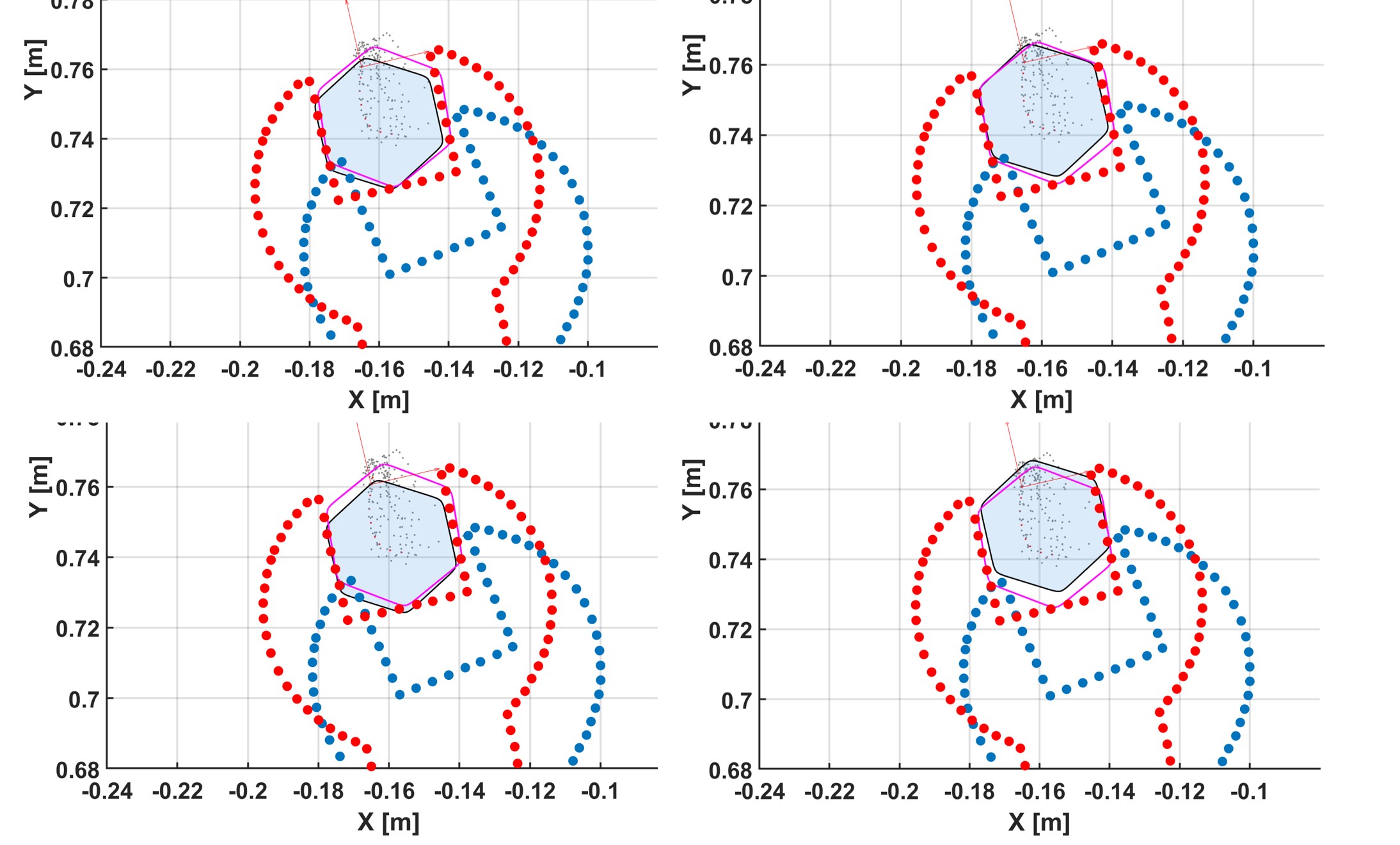}%
        \label{fig:mismatch_sub_b}
    } \\ 
    \vspace{1pt}
    \subfloat[Representative samples of $s^{(3)}$: Squ-19 (Undersized)]{
        \includegraphics[width=0.9\columnwidth]{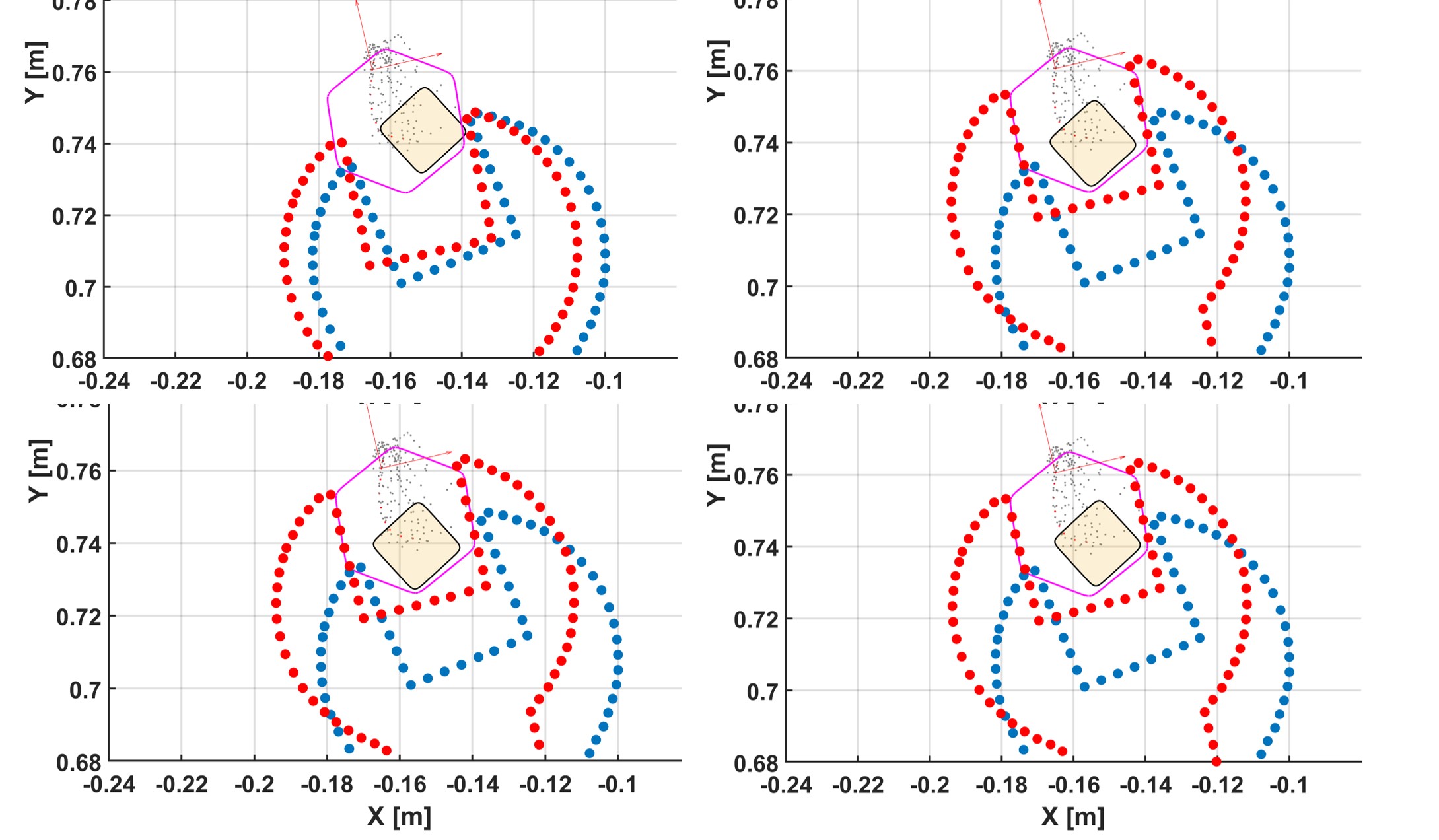}%
        \label{fig:mismatch_sub_c}
    }
    \caption{Visualization of haptic mismatch for discrete hypotheses $s^{(1)}$-$s^{(3)}$. Red and blue point clouds represent the predicted equilibrium $\V z^*$ and observed pose $\overline{\V z}$, respectively, which corresponds to haptic mismatch due to impedance control.}
    \label{fig:haptic_mismatch}
\end{figure}

\subsection{Three-Hypothesis Study }\label{sec:3class}
Following the simulation studies, the real world evaluation begins with three-hypothesis scenarios using various tool-screw combinations. For each manipulation task, the agent is initialized with a hypothesis space $\mathcal{S} = \{s^{(1)}, s^{(2)}, s^{(3)}\}$ with three candidates. Each discrete index $s^{(i)}$ corresponds to a unique tool-screw pairing that defines the specific manipulation potential $W^{(s_i)}$ and its induced equilibrium manifold. The primary objective is to verify whether the Haptic SLAM module can accurately identify the correct physical configuration and estimate its pose across varying geometries, categorized as {oversized, matched, undersized}, while also validating the capability of the planner and controller. To systematically analyze the performance of the proposed framework, we present key analyses as described below.

\subsubsection{Haptic Mismatch Analysis}
To analyze the underlying mechanics of the proposed framework, we visualize the haptic mismatch at a representative snapshot during convergence. As shown in Fig.~\ref{fig:haptic_mismatch}, we present one AOP for four representative particles per hypothesis. The magenta geometry indicates the ground-truth screw, while the red frame denotes the control input $\mathbf{u}_m$. For each shape hypothesis, multiple particles run the haptic SLAM process in parallel.

From the figure, it can be observed that due to the inherent geometric constraints, the tool (spanner-34) becomes obstructed when interacting with an oversized screw, such as the Hex-36 (denoted as $s^{(1)}$ in this trial). Consequently, for particles associated with hypothesis $s^{(1)}$, the expected tool pose $\V z^*(\V u;\V \theta, s)$ (represented by the red point cloud) aligns closely with the measured tool pose (blue point cloud),  as shown in Fig.\ref{fig:mismatch_sub_a}. At this stage, several particles have already converged, resulting in small pose variation. In contrast, when the haptic SLAM hypothesizes a screw size smaller than or matches the physical tool (e.g., Hex-34 $s^{(2)}$ in Fig.\ref{fig:mismatch_sub_b} or Squ-19 $s^{(3)}$ in Fig.\ref{fig:mismatch_sub_c}), the robot expects a successful insertion and thus predicts a minimal interaction wrench $\bm{\mathcal{F}}$. However, as the tool is physically blocked in reality, a large wrench $\overline{\bm{\mathcal{F}}}$ is sensed. Under the system's impedance control law, this force discrepancy is manifested as a pronounced spatial separation between the red and blue point clouds. The geometric distance between the predicted equilibrium state $\mathbf{z}^*$ and the measured state $\overline{\mathbf{z}}$ reflects the magnitude of the haptic mismatch driving the inference, offering a more intuitive interpretation than representing the mismatch directly in wrench space.

\subsubsection{Manipulation planning}
\begin{figure}[!h]
\centering
\includegraphics[width=0.5\textwidth]{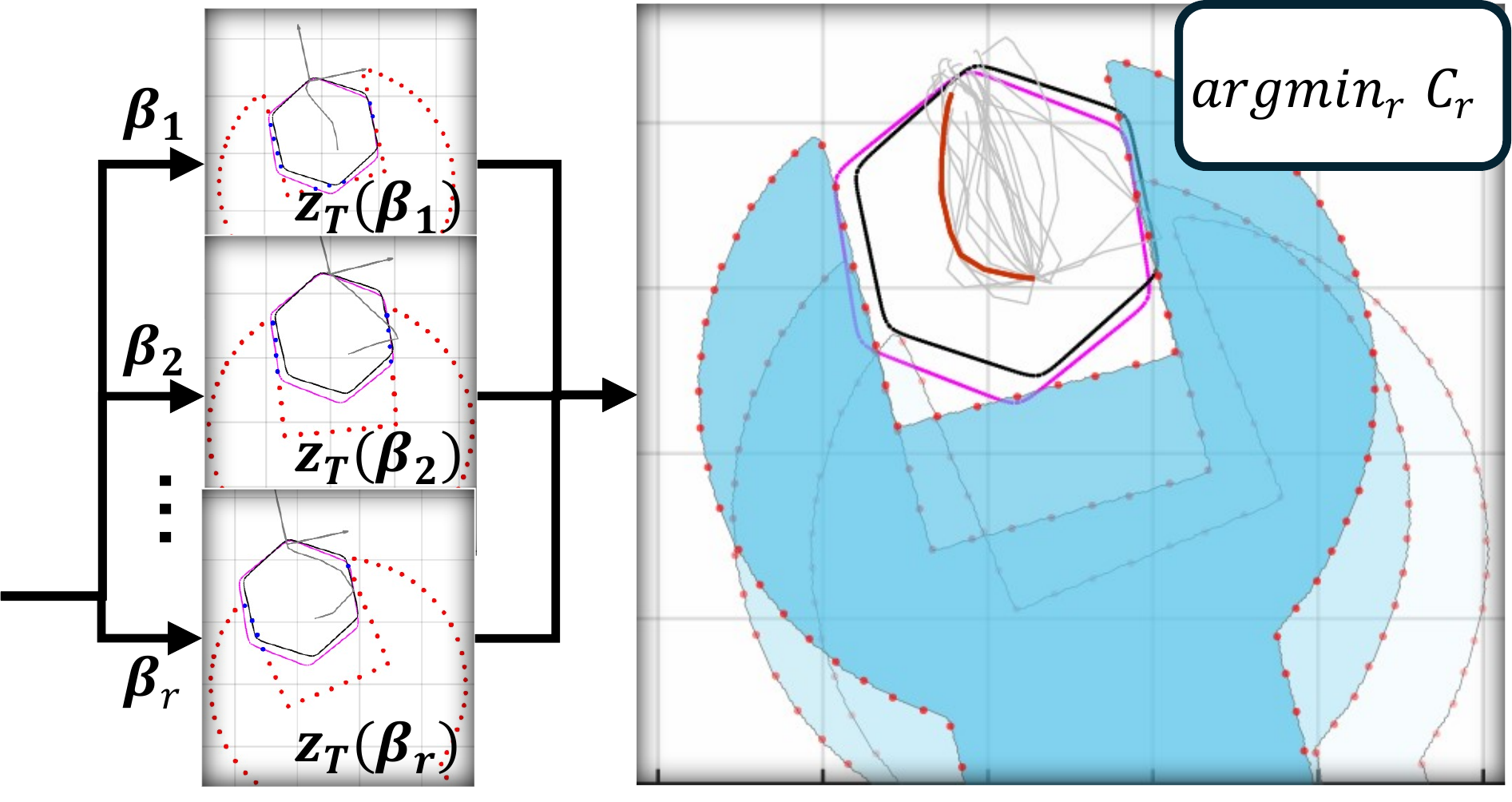}
\caption{Visualization of the online planning process under model mismatch. Gray curves denote sampled trajectories and the red curve indicates the selected minimum cost trajectory. The black contour shows current estimated screw geometry, while magenta contour represents ground truth. The discrepancy between them drives exploratory interaction for state disambiguation.}
\label{fig:MPPI_exp}
\end{figure}

Figure \ref{fig:MPPI_exp} illustrates the online planning process, where multiple control trajectories are generated by sampling perturbations of the DMP parameters $\V \beta^r$ to explore the feasible policy. While some rollouts may result in mechanical jamming or excessive interaction effort, the optimal control policy (highlighted in red) is selected as the elite candidate for execution in the current iteration. The right panel of the figure provides a comprehensive view of all simulated rollouts and the resulting receding-horizon policy.

An interesting aspect of the proposed closed-loop framework is the emergence of a planning gap caused by uncertainty in the estimated shape hypothesis. As shown in the figure, the ground-truth object is an oversized Hex-36 screw (magenta), which is physically incompatible with the tool dimensions (spanner-34). However, the planner operates based on the current haptic state estimate $s^*$, which hypothesizes a Hex-34 screw (black) that permits successful insertion. Consequently, the planner try to insert with $s^*$ and the robot executes the planned trajectory $\V u(t)$ and eventually becomes obstructed at the contact boundary. This deliberate interaction creates the haptic mismatch illustrated in Fig. \ref{fig:haptic_mismatch}, providing the essential sensory feedback, necessary to refine the object hypothesis and ultimately identify the correct screw type.

\begin{figure}[!h]
    \centering
    \subfloat[Particles of $s^{(1)}$: Hex-30 (Oversized)]{
        \includegraphics[width=0.95\columnwidth]{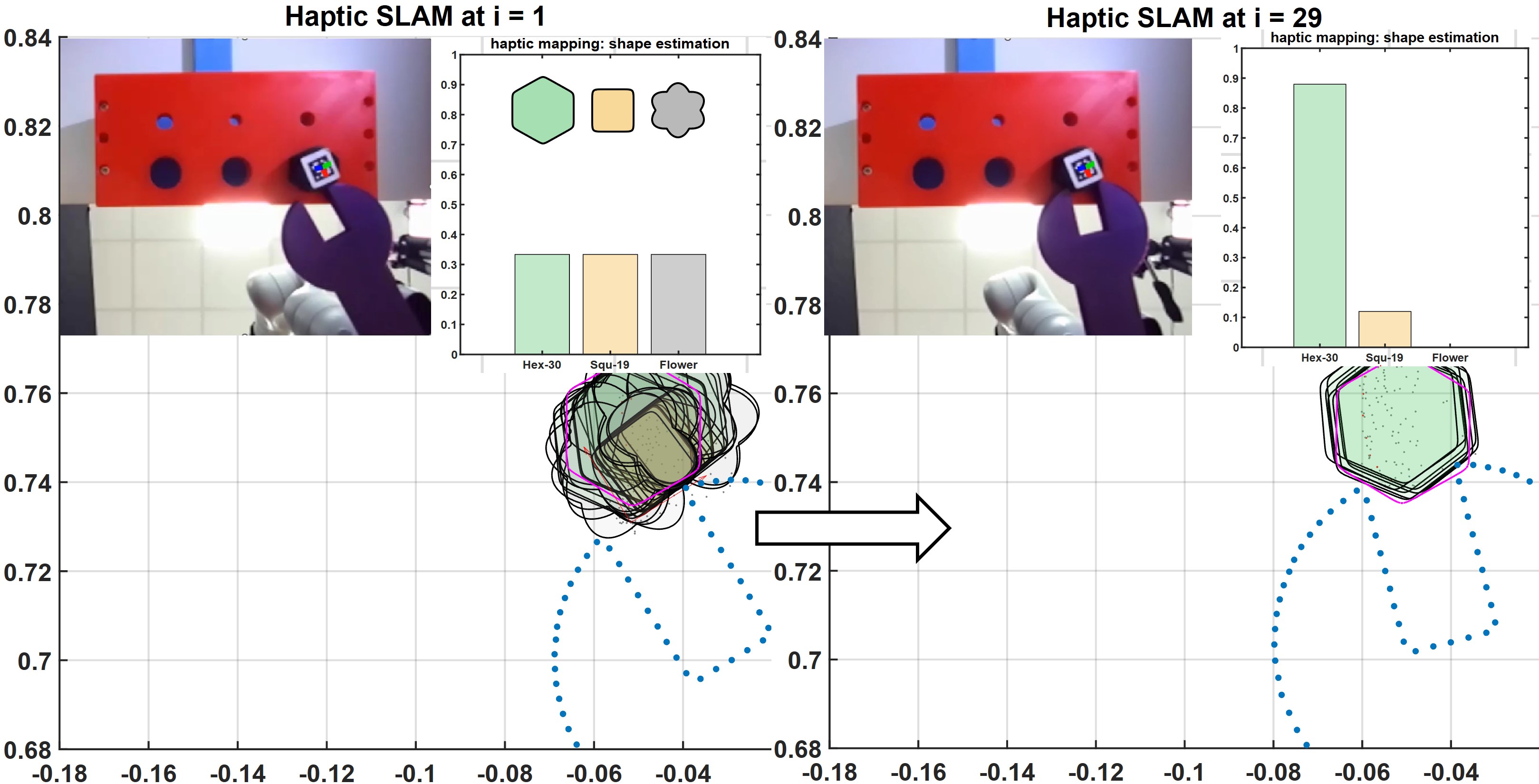}%
        \label{fig:346_3} 
    } \\ 
    \vspace{1pt} 
    \subfloat[Particles of $s^{(2)}$: Squ-19 (Matched)]{
        \includegraphics[width=0.95\columnwidth]{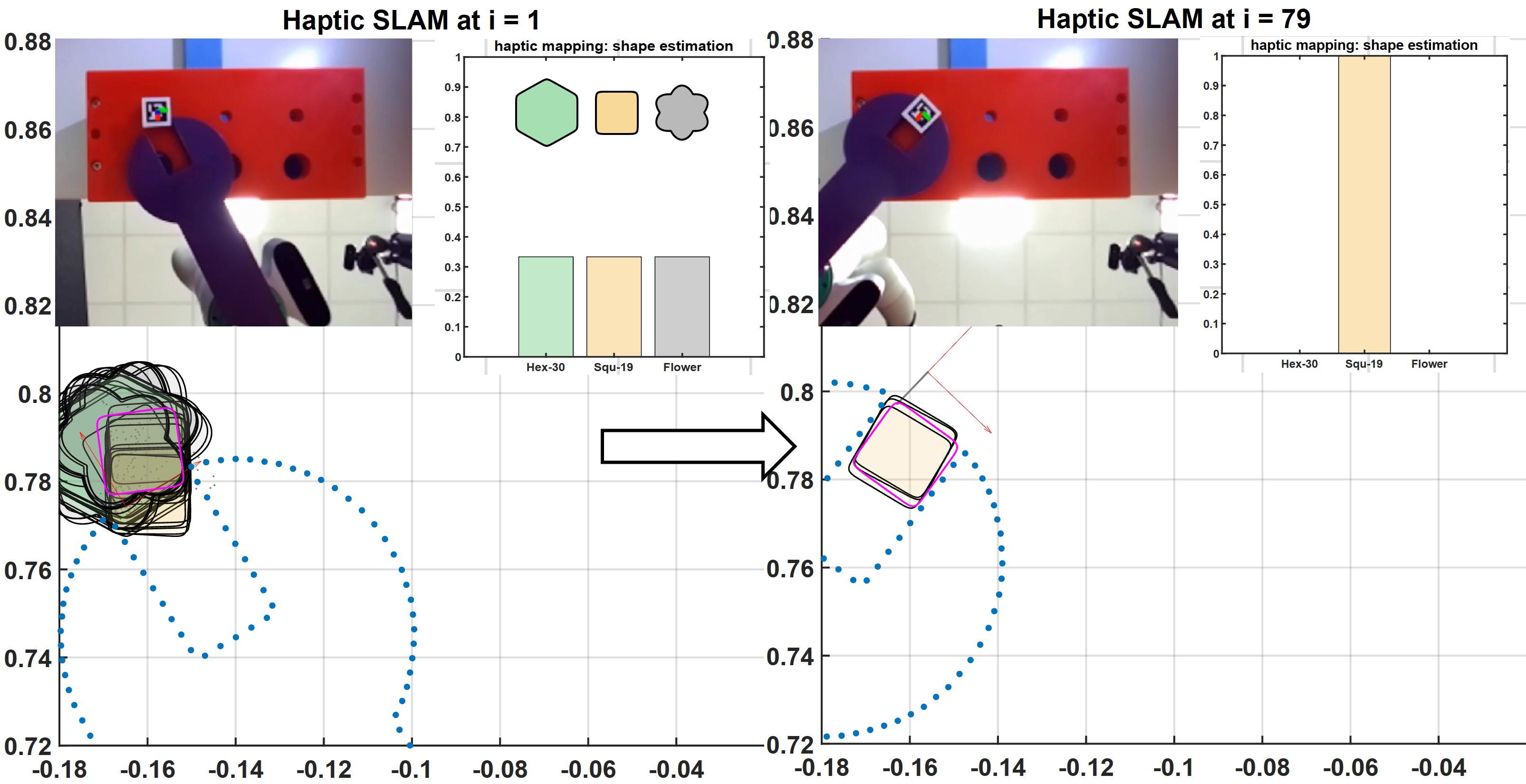}%
        \label{fig:346_4}
    } \\ 
    \vspace{1pt}
    \subfloat[Particles of $s^{(3)}$: Flw-33 (Oversized)]{
        \includegraphics[width=0.95\columnwidth]{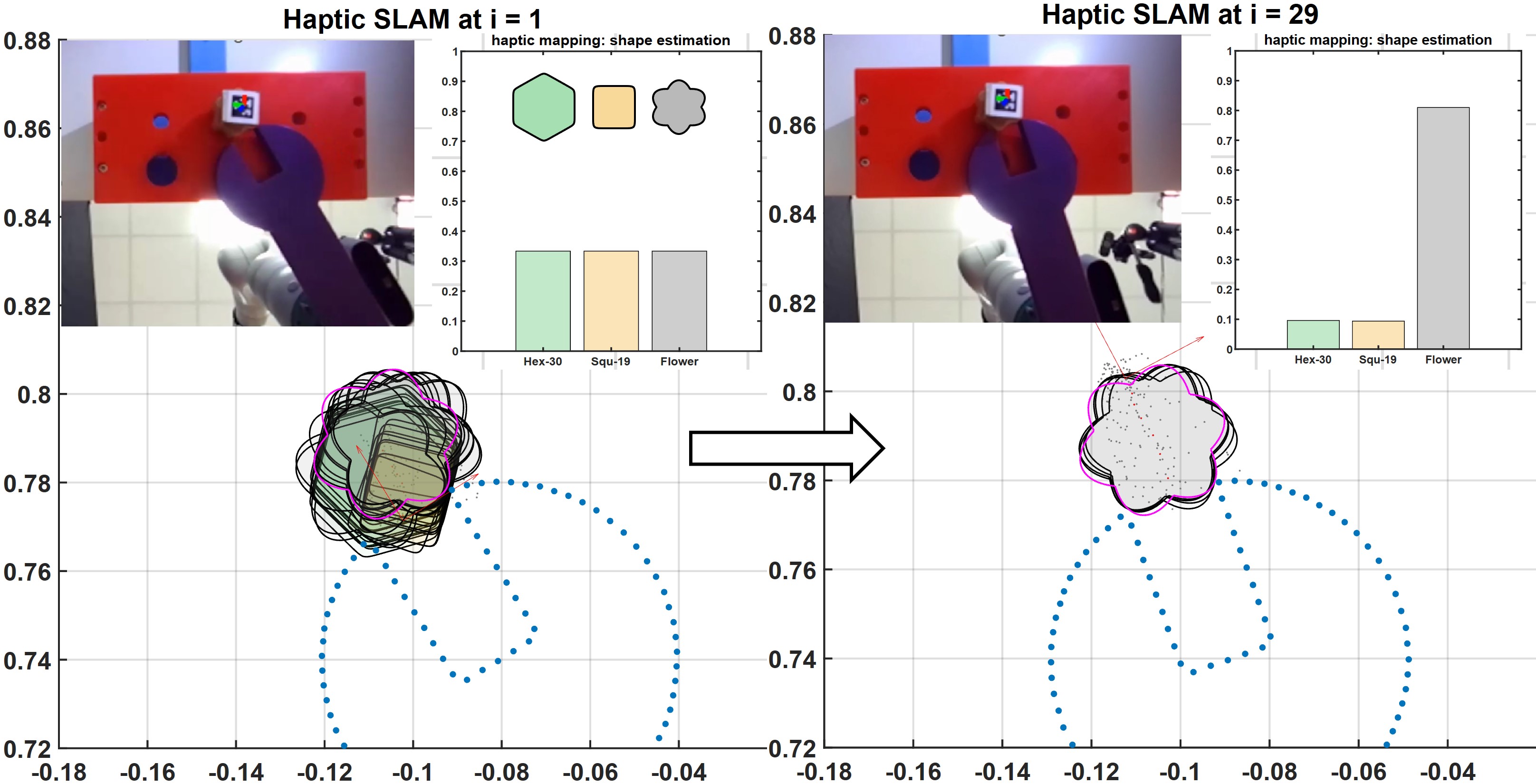}%
        \label{fig:346_6}
    }
    \caption{Particle evolution of haptic SLAM during interaction. Each row shows the transition from the initial contact state (left) to the final converged belief (right). The system identifies the screw shape and tracks its pose during both insertion and rotation.}
    \label{fig:Particle_evolution}
\end{figure}

\subsubsection{Evolution of Particle Distributions}
We present the evolution of particle distributions for each shape hypothesis $\{s^{(i)}, \V \theta^{(j)}[s^{(i)}] \}$. The estimation process is initialized at the moment of first contact, as illustrated on the left side of Fig.\ref{fig:Particle_evolution}. In this scenario, a tool (spanner-21) is used to interact with three screw types: Hex-30, Squ-19, and Flw-33. Due to geometric constraints, the Hex-30 and Flw-33 screws are oversized relative to the tool, while the Squ-19 screw provides a compatible fit.

As shown in each panel of Fig.\ref{fig:Particle_evolution}, the top-left corner provides a ZED2 camera screenshot for visual reference of the real-world setup. The top-right corner displays the probability weight distribution $w^{(i)}$ across the screw hypotheses, which are initially uniform. The lower portion of each panel visualizes the spatial particle distribution, where the tool is represented as a blue point cloud and the ground-truth screw pose is shown in magenta.

By the end of the estimation process (right column), the robot successfully classifies the screw shape, with the weights converging to the correct screw types for $s^{(1)}$ and $s^{(3)}$. For the matched case in Fig.\ref{fig:346_4}, the algorithm identifies the Squ-19 screw, enables successful insertion, and subsequently rotates the tool to loosen it. As illustrated on the right side of the panel, the particles remain tightly clustered on the screw even as the magenta ground truth rotates. These results demonstrate the framework's capability for both static and dynamic pose and type estimation. 

\begin{table*}[t]
\centering
\caption{Quantitative Performance of Haptic SLAM across 3-Hypothesis Experiments}
\label{table:3_class_final}
\resizebox{\textwidth}{!}{
\begin{tabular}{llcccccccccc}
\toprule
\multirow{2}{*}{\shortstack{Exp. \& \\ Spanner}} & \multirow{2}{*}{Screw Type} & \multirow{2}{*}{Condition} & \multicolumn{2}{c}{Success Rate} & \multicolumn{2}{c}{Static Error} & \multicolumn{2}{c}{Dynamic Error} & \multicolumn{2}{c}{Interaction Force (N)} & \multirow{2}{*}{\shortstack{Task \\ Time (s)}} \\ 
\cmidrule(lr){4-5} \cmidrule(lr){6-7} \cmidrule(lr){8-9} \cmidrule(lr){10-11}
& & & $S_{id}$ & $S_{mn}$ & Pos. (mm) & Rot. (deg) & Pos. (mm) & Rot. (deg) & Peak & Avg. & \\ 
\midrule
\multirow{3}{*}{\shortstack{Exp 1 \\ (34 mm)}} & Hex-36 & Oversized & 10/10 & - & $2.65 \pm 1.23$ & $4.09 \pm 3.17$ & - & - & 24.17 & 5.53 & $12.93 \pm 0.28$ \\
& Hex-34 & Matched & 10/10 & 10/10 & $3.33 \pm 1.51$ & $2.60 \pm 1.57$ & $3.95 \pm 1.54$ & $1.48 \pm 0.32$ & 14.97 & 2.35 & $10.96 \pm 3.36$ \\
& Squ-19 & Undersized & 10/10 & - & $3.02 \pm 4.89$ & $4.21 \pm 1.85$ & - & - & 4.21 & 0.36 & $13.03 \pm 0.22$ \\ 
\midrule
\multirow{3}{*}{\shortstack{Exp 2 \\ (34 mm)}} & Hex-36 & Oversized & 10/10 & - & $2.07 \pm 0.45$ & $5.57 \pm 1.86$ & - & - & 24.20 & 7.24 & $17.83 \pm 0.26$ \\
& Rec-20 & Undersized & 10/10 & - & $4.90 \pm 2.50$ & $3.29 \pm 2.70$ & - & - & 8.08 & 1.42 & $16.89 \pm 2.93$ \\
& Flw-33 & Matched & 10/10 & 9/10 & $3.15 \pm 1.58$ & $2.51 \pm 1.70$ & $4.00 \pm 0.87$ & $3.60 \pm 1.53$ & 10.99 & 1.96 & $13.33 \pm 6.19$ \\ 
\midrule
\multirow{3}{*}{\shortstack{Exp 3 \\ (21 mm)}} & Hex-30 & Oversized & 10/10 & - & $1.97 \pm 0.65$ & $3.06 \pm 1.60$ & - & - & 25.79 & 6.96 & $12.85 \pm 0.22$ \\
& Squ-19 & Matched & 10/10 & 10/10 & $4.32 \pm 2.50$ & $2.51 \pm 0.96$ & $4.51 \pm 1.42$ & $2.16 \pm 0.53$ & 8.39 & 1.39 & $11.99 \pm 2.90$ \\
& Flw-33 & Oversized & 10/10 & - & $2.67 \pm 2.14$ & $2.52 \pm 1.73$ & - & - & 24.49 & 6.31 & $12.92 \pm 0.27$ \\ 
\midrule
\multirow{3}{*}{\shortstack{Exp 4 \\ (36 mm)}} & Hex-36 & Matched & 10/10 & 10/10 & $3.80 \pm 1.74$ & $3.79 \pm 1.78$ & $4.09 \pm 1.84$ & $3.10 \pm 1.69$ & 11.11 & 1.90 & $12.69 \pm 5.47$ \\
& Hex-30 & Undersized & 10/10 & - & $3.79 \pm 3.66$ & $3.29 \pm 3.67$ & - & - & 6.27 & 0.91 & $13.26 \pm 0.27$ \\
& Rec-20 & Undersized & 10/10 & - & $3.00 \pm 1.38$ & $2.90 \pm 1.10$ & - & - & 11.19 & 2.17 & $13.42 \pm 0.31$ \\ 
\bottomrule
\end{tabular}
}
\end{table*}

\subsubsection{Qualitative Analysis}
\begin{figure}[!h]
    \centering
    \subfloat[Exp 1: Spanner-34 (Hex-36, Hex-34, Squ-19)]{
        \includegraphics[width=0.45\columnwidth]{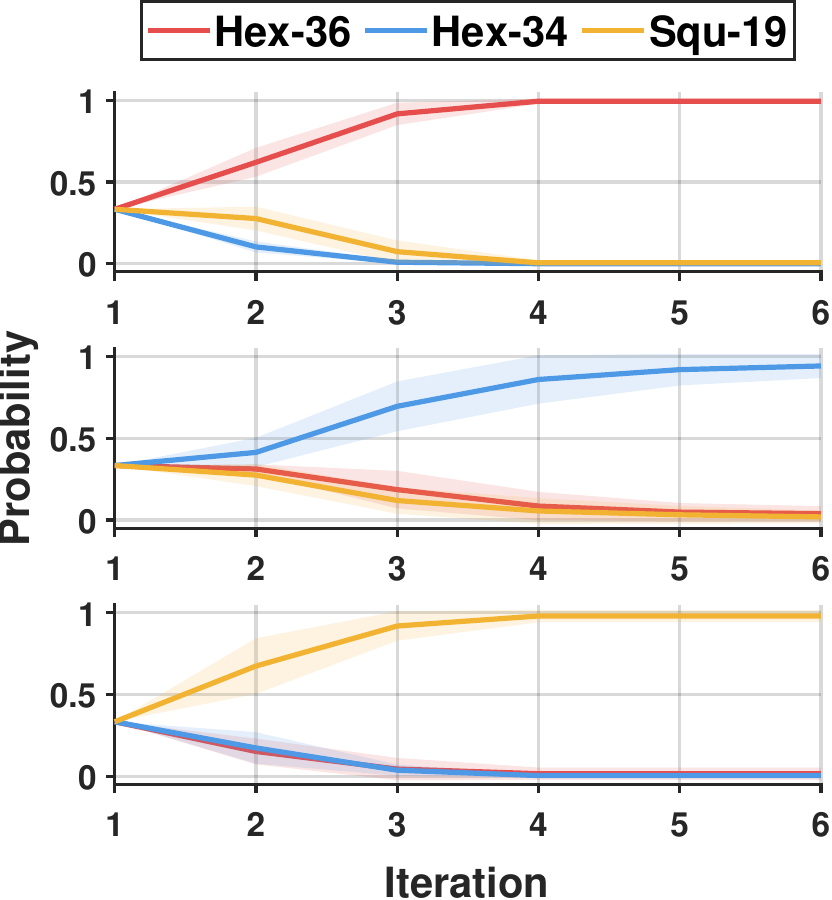}%
        \label{fig:2x2_a_weight}
    }
    \hfil 
    \subfloat[Exp 2: Spanner-34 (Hex-36, Rec-20, Flw-33)]{
        \includegraphics[width=0.45\columnwidth]{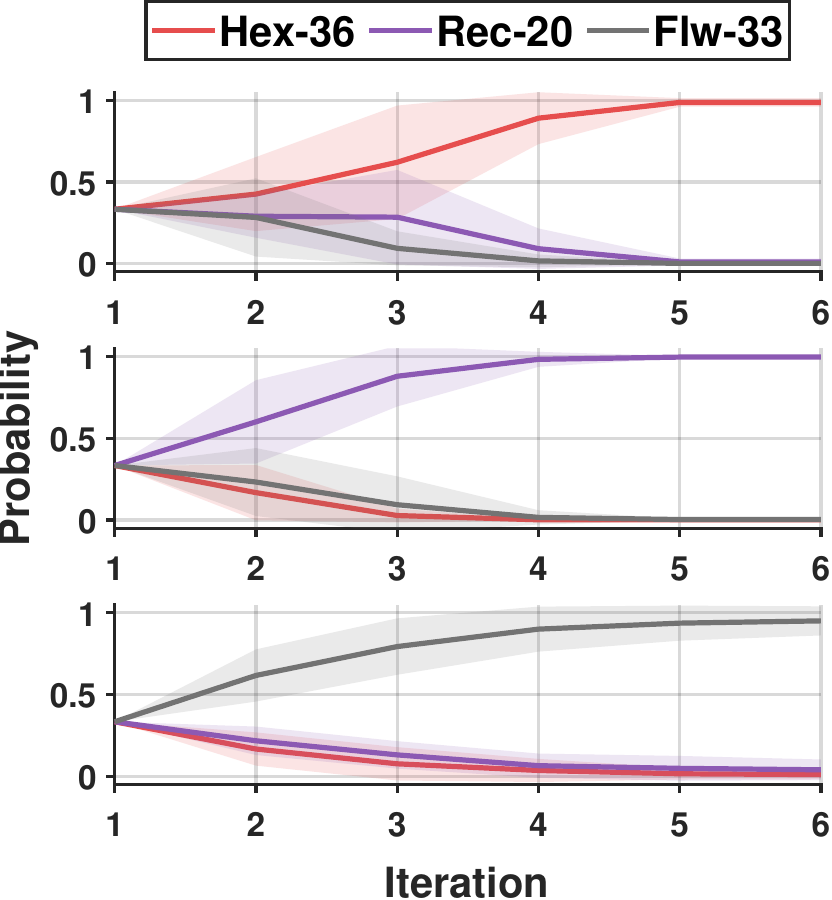}%
        \label{fig:2x2_b_weight}
    }
    \vspace{0.2pt} 
    \subfloat[Exp 3: Spanner-21 (Hex-30, Squ-19, Flw-33)]{
        \includegraphics[width=0.45\columnwidth]{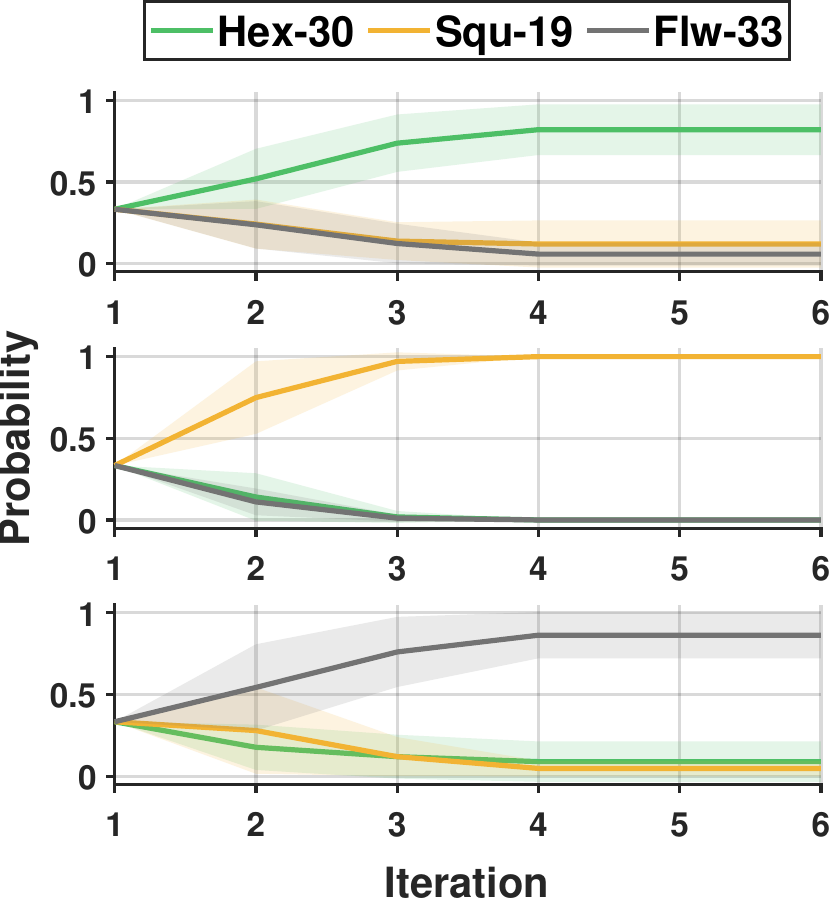}%
        \label{fig:2x2_c_weight}
    }
    \hfil
    \subfloat[Exp 4: Spanner-36 (Hex-36, Hex-30, Rec-20)]{
        \includegraphics[width=0.45\columnwidth]{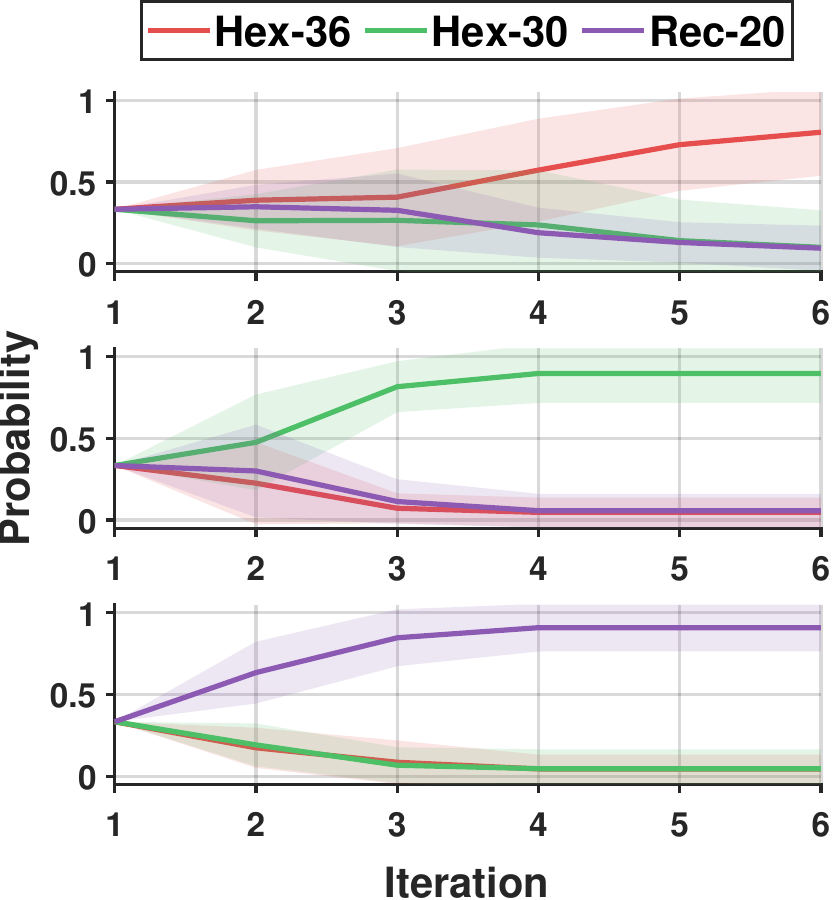}%
        \label{fig:2x2_d_weight}
    }
    \caption{Probability weight evolution across four three hypothesis experiments. Solid lines and shaded regions represent the mean and variance of the screw type weight.}
    \label{fig:overall_2x2_weight}
\end{figure}

\textbf{Overall Performance} Table \ref{table:3_class_final} summarizes the quantitative performance across 120 real world trials (4 Exp $\times$ 3 classes $\times$ 10 trials). We evaluate performance using two metrics: \textit{Identification rate} ($S_{id}$), which measures successful classification during the static interaction phase, and \textit{manipulation success rate} ($S_{mn}$), which indicates successful screw loosening following correct identification. The system achieved a $100\%$ identification rate ($S_{id}$) across all scenarios. Manipulation performance ($S_{mn}$) was similarly robust, with only a single failure observed in the Flw-33 case caused by mechanical interlocking on the non-convex geometry. The static error denotes the pose estimation error during the insertion phase after convergence, whereas the dynamic error represents the mean pose estimation error across temporal batches during the screw-loosening phase. The \textit{interaction force} (peak and average) metrics provide physical validation: oversized scenarios exhibit high forces due to the intentional haptic mismatch, whereas matched cases show minimal force. Furthermore, the entire perception-action loop proves efficient, with an average task time consistently below 18 seconds.

\begin{figure}[!h]
    \centering
    \subfloat[Exp 1: Spanner-34 (Hex-36, Hex-34, Squ-19)]{
        \includegraphics[width=0.45\columnwidth]{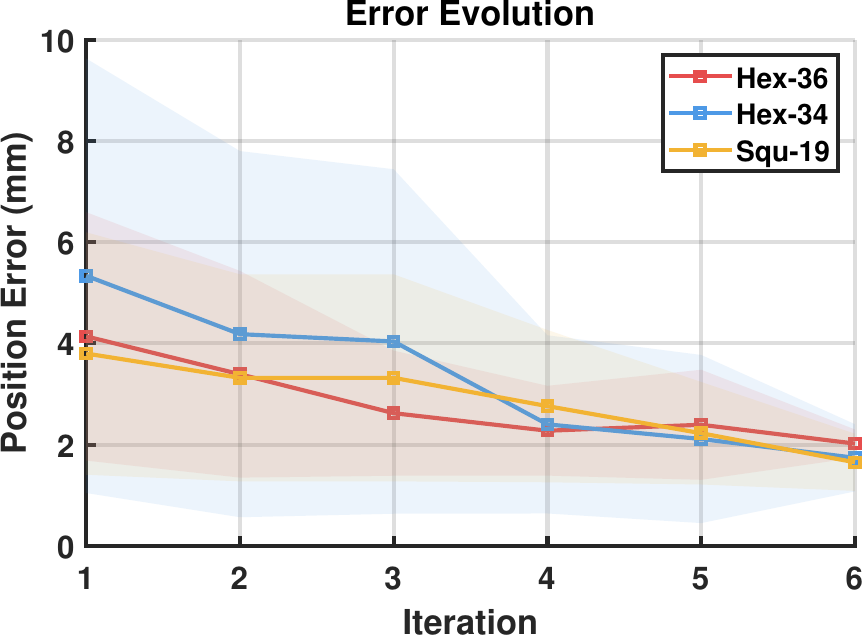}%
        \label{fig:2x2_b}
    }
    \hfil 
    \subfloat[Exp 2: Spanner-34 (Hex-36, Rec-20, Flw-33)]{
        \includegraphics[width=0.45\columnwidth]{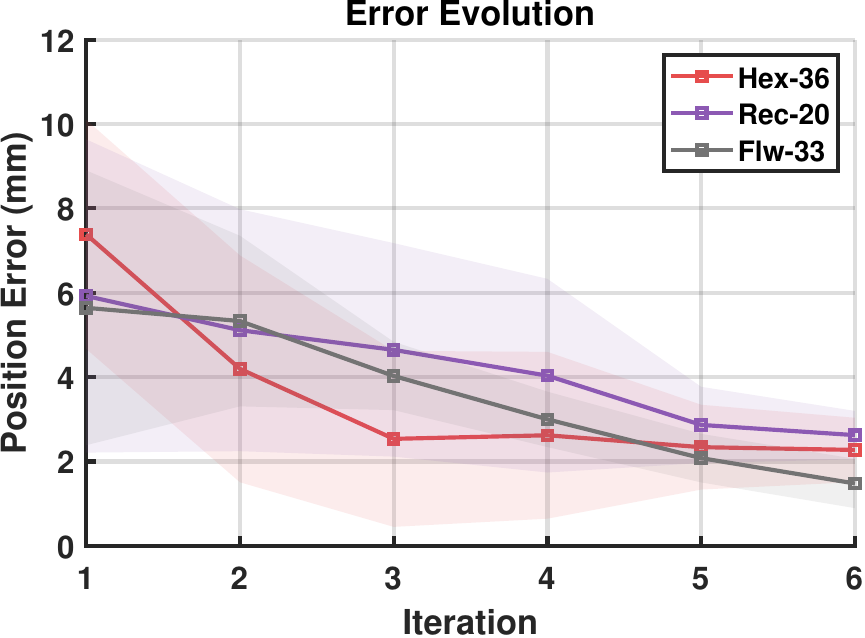}%
        \label{fig:2x2_c}
    }
    \vspace{0.2pt} 
        \subfloat[Exp 3: Spanner-21 (Hex-20, Squ-19, Flw-33)]{
        \includegraphics[width=0.45\columnwidth]{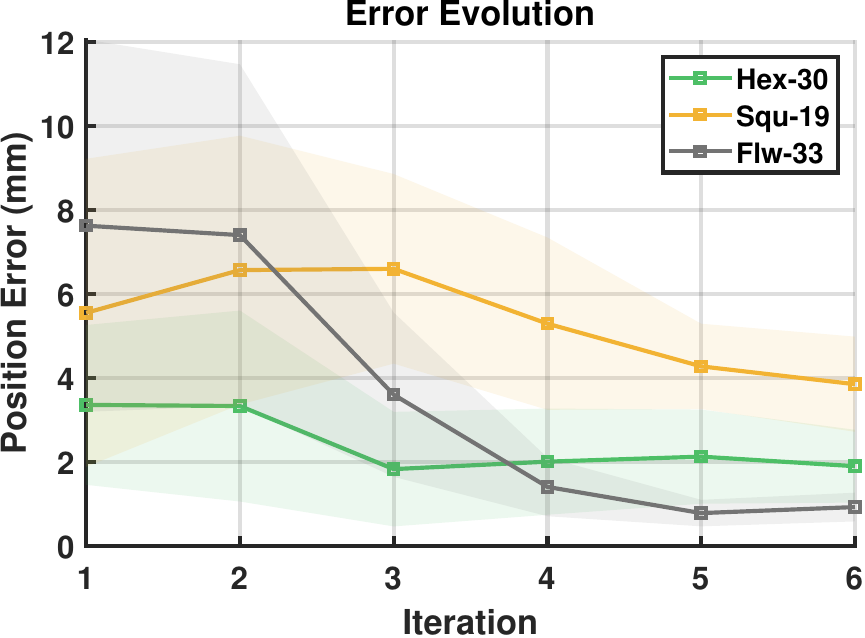}%
        \label{fig:2x2_a}
    }
    \hfil
    \subfloat[Exp 4: Spanner-36 (Hex-36, Hex-30, Rec-20)]{
        \includegraphics[width=0.45\columnwidth]{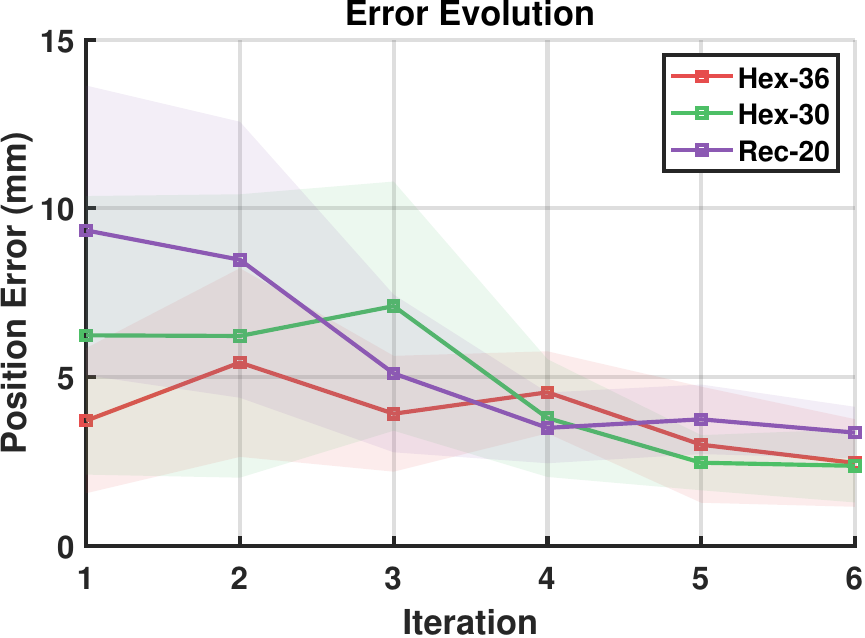}%
        \label{fig:2x2_d}
    }
    \caption{Evolution of position error and pose uncertainty across four three-hypothesis experiments. Shaded areas represent the covariance of the particles exceeding the weight threshold. Despite variations in screw geometry, the position error converges.}
    \label{fig:overall_2x2_error}
\end{figure}

\textbf{Shape Identification}: To further validate the robustness of the identification rate ($S_{id}$), Fig.~\ref{fig:overall_2x2_weight} illustrates the evolution of screw-type probability weights across ten trials for each experiment. The weights exhibit clear convergence toward the correct shape. While most scenarios demonstrate rapid convergence, Exp.~4 (Fig.~\ref{fig:2x2_d_weight}) presents unique challenges due to the geometric ambiguity of the rectangular screw (Rec-20). If the tool initially engages along the short axis, the resulting obstruction can be misinterpreted as an oversized Hex-36 screw. Our framework effectively reorients the tool to explore alternative axes based on the current estimate $\mathbf{\theta}^*$, such as transitioning from long-axis engagement to a short-axis probe, thereby gathering the discriminative haptic observations necessary to resolve the ambiguity. Consequently, Exp.~4 requires a greater number of iterations to achieve stable convergence compared with the other experiments.

\textbf{Pose Convergence}: The convergence behavior of the estimated pose is illustrated in Fig. \ref{fig:overall_2x2_error}, which tracks the mean and covariance of the particle set across iterations. For each experimental scenario, we evaluate the representative trials by considering only particles with weights exceeding a threshold. The results show that the pose error, measured relative to the AprilTag ground truth, converges to approximately 2\,mm.

\begin{figure}[!h]
\centering
\includegraphics[width=0.85\columnwidth]{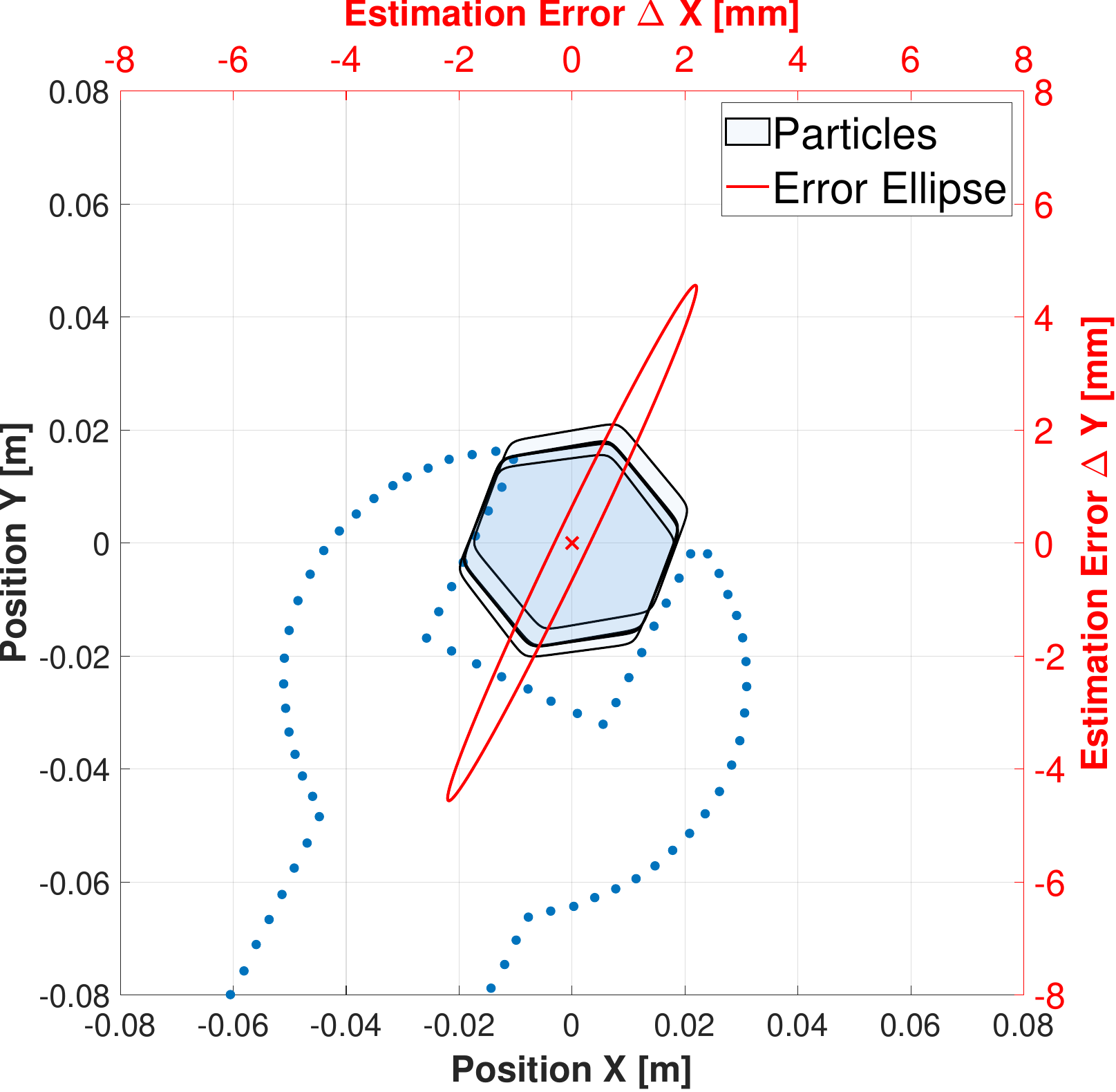}
\caption{Analysis of anisotropic pose uncertainty. Blue hexagons denote the particle distribution, while the red curve indicates the associated error ellipse. The anisotropy reflects the contrast between stiff lateral constraints and the geometric clearance along the tool opening.}
\label{fig:error_ana}
\end{figure}

\begin{table*}[t]
\centering
\caption{Quantitative Results of the 6-Hypothesis Challenge (34 mm Spanner)}
\label{table:6_class_final}
\resizebox{\textwidth}{!}{
\begin{tabular}{llcccccccccc}
\toprule
\multirow{2}{*}{\shortstack{Exp. \& \\ Spanner}} & \multirow{2}{*}{Screw Type} & \multirow{2}{*}{Condition} & \multicolumn{2}{c}{Success Rate} & \multicolumn{2}{c}{Static Error} & \multicolumn{2}{c}{Dynamic Error} & \multicolumn{2}{c}{Interaction Force (N)} & \multirow{2}{*}{\shortstack{Task \\ Time (s)}} \\ 
\cmidrule(lr){4-5} \cmidrule(lr){6-7} \cmidrule(lr){8-9} \cmidrule(lr){10-11}
& & & $S_{id}$ & $S_{mn}$ & Pos. (mm) & Rot. (deg) & Pos. (mm) & Rot. (deg) & Peak & Avg. & \\ 
\midrule
\multirow{6}{*}{\shortstack{6-Class \\ Batch \\ (34 mm)}} & Hex-36 & Oversized & 10/10 & -- & $2.31 \pm 0.93$ & $3.36 \pm 1.66$ & -- & -- & 24.42 & 6.86 & $24.09 \pm 0.85$ \\ 
& Hex-34 & Matched & 8/10 & 10/10 & $2.71 \pm 2.23$ & $1.94 \pm 1.05$ & $3.24 \pm 2.64$ & $4.11 \pm 1.65$ & 10.75 & 2.08 & $14.39 \pm 5.31$ \\ 
& Hex-30 & Undersized & 6/10 & -- & $1.98 \pm 0.81$ & $2.51 \pm 2.07$ & -- & -- & 5.94 & 0.94 & $23.17 \pm 6.38$ \\ 
& Squ-19 & Undersized & 7/10 & -- & $2.83 \pm 1.86$ & $2.51 \pm 1.15$ & -- & -- & 4.45 & 0.73 & $24.88 \pm 0.68$ \\ 
& Rec-20 & Undersized & 6/10 & -- & $3.57 \pm 2.39$ & $3.08 \pm 1.65$ & -- & -- & 7.29 & 1.36 & $28.08 \pm 5.55$ \\ 
& Flw-33 & Matched & 5/10 & 8/10 & $4.28 \pm 3.08$ & $3.54 \pm 3.00$ & $3.34 \pm 2.60$ & $4.04 \pm 4.21$ & 7.91 & 1.66 & $17.40 \pm 5.71$ \\ 
\bottomrule
\end{tabular}
}
\end{table*}
\textbf{Error Analysis}: 
The residual error can be attributed to modeling approximations, numerical optimization, and the physical interaction process. To further elucidate the latter, we examine the mechanical interaction illustrated in Fig.~\ref{fig:error_ana}. By extracting the pose covariance matrix $\mathbf{\Sigma}_{\mathbf{\theta}}$, we construct a 95\% confidence error ellipse to characterize the estimation uncertainty. To facilitate clear visualization of the sub-millimeter distribution, this ellipse is indicated by the red secondary axes, together with the corresponding estimation errors ($\Delta X, \Delta Y$ in mm).
Fundamentally, this anisotropic uncertainty is governed by the richness of the haptic observations, specifically the extent to which the tool contacts multiple edges of the screw during interaction. The principal axes of the ellipse reveal that while the tool provides stiff lateral constraints, the longitudinal direction (along the spanner opening) often remains relatively unconstrained. This elongated variance typically emerges when the tool observes only a limited number of edges, such as during the rotational loosening phase where contact is primarily maintained on the driving faces, leaving the sliding direction under-observed. It is important to note that this anisotropic distribution represents a typical statistical behavior rather than a strict deterministic outcome for every individual trial. In cases where the haptic observations cover multiple edges of the screw, the pose can become tightly constrained in all directions. Furthermore, for loose-fitting geometries such as Squ-19, the screw's valid pose effectively floats within a region defined by the mechanical tolerances, leading to larger overall estimation variance.

\subsection{Six-Hypothesis Stress Test}\label{sec:6class}
To investigate the scalability and robustness of the Haptic SLAM framework under increased environmental complexity, we conduct an extended stress-test evaluation using the 34 mm tool (spanner-34) against the full library of six candidate geometries, $\mathcal{S} =\{ {s^{(1)}, \dots, s^{(6)}} \}$.
By doubling the hypothesis space and increasing structural ambiguity, this experiment probes the system's ability to resolve geometrically similar geometries (e.g., Hex-34 vs. Flw-33). The robot must decode subtle haptic mismatches between predicted and sensed wrenches to simultaneously disambiguate the physical model and track object pose within a high dimensional manifold hypothesis space.

\subsubsection{Quantitative Results}
Table \ref{table:6_class_final} summarizes the quantitative performance for the 6-class challenge. While the oversized Hex-36 is identified and handled with $100\%$ ($10/10$) success, scenarios involving geometrically matched or undersized screws exhibit a notable divergence between identification rates ($S_{id}$) and manipulation success rate ($S_{mn}$). Additionally, the average task time is longer (approx. $24-28$ s) compared to the 3-class experiments ($\approx 13$ s). We attribute these performance shifts to two key factors:

\begin{itemize}
\item {Task Robustness under Ambiguity:}
The manipulation success rate often exceeds identification accuracy for compatible geometries. For instance, the Hex-34 screw achieves perfect manipulation success (10/10) despite some misclassifications (8/10), demonstrating robustness to semantic errors. In contrast, the Flw-33 screw is more sensitive due to its inherently wider particle distribution. When misclassified as Hex-34 (5/10), this higher variance leads to significant pose estimation errors, which directly result in some manipulation failures (8/10).

\item {Exploration for Information Accumulation:} 
The increased duration arises from the under-constrained nature of interactions within a larger hypothesis space. To resolve structural ambiguities among six candidates, the planner must perform more extensive exploratory motions to accumulate sufficient observations for confident disambiguation.
\end{itemize}

\subsubsection{Overall particle evolution}
\begin{figure}[!htb]
\centering
\includegraphics[width=0.5\textwidth]{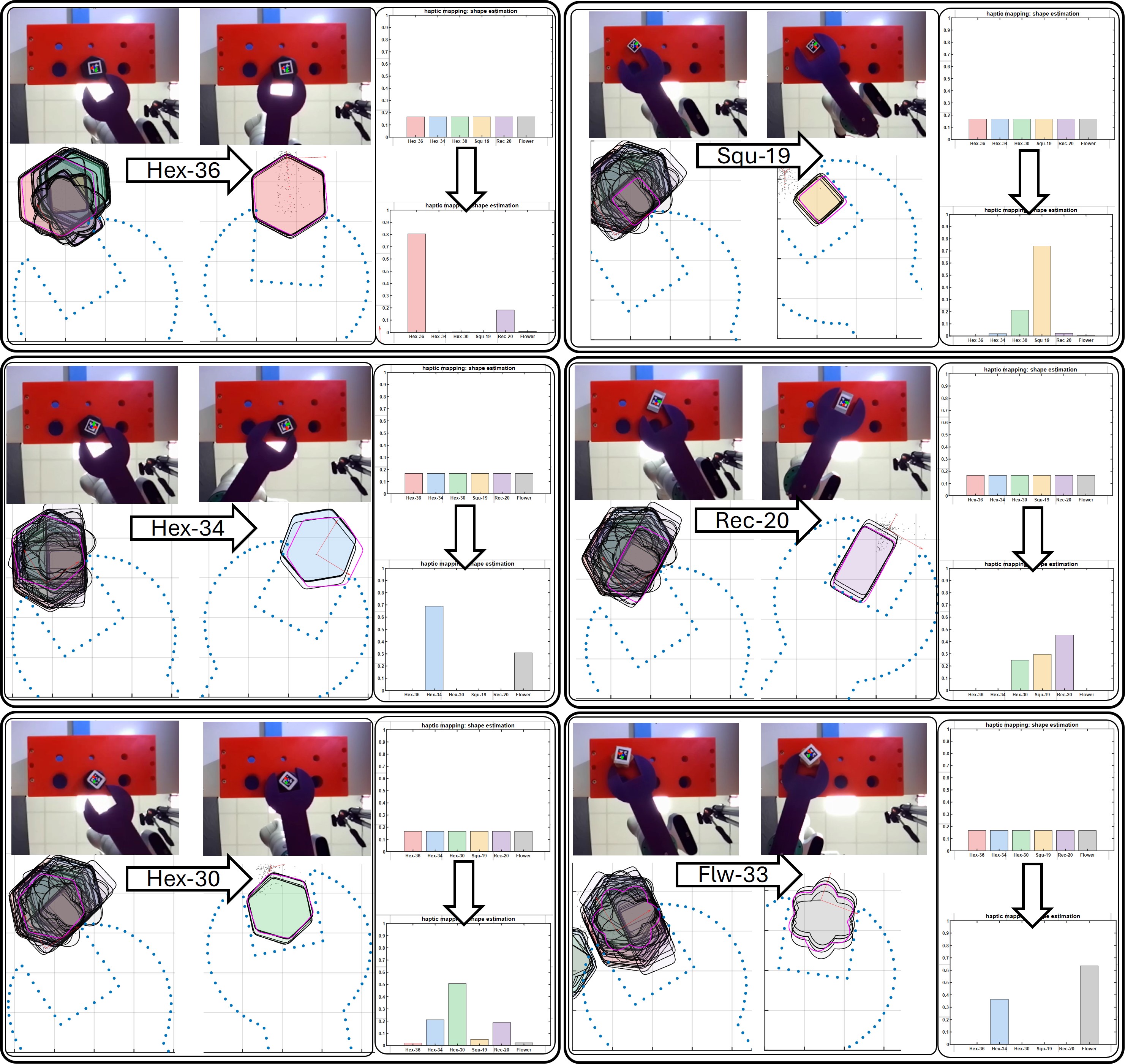}
\caption{Typical particle evolution for the 6-class challenge. The system must simultaneously reason over six discrete shape hypotheses and their associated continuous poses. Unlike the rapid convergence in simpler tasks, the system here may compete weight of shape.}
\label{fig:6class}
\end{figure}

We visualize the belief and pose evolution for a representative trial of the 6-hypothesis challenge in Fig.~\ref{fig:6class}. Unlike the unambiguous 3-hypothesis scenarios, this case highlights the difficulty of distinguishing between geometrically similar structures (e.g., Hex-34 vs. Flw-33). As shown in the weight evolution plot (top-right panel), the probability distribution does not converge to a single mode. Instead, the weights maintain competing hypotheses for an extended period, reflecting the haptic ambiguity inherent to the task. This behavior aligns with the quantitative findings in the confusion matrix and explains the increased exploration time required in this setting.

\subsubsection{Ambiguity Analysis: Confusion Matrix and Belief Evolution}
\begin{figure}[!b]
\centering
\includegraphics[width=0.9\columnwidth]{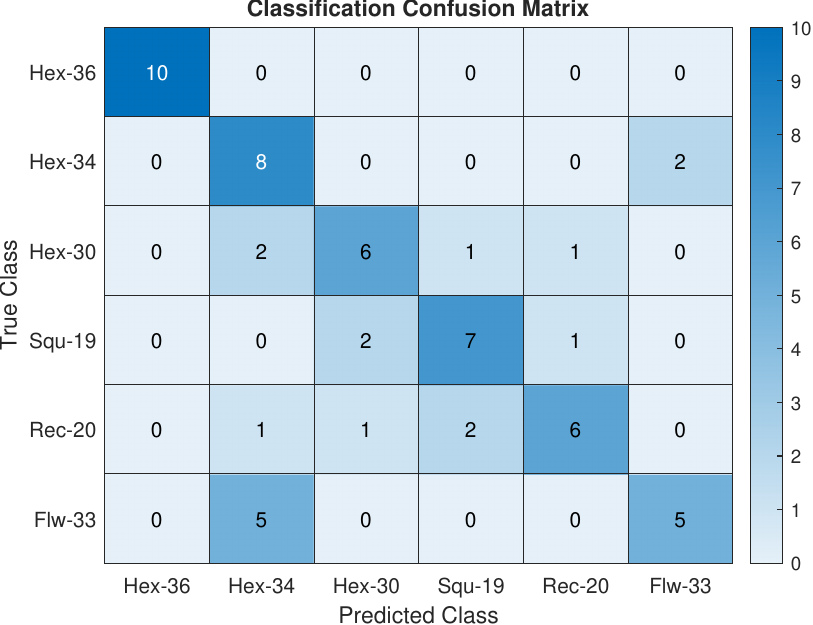}
\caption{Confusion matrix for 6-hypothesis stress test. The system exhibits high accuracy for oversized screw (Hex-36) but faces ambiguity where shapes are geometrically similar (Hex-34 vs. Flw-33) or interactions are under-constrained (undersized screws).}
\label{fig:Confusion}
\end{figure}

\begin{figure}[!htb]
\centering
\includegraphics[width=0.5\textwidth]{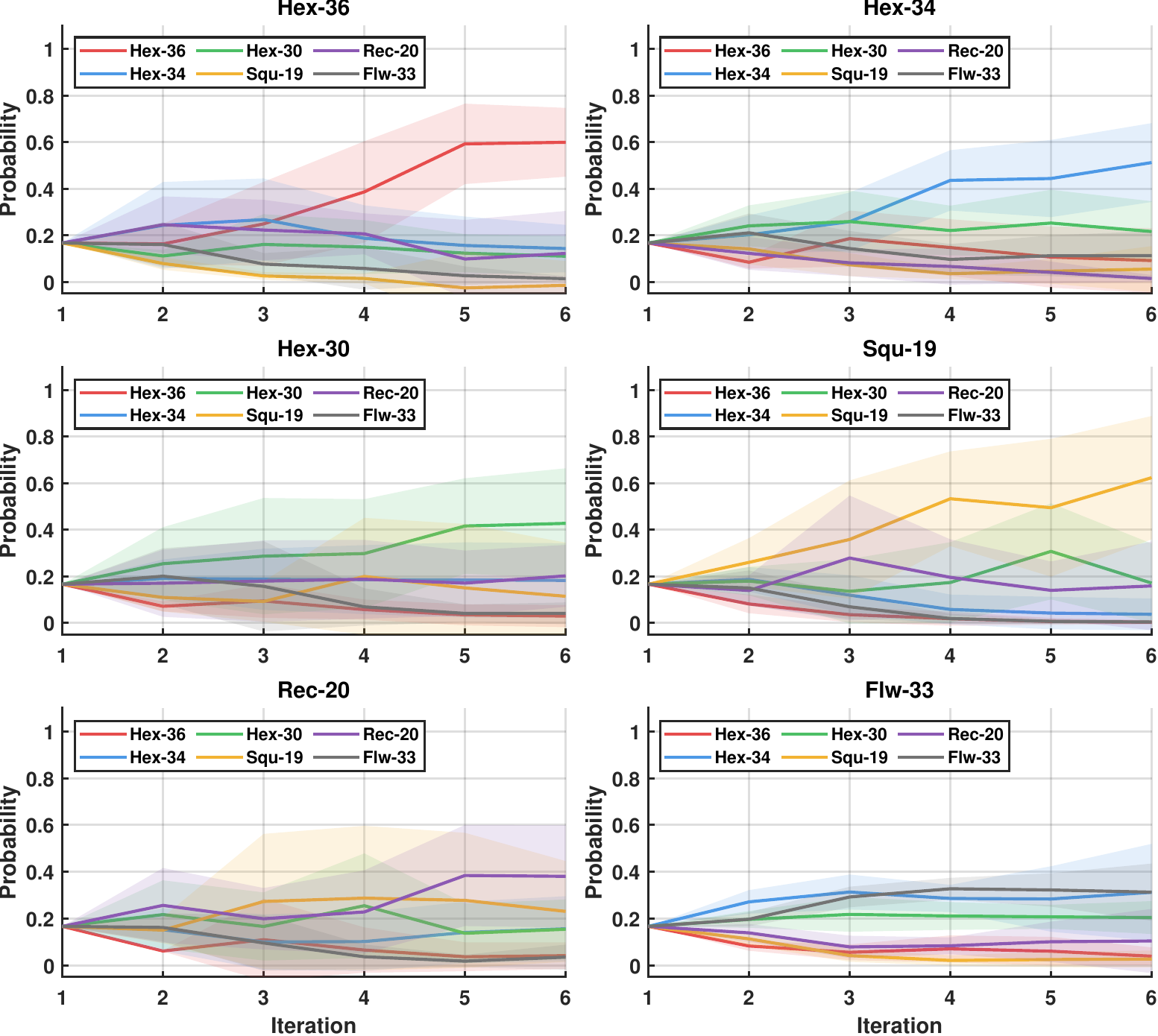}
\caption{Evolution of probability weights across different screw types in 6-hypothesis stress test.}
\label{fig:Weight_Evolution}
\end{figure}

The confusion matrix (Fig.~\ref{fig:Confusion}) and probability weight evolution (Fig.~\ref{fig:Weight_Evolution}) reveal the classification behavior across different geometries. For oversized screws such as Hex-36, classification is unambiguous ($10/10$ accuracy) because the physical obstruction generates a distinct, high-force haptic mismatch that allows the particle filter to reject all smaller hypotheses.

However, significant ambiguity arises between the physically compatible Hex-34 and Flw-33 screws. Since both geometries permit tool insertion, the haptic feedback during the insertion phase is often insufficient to distinguish between them. This is evident in the weight evolution for the Flw-33 trial (Fig.~\ref{fig:Weight_Evolution}, bottom-right), where the probability weights for Flw-33 and Hex-34 remain entangled and converge to nearly equal values, reflecting the system's uncertainty. Interestingly, the confusion is asymmetric: while Hex-34 is correctly identified in 8/10 trials, Flw-33 is frequently misclassified as Hex-34 (5/10 error rate). This suggests that the internal model tends to favor the Hex-34 hypothesis.

Finally, the undersized screws (Hex-30, Squ-19, Rec-20) exhibit mutual confusion, which is a direct consequence of the task-driven nature of our planner. Unlike active exploration strategies designed to maximize information gain by tracing object contours, our planner aims to achieve successful insertion. When the tool fits loosely over a small screw, the planner can complete the insertion without necessarily contacting the discriminative features required for unique identification. Consequently, the system prioritizes efficient task completion over semantic certainty, leading to successful manipulation despite lower identification accuracy.

\subsection{Ablation Study}\label{sec:abl}

\begin{figure*}[!t]
    \centering
    \subfloat[Manipulation success rates across different settings.]{
        \includegraphics[width=0.355\textwidth]{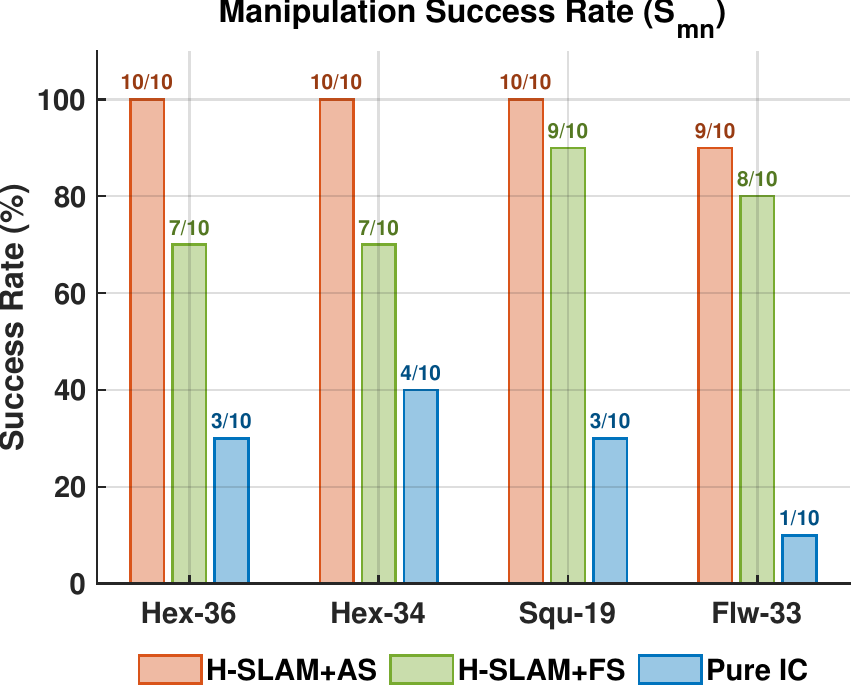}
        \label{fig:ablation_a}
    }
    \hspace{2mm}
    \subfloat[Box-plot analysis of interaction forces and torques across different ablation settings.]{
        \includegraphics[width=0.58\textwidth]{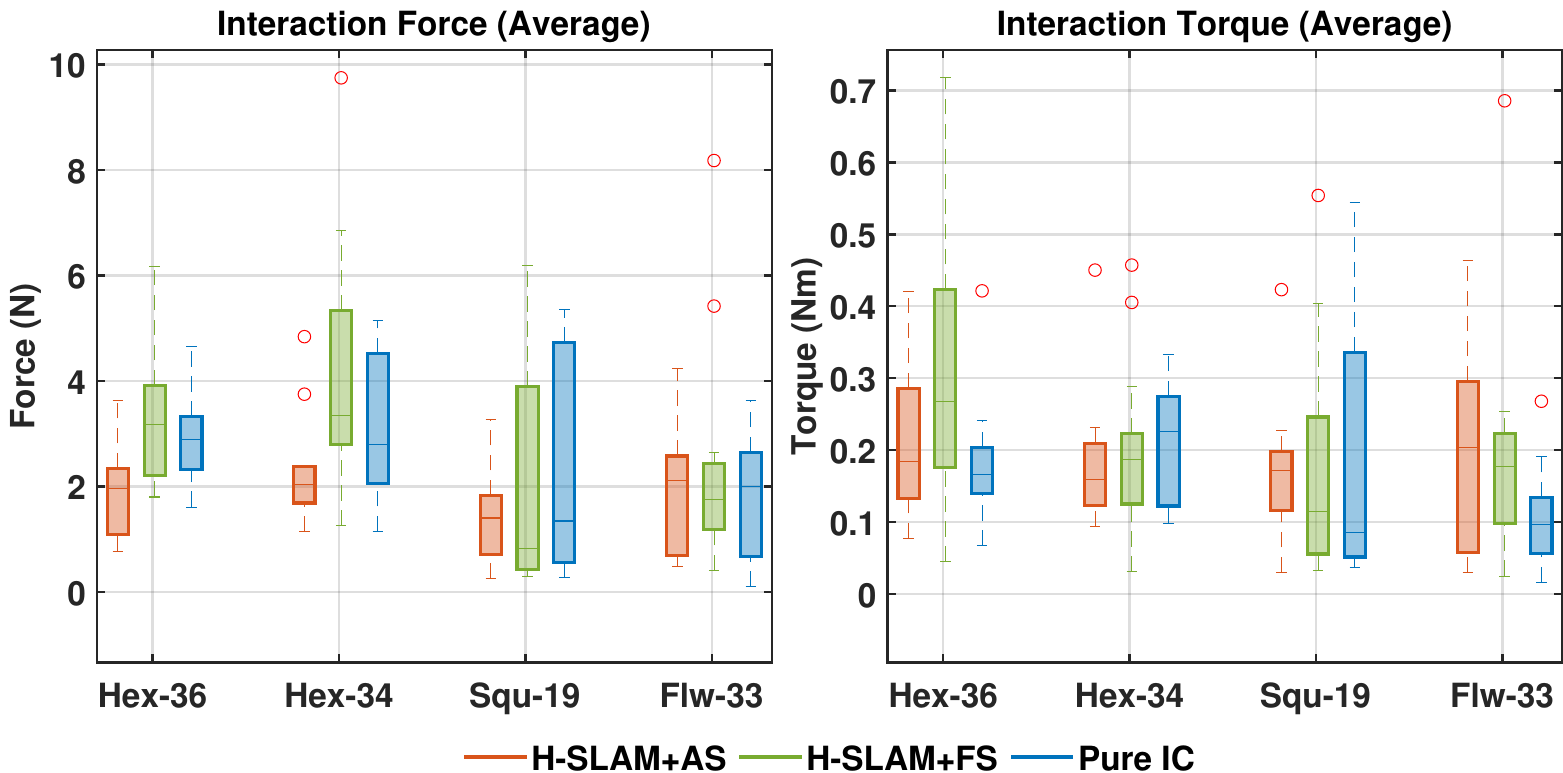}
        \label{fig:ablation_b}
    }
    \caption{Quantitative results of the ablation study across different screw geometries. (a) Manipulation success rates. The proposed framework (H-SLAM+AS) consistently achieves the highest success rates compared to fixed control stiffness ablation (H-SLAM+FS) and the baseline pure impedance control (Pure IC). (b) Distributions of average interaction forces and torques. The box plots demonstrate that the full H-SLAM+AS framework maintains safe and compliant contact, generally yielding lower and more stable interaction wrenches during the tasks, whereas baselines exhibit higher variance and larger contact forces.}
    \label{fig:ablation}
\end{figure*}

To our knowledge, no prior model-based works addressed tool-mediated, contact-rich manipulation through a unified framework that tightly integrates haptic state estimation, online planning, and adaptive control. Since existing physics-based methods typically decouple these components or assume a priori known environments, we evaluate the individual contributions of our perception and control modules by comparing the complete framework (H-SLAM + AS) against two ablated baselines:
\begin{enumerate}
\item \textbf{Pure impedance control (Pure IC):} A blind compliant control that relies solely on the initial pose estimated from the wrench-axis (Sec. \ref{sec:wrench_axis}) without subsequent Haptic SLAM updates. This evaluates the necessity of continuous state estimation.
\item \textbf{Haptic SLAM with fixed stiffness (H-SLAM + FS):} This baseline performs full state estimation but uses a constant control stiffness $\V K_c$ without uncertainty driven modulation. It reflects conventional or learning-based controllers that rely on fixed or predefined impedance schedules and treat perception and action as loosely coupled, without adapting stiffness under  uncertainty.
\end{enumerate}
Figure \ref{fig:ablation} summarizes the performance across four representative screw types.

\subsubsection{Success Rate Analysis}
As shown in Fig.~\ref{fig:ablation_a}, the complete H-SLAM + AS framework demonstrates superior robustness, consistently achieving the highest success rates. The baseline Pure IC yields the lowest success rate (10\%–40\%), confirming that the coarse wrench-axis initialization lacks the millimeter-level precision required for tight-tolerance insertion and frequently results in jamming. While introducing state estimation (H-SLAM + FS) drastically improves performance, failures still occur. Under fixed control stiffness, even minor orientation errors caused by sensor noise can generate excessive internal forces, causing the tool to jam before insertion is completed. In contrast, H-SLAM + AS dynamically lowers rotational stiffness under uncertainty, compliantly accommodating these minor misalignment while maintaining the necessary insertion force.

\subsubsection{Force and Interaction Analysis}
Figure~\ref{fig:ablation_b} illustrates the distribution of interaction wrenches. Our adaptive method (H-SLAM + AS) generally ensures the lowest and most consistent interaction forces. A notable exception is the plastic Flw-33 screw. Its lower friction allows the tool to naturally slide into the screw even under fixed stiffness, thereby mitigating force accumulation. For high-friction steel screws, however, the compliance provided by AS is critical for relieving internal stresses and preventing jamming.

Furthermore, while forces accumulate significantly during jamming (particularly in the FS baseline), the interaction torques do not exhibit the same drastic spikes. This behavior stems from the geometric stability of the spanner–screw interface: the two jaws of the spanner establish a stable two-point contact with the screw, effectively constraining rotational motion. Consequently, when the spanner becomes stuck under impedance control, the forward insertion command causes a rapid accumulation of force, while the structural stability of the spanner prevents excessive torque buildup.

\section{Conclusion}
This paper presented a unified, physics-based framework for Haptic SLAM and tool-mediated manipulation. Addressing the fundamental challenges of indirect sensing and visual occlusion inherent in tool use, the proposed approach tightly couples perception, planning, and control through a shared internal model based on parameterized equilibrium manifolds. Within this formulation, sparse local haptic feedback becomes sufficient to resolve geometric ambiguity and track object states through closed-loop interaction.

From a modeling perspective, the framework maintains a structural separation between conservative equilibrium constraints, which define the manifold used for inference, and dissipative effects that regularize forward interaction prediction. This separation preserves analytical tractability for gradient-based estimation while retaining physically consistent behavior.

Extensive simulations and real-world experiments demonstrate the robustness of the framework, achieving reliable object identification and millimeter-level pose tracking. The results highlight the importance of closing the perception–action loop, where estimation uncertainty directly informs adaptive impedance control to prevent jamming and ensure safe interaction.

Although a broad class of tool-mediated manipulation tasks can be addressed within the proposed quasi-static framework, certain tasks such as surface finishing, cutting, or handling deformable environments require extensions beyond the current formulation. Another limitation is the need for pre-tuning friction-related parameters (see Appendix~\ref{app:friction_effect}). Future work will extend the framework to incorporate online estimation of interaction parameters such as friction coefficients, full 3D manipulation with more complex multi-point contacts, and broader classes of tools and object geometries.

\section{Appendix}\label{Appendix}
\subsection{Derivation of the Generalized Friction Correction}
\label{app:friction_derivation}

This appendix details the construction of the generalized friction wrench $\bm{\mathcal{F}}^{fri}_i$ used in Section~\ref{subsec:fri}. In our formulation, friction is treated as a smooth, velocity-regularized dissipative correction that opposes tangential slip. 
In the quasi-static continuous-time interpretation, this regularization vanishes as slip velocity approaches zero and therefore does not alter the equilibrium manifold defined by $\partial_{\mathbf{z}} W = 0$; it modifies only the convergence path toward equilibrium. In practice, we realize this behavior using a predictor--corrector update in which friction is evaluated from a frictionless predictor velocity, yielding an explicit correction that preserves the same quasi-static equilibrium.

\subsubsection{Point Cloud Kinematics}
For each point cloud node $\mathbf{c}_i'$ fixed in the tool frame, its world-frame velocity is computed from the frictionless predictor update $\dot{\mathbf{z}}_{\mathrm{free}} = [\dot{x}, \dot{y}, \dot{\phi}]^T$:
\begin{equation}
    \dot{\mathbf{c}}_i
    =
    \dot{\phi}
    \!\begin{bmatrix}
        0 & -1 \\
        1 & 0
    \end{bmatrix}
    \!\mathbf{R}(\phi) \, \mathbf{c}_i'
    \,+ 
    \begin{bmatrix}
        \dot{x}\\
        \dot{y}
    \end{bmatrix}.
\end{equation}

The tangential slip velocity is obtained by projecting $\dot{\mathbf{c}}_i$ onto the local surface tangent:
\begin{equation}
    \dot{\mathbf{c}}_i^\Vert
    =
    \dot{\mathbf{c}}_i
    - (\dot{\mathbf{c}}_i^T \mathbf{n}_i) \, \mathbf{n}_i\,,
    \quad
    \mathbf{n}_i = \frac{\mathbf{f}_{Ni}}{\|\mathbf{f}_{Ni}\|},
\end{equation}
where $\mathbf{f}_{Ni} = -\V R(\phi) \, \partial_{\V{c}_i'} W_i$ is the normal contact force derived from the conservative contact potential.


\subsubsection{Velocity-Regularized Friction Model}
The tangential friction force is defined using a smooth saturation function:
\begin{equation}
\mathbf{f}^{fri}_i
=
-\mu \|\mathbf{f}_{Ni}\|
\tanh \!\left(
    \frac{b \|\dot{\mathbf{c}}_i^\Vert\|}
         {\mu \|\mathbf{f}_{Ni}\|}
\!\right)\!
\frac{\dot{\mathbf{c}}_i^\Vert}{\|\dot{\mathbf{c}}_i^\Vert\|}\,,
\end{equation}
where $\mu$ is the friction coefficient and $b$ is a regularization constant. 
At low velocities, this reduces to a linear viscous term $\mathbf{f}_i^{fri} \approx -b \, \dot{\mathbf{c}}_i^\Vert$, while at higher velocities it saturates at a Coulomb-like bound $\mathbf{f}_i^{fri} \approx -\mu \|\mathbf{f}_{Ni}\|$.
In practice, we add a small epsilon to the norm in the normalization terms to avoid numerical issues when denominator is near zero.

\subsubsection{Generalized Wrench Expansion}
Finally, the Cartesian friction force is lifted into the generalized state space to form a friction wrench:
\begin{equation}
    \bm{\mathcal{F}}^{fri}_i
    =
    \begin{bmatrix}
        \mathbf{f}^{fri}_i \\
        [\mathbf{R}(\phi) \, \mathbf{c}_i']^T
        \begin{bmatrix}
            0 & -1 \\
            1 & 0
        \end{bmatrix}
        \mathbf{f}^{fri}_i
    \end{bmatrix}.
\end{equation}

This generalized wrench is used as an explicit correction term in the forward quasi-static update and is not included in the equilibrium constraint or the implicit differentiation used for SLAM. Since the friction regularization vanishes in the zero-velocity limit, it does not support a nonzero potential gradient at equilibrium and therefore does not define a state-based equilibrium constraint. Accordingly, the SLAM equilibrium remains defined by $\partial_{\mathbf{z}} W = 0$, with friction used only to regularize forward rollouts.

\subsection{Effect of Friction on Tool-Screw Contact}
\label{app:friction_effect}
To evaluate the role of friction in tool-screw interaction, we compare a frictionless ODE (Eq.~\ref{eq:ode0}) with a friction-aware ODE (Eq.~\ref{eq:ode_fri}), while keeping all other parameters identical. This simulation illustrates how friction alters the contact outcome under the same control policy and horizontal misalignment.

\begin{figure}[!ht]
    \centering
    \includegraphics[width=0.9\columnwidth]{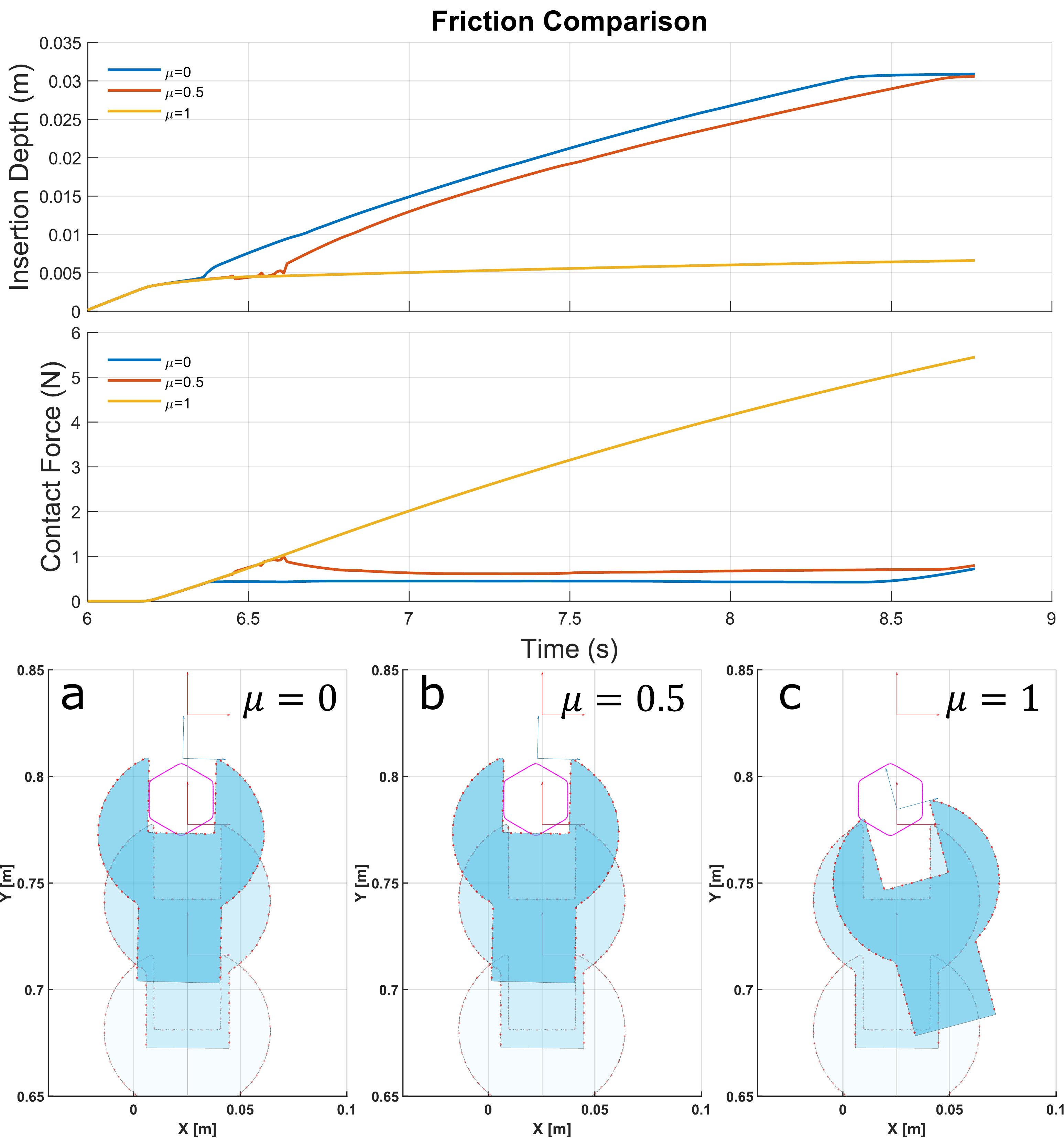}
    \caption{Effect of friction on direct insertion under horizontal misalignment. Low friction enables the tool to slide into the screw head, whereas high friction causes jamming at the contact boundary.}
    \label{fig_fri}
\end{figure}

As shown in Fig.~\ref{fig_fri}, when friction is not modeled ($\mu = 0$), the tool smoothly slides into the screw head despite the horizontal offset. The insertion depth increases rapidly and monotonically, and the contact force remains low and nearly constant. For a moderate friction coefficient ($\mu = 0.5$), insertion is still successful but significantly slower. This slowdown is accompanied by an elevated contact force, indicating increased resistance along the contact boundary. When friction is further increased ($\mu = 1$), the tool becomes jammed at the contact boundary and fails to enter the screw. In this case, the insertion depth quickly saturates, while the contact force grows monotonically over time. These results show that, under horizontal misalignment, low friction enables passive sliding into the screw, whereas high friction leads to contact jamming.

\bibliographystyle{ieeetr}
\bibliography{ref}


 





\end{document}